\documentclass[journal]{IEEEtran}

\usepackage{moreverb,url}
\usepackage{subfiles}
\usepackage{algorithm}
\usepackage[noend]{algpseudocode}
\usepackage{tablefootnote}

\usepackage[colorlinks,bookmarksopen,bookmarksnumbered,citecolor=red,urlcolor=red]{hyperref}
\usepackage[justification=centering]{caption}

\usepackage[noadjust]{cite}
\usepackage{filecontents}
\usepackage{cite}

\usepackage{epsfig}
\usepackage{graphicx}
\usepackage{amsmath}
\usepackage{amssymb}
\usepackage{mathrsfs}

\usepackage{float}
\usepackage{nth}

\ifCLASSINFOpdf

\else

\fi

\begin{document}

\title{Point-cloud-based place recognition using CNN feature extraction}

\author{Ting Sun$^{1}$, Ming Liu$^{1}$, Haoyang Ye$^{1}$, Dit-Yan Yeung$^{2}$
	\thanks{$^{1}$ Department of Electronic \& Computer Engineering, Hong Kong University of Science and Technology, Hong Kong, China. {\tt\small (tsun, eelium)@ust.hk} }
	\thanks{$^{2}$ Department of Computer Science, Hong Kong University of Science and Technology, Hong Kong, China. {\tt\small dyyeung@cse.ust.hk}}%
}

\markboth{Journal of \LaTeX\ Class Files,~Vol.~6, No.~1, January~2007}%
{Shell \MakeLowercase{\textit{et al.}}: Bare Demo of IEEEtran.cls for Journals}

\maketitle

\begin{abstract}
This paper proposes a novel point-cloud-based place recognition system that adopts a deep learning approach for feature extraction.  By using a convolutional neural network pre-trained on color images to extract features from a range image without fine-tuning on extra range images, significant improvement has been observed when compared to using hand-crafted features.  The resulting system is illumination invariant, rotation invariant and robust against moving objects that are unrelated to the place identity. Apart from the system itself, we also bring to the community a new place recognition dataset containing both point cloud and grayscale images covering a full $360^\circ$ environmental view. In addition, the dataset is organized in such a way that it facilitates experimental validation with respect to rotation invariance or robustness against unrelated moving objects separately.
\end{abstract}


\IEEEpeerreviewmaketitle

\section{Introduction}
\label{sec:Introduction}
In autonomous driving, place recognition is to recognize a previously memorized place when the vehicle revisits it.  With a stored map, place recognition can be used for localization when a high quality global positioning system (GPS) signal is unavailable.  In simultaneous localization and mapping (SLAM), place recognition performs loop-closure detection, which is crucial for drifting error correction.  In contrast to localization, place recognition only concerns the location of the robot, regardless of its orientation.  A well-performing place recognition system is expected to correctly identify a previously visited place with a high probability in real-time.

Most of the place recognition methods are based on images \cite{scene-sequences,FAB-MAP,SeqSLAM,correct-loop,HMM,binary-sequences,slam-across-seasons,location_models,covisibility-graph}, and few have reported using LiDAR \cite{3Dlidar}. The limitation of using images is that they are variant to illumination change, as cameras are passive photoreceptive sensors.  LiDARs are active illumination invariant sensors.  In this work we adopt Velodyne\footnote{\url{http://velodynelidar.com/}} for input, since its generated point-cloud covers $360^\circ$ in the horizontal direction isotropically, and this enables us to design a rotation-invariance system, which is necessary for place recognition.  As similar sensors become increasingly popular in autonomous driving \cite{zhang2010lidar,zhu20123d,kitti,ibisch2013towards} and the cost has been greatly reduced, it is promising and practical to conduct place recognition using similar sensors.

The traditional place recognition methods first detect key points, where some hand crafted features like SIFT \cite{SIFT} are extracted.  These features are commonly with high computational cost, and are further encoded by bag-of-words, then recognition is conducted by matching the encoded indexes \cite{FAB-MAP,steder2011place}.  Since deep learning approaches have achieved state-of-the-art performance in many challenging computer vision tasks \cite{Imagenet,DeCAF,cnn-feature-off-the-shelf,very-deep,going-deeper}, there is a natural momentum to attempt place recognition using deep learning.  In particular, convolutional neural networks (CNNs) \cite{convnet} provide a powerful end-to-end framework for image-based vision tasks. Such a structure would also greatly inspire place recognition using LiDAR. The advantages of CNNs include the following:

\begin{itemize}
	\item It directly takes raw images as input;
	\item It extracts hierarchical features automatically;
	\item Both the convolutional layer and pooling layer make the features of the higher layer shift invariant to some extent;
	\item Due to common characteristics of natural images, such as analogy, the CNN models trained on one dataset can be transferred to other datasets.
\end{itemize}

However, it is highly nontrivial to leverage a CNN to extract features from an unstructured point cloud. This is because images are typically a ``dense representation'', for which each pixel in the image configuration space has a defined intensity value.  Conversely, point-cloud is a ``sparse representation'', for which not \textit{all} (abstractly) points in the configuration space are defined. Only the locations with point observations are informative.  Point clouds are much more expensive to obtain than color images, but training a deep CNN is data demanding.  The features extracted by a deep CNN in different layers have different levels of abstraction, and its worth studying how the performance varies with the choice of layers.  In brief, the solutions to the following issues are critical to use CNN with point-clouds:

\begin{itemize}
	\item A mapping to convert an unstructured point cloud to an image is to be defined;
	\item The selection of a proper CNN model should be advised;
	\item The representation of the extracted deep feature should be associated with a layer of the network;
	\item The features directly obtained from a pre-trained CNN are redundant and noisy. Efficient and sufficient post-processing is necessary for more compact feature description.
\end{itemize}

In this paper, we propose a point-cloud-based place recognition method using a CNN for feature extraction.  A point-cloud is first aligned with its principal directions, then converted to a range image.  A CNN pre-trained on abundant RGB images is used to extract features from the range images.  Principal component analysis (PCA) is used for further dimension reduction following our previous guideline \cite{tai2016pca}.  A scoring mechanism that concerns both cosine similarity and the discrimination of the best match among the top matches is used to make a decision. 

A good place recognition system is expected to be invariant to illumination change and rotation, as well as to moving objects that are irrelevant to the place \cite{sun2017improving}.  However, to the limit of our knowledge, there is no place recognition dataset that deliberately separates its content according to these three aspects in this context.  In this paper, we introduce a place recognition dataset containing grayscale images (the images are not directly used in this paper) and point-clouds.  The images are taken under severe illumination change and both types of data cover a full $360^\circ$ environmental view.  The content of our dataset is organized to especially facilitate tests of yaw rotation, which is the primary rotation case in autonomous driving, invariance, and robustness to moving objects separately.  Details of the dataset are give in Subsec.~\ref{subsec:dataset}.  

To summarize, we stress the following contributions in this paper:

\begin{enumerate}
	\item We propose a novel end-to-end point-cloud-based place recognition system using CNN feature extraction, which is:
	\begin{itemize}
		\item illumination invariant,
		\item rotation invariant, and
		\item robust to moving objects.
	\end{itemize}
	\item We analyze the properties of the system by testing the effectiveness of each module;
	\item We introduce a new dataset for place recognition.
\end{enumerate}

The remainder of this paper is organized as follows. Sec.~\ref{sec:Related work} reviews some previous work on place recognition and the related CNN study. Our proposed method is presented in Sec.~\ref{sec:Proposed method}, which is then followed by experiment results and discussion in Sec.~\ref{sec:Experiments}. Sec.~\ref{sec:Conclusion} concludes the paper.

\section{Related work}
\label{sec:Related work}
\subsection{Place recognition}
Place recognition contains two main steps: feature extraction and feature retrieval. We separately introduce the related work in the two domains as follows.

\subsubsection{Feature extraction and description}

Descriptors are crucial, and improvement of illumination invariance, viewpoint invariance and calculation efficiency are the three main research directions of feature extraction and description. 

In the traditional visual place recognition methods, descriptors roughly fall into two categories \cite{survey}: local descriptors and global descriptors (or holistic descriptors \cite{BRIEF-Gist}).  Local descriptors are extracted around detected keypoints like corners, while global descriptors describe the whole image.  Widely adopted local descriptors include scale-invariant feature transforms (SIFT) \cite{SIFT}, speeded-up robust features (SURF) \cite{SURF}, binary robust independent elementary features (BRIEF) \cite{BRIEF}, binary robust invariant scalable keypoints (BRISK) \cite{BRISK}, oriented FAST and rotated BRIEF (ORB) \cite{ORB}, local difference binary (LDB) \cite{LDB,D-LDB} etc.  Since the number of detected keypoints in each frame varies and directly matching the features can be inefficient \cite{survey}, the bag-of-words model is used to further encode the local features to facilitate frame-wise comparison.  Global descriptors of a frame can be obtained by integrating the local descriptors \cite{BRIEF-Gist,WI-SURF}, or by directly extracting them from the whole image \cite{color-hist-map,cnn-performance}.  A trade-off between local and global descriptors is explored in \cite{landmarks-cnn,semi-semantic}, where features are extracted from the object region proposed by edge boxes \cite{edge-boxes}, and \cite{liu2014topological} proposes a lightweight adaptive line descriptor based on color features and geometric information.

As a variety of range sensors become popular, place recognition starts to benefit from this new type of input.  In \cite{range-salient},  range value is used to obtain scale information assisting the detection of salient regions.  \cite{pcl-loop-clousure} manually extracts 41 rotation invariant features from each frame of a 3D point cloud, and adopts AdaBoost to train a binary classifier to distinguish positive and negative laser pairs.  \cite{pcl-descriptors} designs a local feature extracted from a point cloud, called a neighbor-binary landmark density descriptor (NBLD), and extracts the NBLD from detected keypoints to recognize places through a voting framework.  However, point-cloud-based feature extraction methods are far from mature compared with image-based methods.  It is more difficult to identify keypoints, lines, and objects in an unstructured point cloud than in an image, and the increased dimension also intensifies the computational cost.  Some extraordinary frameworks like CNN are not designed for point clouds and it is definitely worth trying to design a point-cloud-based system so that the power of deep learning approaches can be leveraged.

\subsubsection{Feature Retrieval}

With the extracted descriptor of a new frame, retrieval is used to tell whether the current place matches a previously visited place, and if the answer is yes, where the place is.  In most cases, the retrieval method is designed independently of descriptors, and is less time consuming to compute than feature extraction for most of the datasets, unless the stored map is of a very large scale.

Based on different concerns and assumptions, well-designed retrieval methods help to increase the precision-vs-recall performance.  \textit{E.g.} by exploring the structure of the bag-of-words data \cite{FAB-MAP,liu2012dp,Generative} or covisibility of landmarks \cite{location_models,covisibility-graph} to reduce perceptual aliasing; using the assumption that sequential frame queries are of adjacent or the same place as prior to narrow down the search range \cite{FAB-MAP}; based on the assumption that the vehicle will repeat the same path rather than just revisit a single place, sequence matching is adopted \cite{scene-sequences,SeqSLAM,correct-loop,HMM,slam-across-seasons,binary-sequences}; \cite{correct-loop} propose a method to correct the wrong recognition when new information comes; the time consumed by many retrieval methods is propotional to the number of places stored, and this time can be reduced by narrowing down the search range as mentioned before, or using a kd-tree \cite{2D-lidar}, hashing \cite{GmH} etc.

Our focus in this work is point-cloud-based feature extraction using a CNN.  In order to test the feature quality, we show precision-vs-recall performance using single scan feature matching without special prior.  However, if the assumptions hold, any retrieval approaches can be adopted to further improve the performance.

\subsection{Convolutional neural networks}

Deep learning \cite{deep-learning,deep-learning-overview} approaches focus on learning features directly from raw data in a hierarchical manner.  In particular, CNNs \cite{convnet} provide a powerful end-to-end framework that achieves state-of-the-art performance in many challenging computer vision tasks \cite{Imagenet,DeCAF,cnn-feature-off-the-shelf,very-deep,going-deeper,my_fine_grained}.  The rich features learned by a deep CNN ranging from low-level to high-level representations in the hidden layers have also aroused extensive research interest in investigating how to take advantage of them \cite{Visualizing-cnn,learning-deep-features,scene-recognition,my_fine_grained}.  Moreover, the public availability of efficient CNN implementations \cite{Caffe}, powerful pre-trained CNN models \cite{Imagenet,return-devil,very-deep,going-deeper,caffenet} and their ability to transfer to work on other tasks even without fine-tuning has further popularized the pervasive use of CNNs for various applications. Not surprisingly, visual place recognition also turns to CNNs for feature extraction \cite{cnn-performance,landmarks-cnn,slam-across-seasons,training_cnn,survey,semi-semantic}.  Our method is point-cloud-based, and in the next section we demonstrate how we leverage the feature extraction power of a CNN.

\section{Proposed method}
\label{sec:Proposed method}
\begin{figure*}
 	\centering
 	\captionsetup{justification=centering}
 	\includegraphics[width=\textwidth]{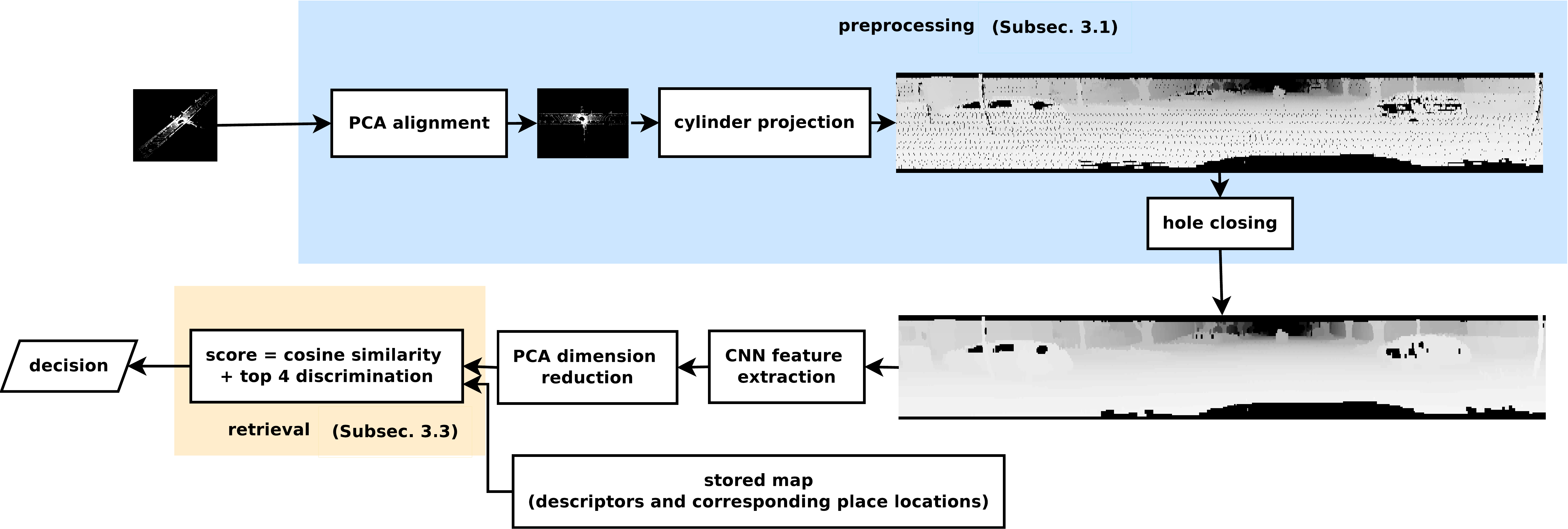}
 	\caption{System overview}
 	\label{fig:system_overview}
 \end{figure*}

 The overview of our proposed place recognition system is shown in Figure~\ref{fig:system_overview}.  A point cloud is first aligned with its principal directions, then is projected onto a cylinder image plane.  After hole-closing, a CNN is used for feature extraction, followed by PCA dimension reduction.  Retrieval is based on a score concerning both similarity and discrimination.  A threshold applied to the score is used to trade off between precision and recall.  The motivation and design details of each module are illustrated in the following.  The last subsection will introduce our new data set for place recognition.  
 
 \subsection{Preprocessing}
 \label{subsec:Preprocessing}
\begin{figure}
 	\centering
 	\captionsetup{justification=centering}
 	\includegraphics[width=0.45\textwidth]{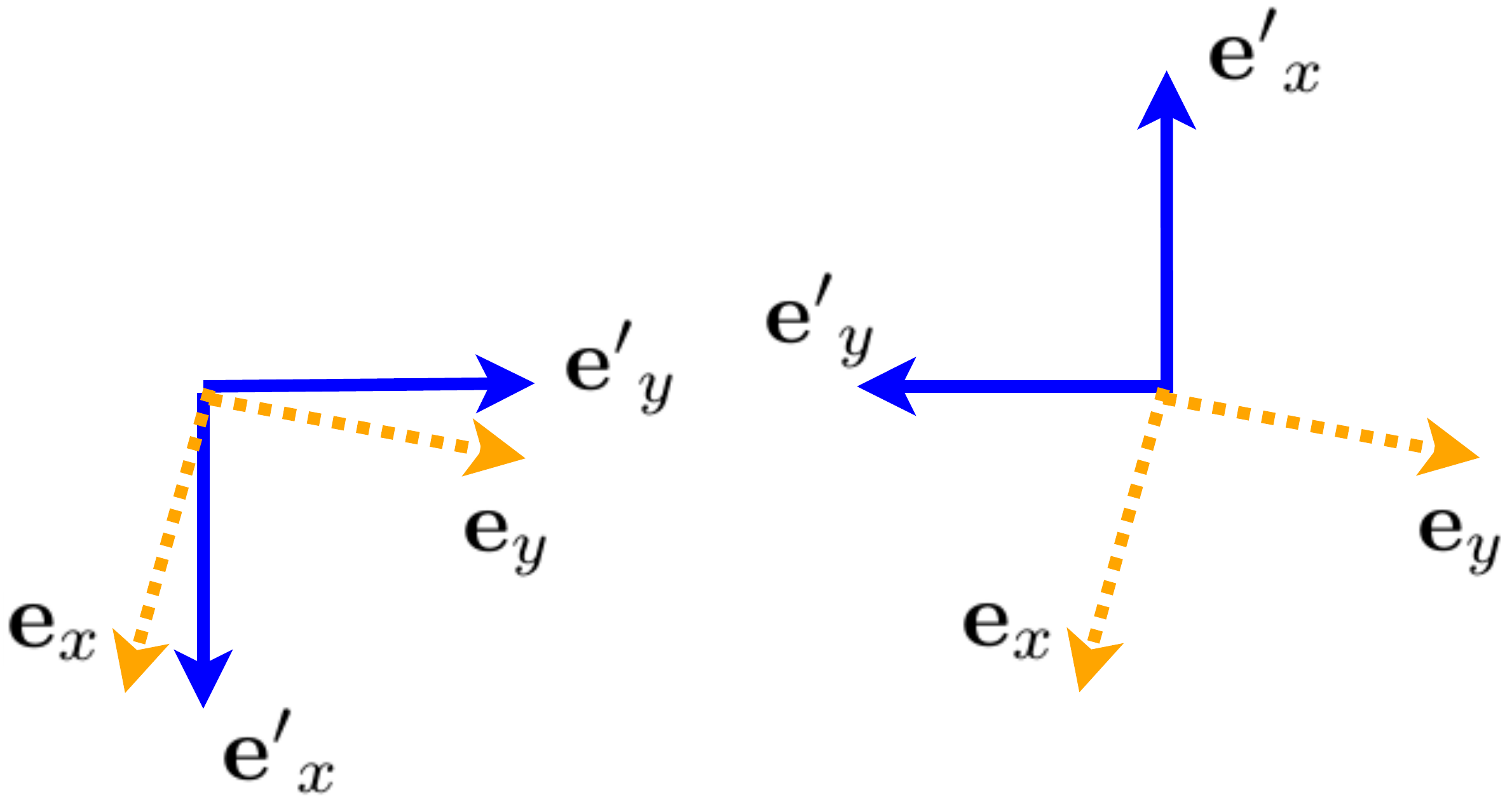}
 	\caption{Two cases of PCA alignment.}
 	\label{fig:pca align}
 \end{figure}

 The preprocessing module is highlighted by a light blue rectangle in Figure~\ref{fig:system_overview}.  The output of this module is a range (i.e. grayscale) image that can be put into a CNN.  The input we consider is a 3D point cloud that covers a full $360^\circ$ environmental view.  This kind of point cloud is commonly created by a Velodyne LiDAR.  In order to align the point cloud, PCA is adopted to find the orthogonal directions sorted by the variance.  However, the resultant bases are not unique, there are 8 possible combinations of signs.  Considering the typical autonomous driving environment, we narrow down the cases to two.  Let's denote the original basis of the point cloud as $ \mathbf{B} = [\mathbf{e}_x, \mathbf{e}_y, \mathbf{e}_z ] $, the basis obtained by PCA as $\mathbf{B'} = [ \mathbf{e'}_x, \mathbf{e'}_y, \mathbf{e'}_z ] $ and $ \mathbf{B'} = \mathbf{T} \mathbf{B}$.  The two cases that we keep satisfy one of the two following constraints:
 
 \begin{equation}
 \label{equ:cond1}
 t_{11}, t_{22}, t_{33} >= 0 
 \end{equation}
 
 or
 
 \begin{equation}
 \label{equ:cond2}
 t_{11}, t_{22} < 0; t_{33} >= 0,
 \end{equation}
 
 where $t_{ij}$ is the element of $\mathbf{T}$ in $i^{th}$ row and $j^{th}$ column.  Following the setup of the KITTI dataset~\cite{kitti}, i.e. $\mathbf{e}_x$ points to the front of the car and $\mathbf{e}_z$ points up.  For a common street view point cloud, it is reasonable to assume in the PCA basis $\mathbf{B'}$, $\mathbf{e'}_x$ and $ \mathbf{e'}_y$ span a plane which is roughly parallel to the ground since they are the directions that capture most of the variance.  In order to satisfy either (\ref{equ:cond1}) or (\ref{equ:cond2}), $\mathbf{e'}_z$ will always point up, and $[\mathbf{e'}_x, \mathbf{e'}_y]$ will be one of the cases in Figure~\ref{fig:pca align}.  As will be shown in Sec.~\ref{sec:Experiments}, for unidirectional loop closure, using one alignment is enough, considering both cases is mainly to handle bidirectional loop closure.
 
 We then create the range image from the aligned point cloud using the Point Cloud Library (PCL) \cite{pcl} implementation.  One parameter we would like to mention is angular resolution\footnote{The term `resolution' here means the angle range corresponding to one pixel in the range image.  In the next subsection, `resolution' means the number of pixels of an image.}, which decides how fine-grained the grid in the cylinder image is when the range image is generated.  If the angular resolution is set too small, there will be a lot of empty grids; and if it is too big, the grid will be too rough to preserve the details.  We suggest setting the angular resolution similar to that of the LiDAR sensor used, then filling the small holes using morphological closing, i.e. dilation followed by erosion.

 \subsection{CNN feature extraction}
 \label{subsec:CNNs}
 As mentioned in Sec.~\ref{sec:Introduction}, leveraging a CNN to extract features from an unstructured point cloud is highly nontrivial.  It is laborious and expensive to collect a large amount of point cloud data to train a model from scratch, and we want to take advantage of the abundant pre-trained ones.  Currently most of these CNN models are trained on RGB images of a square shape \textit{e.g.} 224:224, but the range images we generate from point clouds are grayscale and of a very long rectangle shape about 4000:100.  The range image is simply repeated 3 times to fit the color channels; but for the spatial resolution, naive resizing will severely distort the ratio, and brutal down sampling will cause great loss of details.  We propose to preserve the original horizontal resolution of the image and may increase its vertical resolution to ensure that it will not reduce to zero through the pooling layers of a CNN.  
 
 A deep CNN trained on a large dataset can extract hierarchical generic features for other tasks \cite{DeCAF}.  From the lower layer to the higher layer, the precision of the hidden activation decreases while the abstraction and invariance increase \cite{Visualizing-cnn,learning-deep-features,scene-recognition}.  We try to find the proper precision-abstraction trade-off by testing multiple layers of the networks \cite{return-devil}, as presented in Sec.~\ref{sec:Experiments}.  Our range image is generated by projecting the point cloud on a cylinder plane then unfolding it,  so the rotation of the point cloud around the axis of the cylinder becomes a shift in the resultant range image.  The feature extracted by a CNN is spatially invariant to some extent,  which contributes to the rotation invariance property of our system. 
 
 In general, a deeper and wider network is potentially more powerful \cite{caffenet}.  We test our proposed method on three CNNs i.e. AlexNet \cite{Imagenet}, VGG-CNN-S \cite{return-devil} and Places-CNDS-8 \cite{caffenet}.  VGG-CNN-S \cite{return-devil} is modified from AlexNet \cite{Imagenet} by increasing the channels of the last three convolutional layers, and Places-CNDS-8 \cite{caffenet} has a deeper network structure.  Because of the length of the input image, the size of the hidden layer maps is large.  Notice that although the linear transformations conducted by fully connected layers perform reshaping and dimension reduction, we cannot use the features processed by them; since the shape of our range image is different from that of the original CNN model input, the pre-trained weights of fully connected layers can not be loaded due to parameter number mismatch.
 
 The features directly obtained from the CNN are highly redundant.  The redundancy is introduced by: 
 
 \begin{enumerate}
 	\item the repetition of the 3 color channel input;
 	\item the uninformative region of the range image, like the boundary and the floor; and
 	\item the domain knowledge that only applies to the data used to train the CNN, but not to our application scenario.
 \end{enumerate}
 
 In our system, features are extracted from convolutional layers or pooling layers.  One more pooling layer may be manually inserted on top of the chosen hidden layer for preliminary dimension reduction.  PCA is adopted as a postprocessing step to obtain more compact features.  This is due to the nature of the place recognition task and the global descriptor we use.  Since one place is supposed to have one descriptor, the variance of each dimension indicates its discrimination ability. 
 
 \subsection{Retrieval}
 \label{subsec:retrieval}
 We normalize the descriptor vector of each point cloud and use cosine distance as the similarity metric.  Most image-based retrieval methods only consider the best match in the memory.  On the contrary, our method jointly considers the top k best match and shows that this design leads to a better result.  The proposed retrieval method is described in Algorithm~\ref{alg:retrieval}, where $\mathbf{f}^{i}$ is the feature of place $i$, $\mathcal{S}^{stored}_{\mathbf{f}}$ is the set of stored features of previously visited places, $\mathbf{f}^{query}$ is the feature vector of the current scan,  $\mathcal{C}(\mathbf{f}^{i}, \mathbf{f}^{j})$ returns the cosine similarity between $\mathbf{f}^{i}$ and $\mathbf{f}^{j}$, $s$ is a score we define, and $threshold$ is used to trade off between precision and recall.

 \begin{algorithm}[h!!]
 	\caption{Retrieval}
 	\label{alg:retrieval}
 	\begin{algorithmic}[1]
 		\Require ~~\ 
 		$\mathbf{f}^{query}$, $\mathcal{S}^{stored}_{\mathbf{f}}$
 		\Ensure ~~\ 
 		return place identity if match found, otherwise report not found 
 		\State $\mathcal{C}(\mathbf{f}^{query}, \mathbf{f}^{i_1}) > \mathcal{C}(\mathbf{f}^{query}, \mathbf{f}^{i_2}) \cdots > \mathcal{C}(\mathbf{f}^{query}, \mathbf{f}^{i_k}) > \cdots$ \Comment{find top k match}
 		\State $s =  \mathcal{C}(\mathbf{f}^{query}, \mathbf{f}^{i_1}) \times 2 - \mathcal{C}(\mathbf{f}^{query}, \mathbf{f}^{i_{k}})$
 		\If {$s > threshold$}
 		\State return ${i_1}$
 		\Else 
 		\State report not found
 		\EndIf
 	\end{algorithmic}
 \end{algorithm}

\begin{figure*}
 	\centering
 	\includegraphics[width=0.6\textwidth]{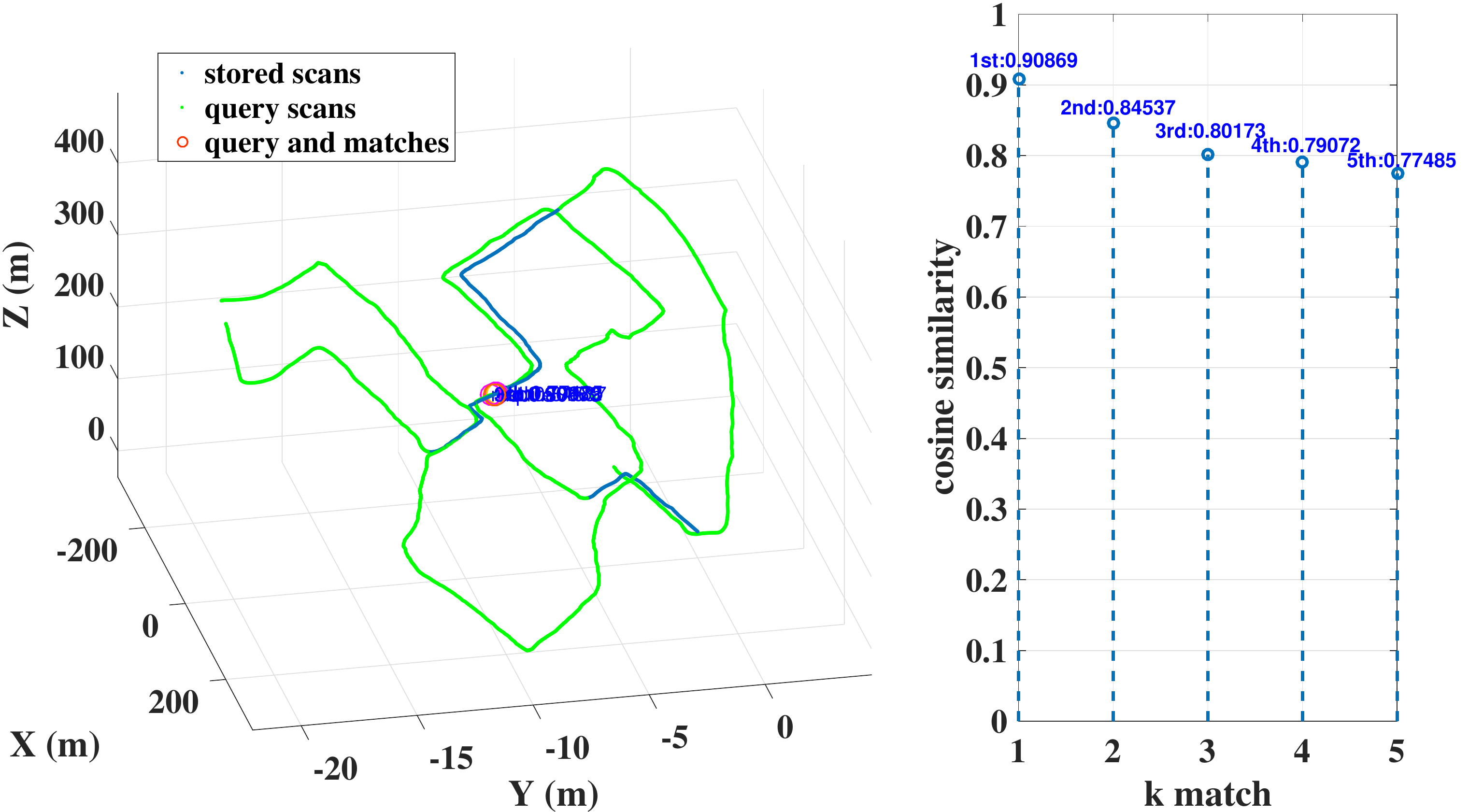}
 	\includegraphics[width=0.3\textwidth]{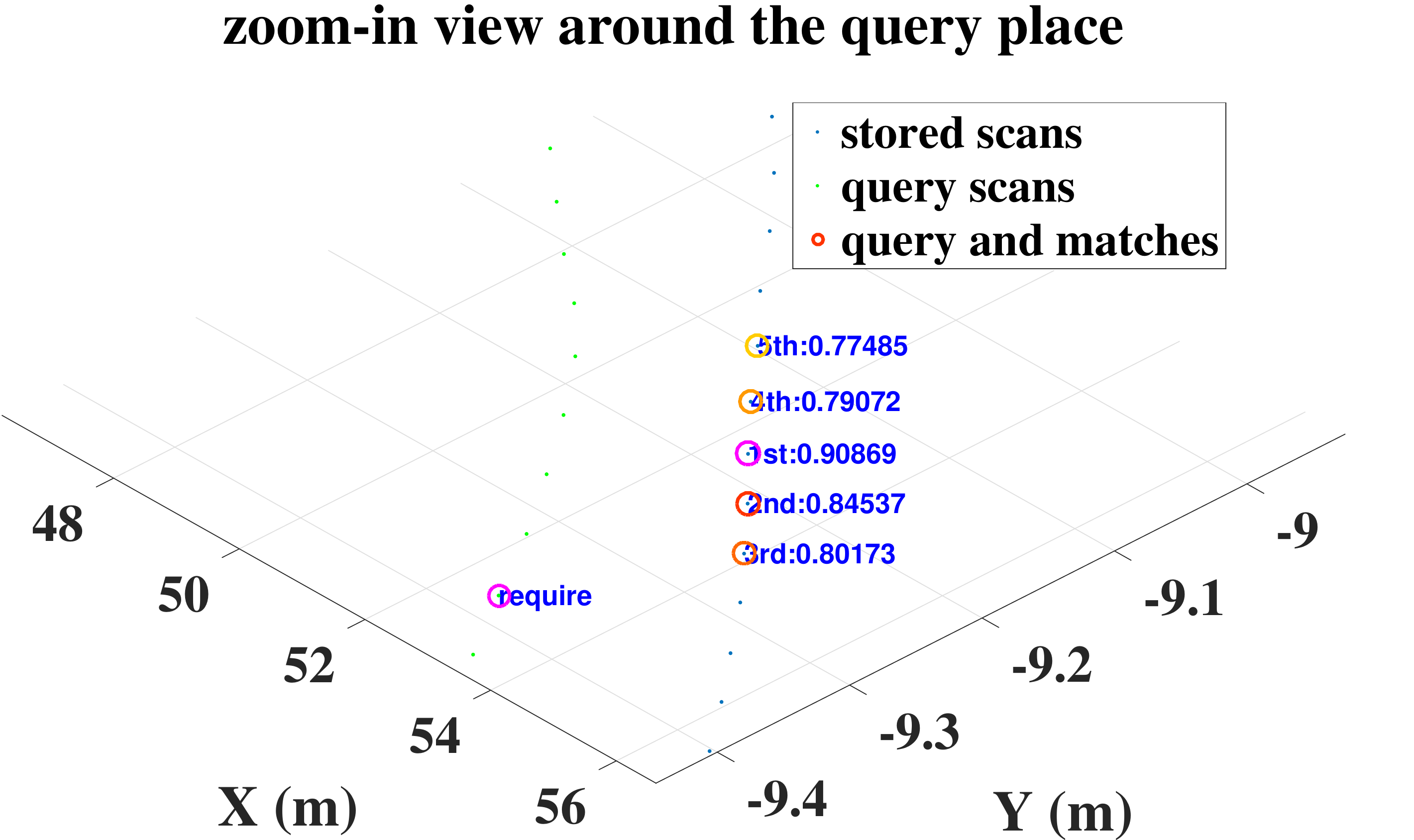}
 	\includegraphics[width=0.6\textwidth]{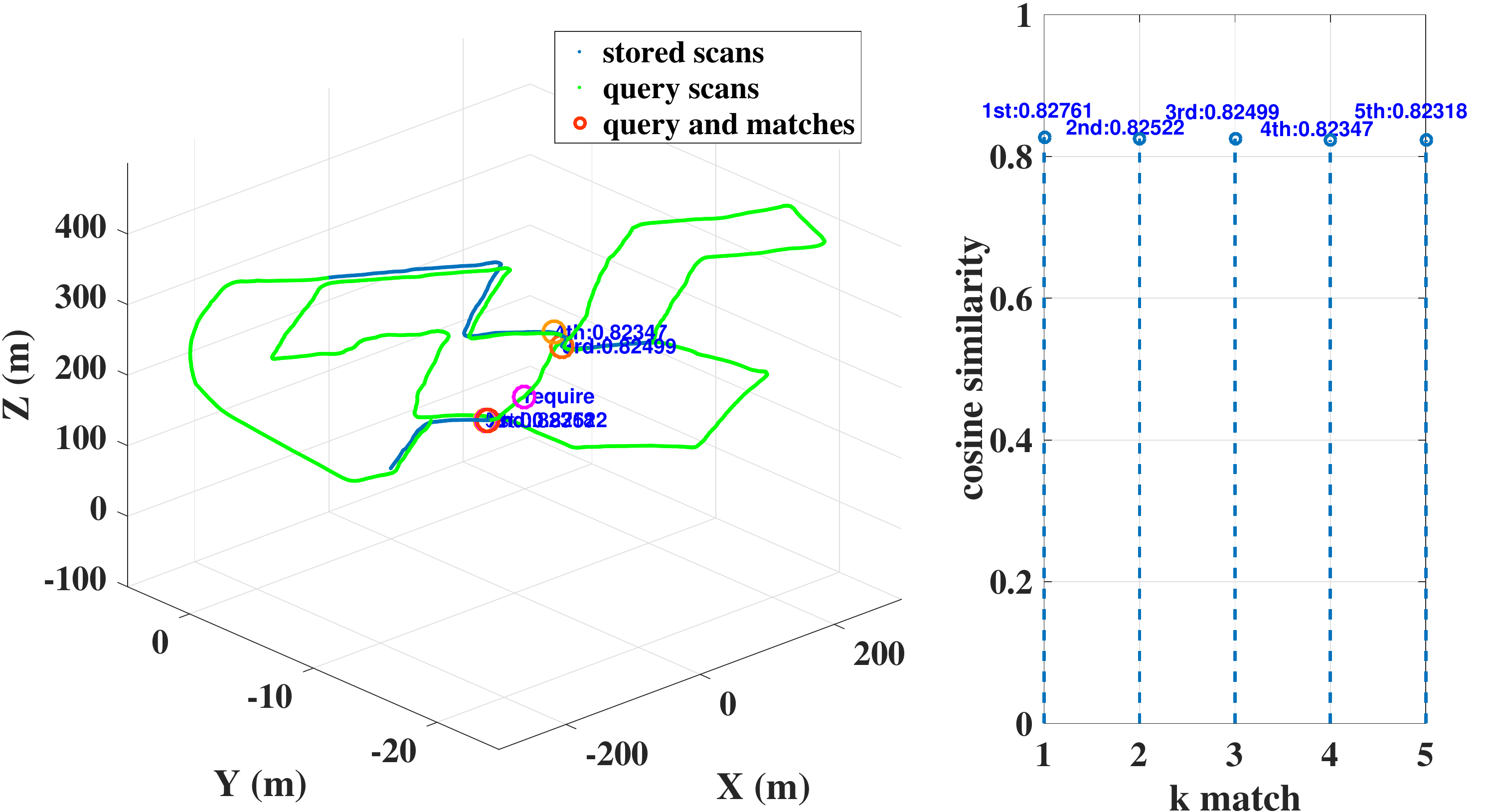}
 	\includegraphics[width=0.3\textwidth]{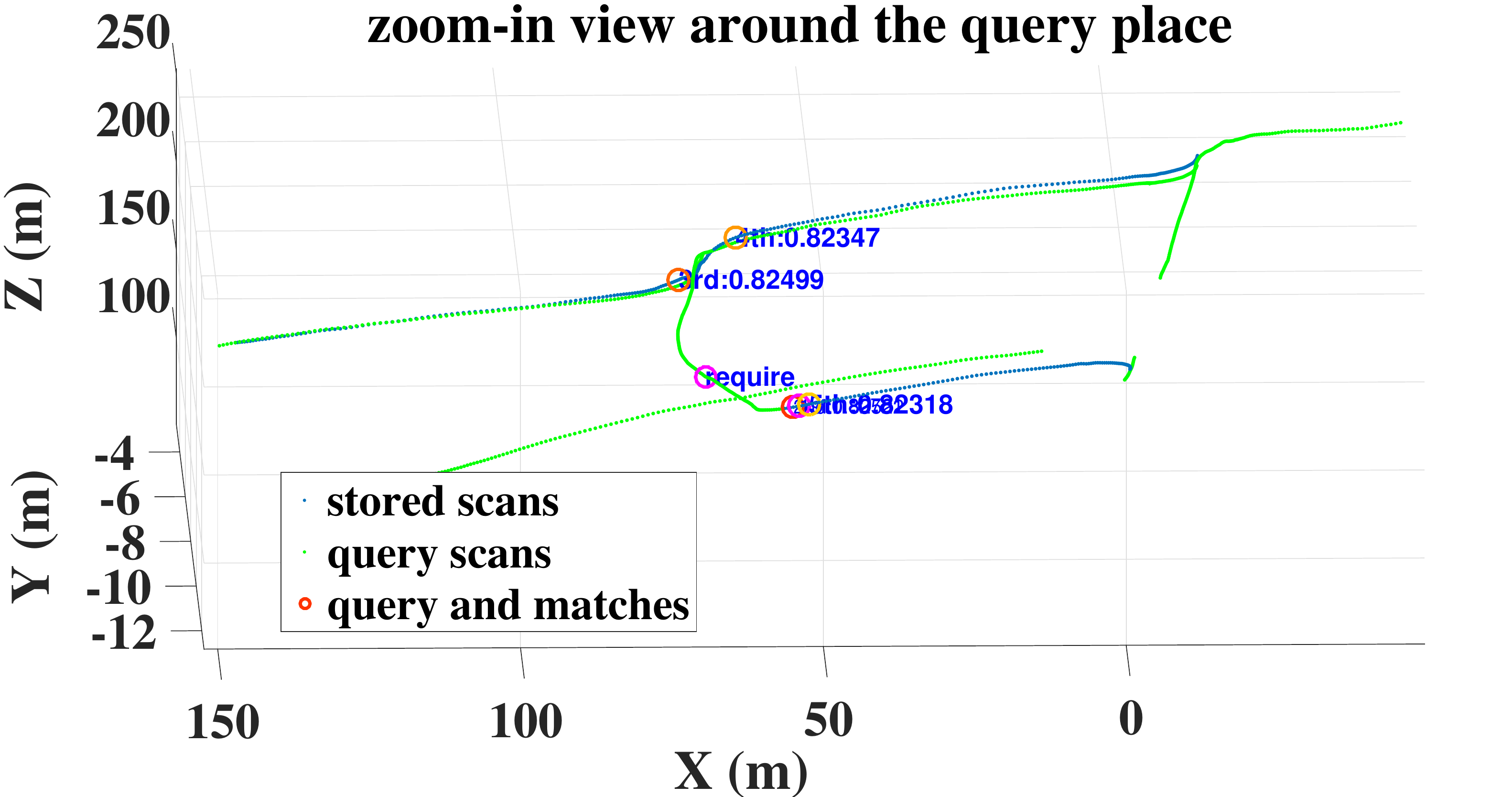}
 	\caption{Two typical retrieval examples in the sequence 00 of the KITTI dataset \cite{kitti} using our proposed method, to illustrate the motivation of the design of score $s$ in Algorithm~\ref{alg:retrieval}.  The blue dots are the manually selected stored scans and the green dots are queries.  The top one shows a case when the required place feature has a match in the memory, while the bottom one shows a case when there is no match.  The left subfigure shows the location of the query and that of its top 5 matches.  The middle subfigure shows the cosine similarity between the query feature and its top 5 matches in the memory.  The right subfigure shows the zoom-in view around the query place.  A more detailed description can be found in the text.}
 	\label{fig:top_5_similarity}
 \end{figure*}

\begin{figure*}
 	\centering
 	\includegraphics[width=0.32\textwidth,height=0.2\textwidth]{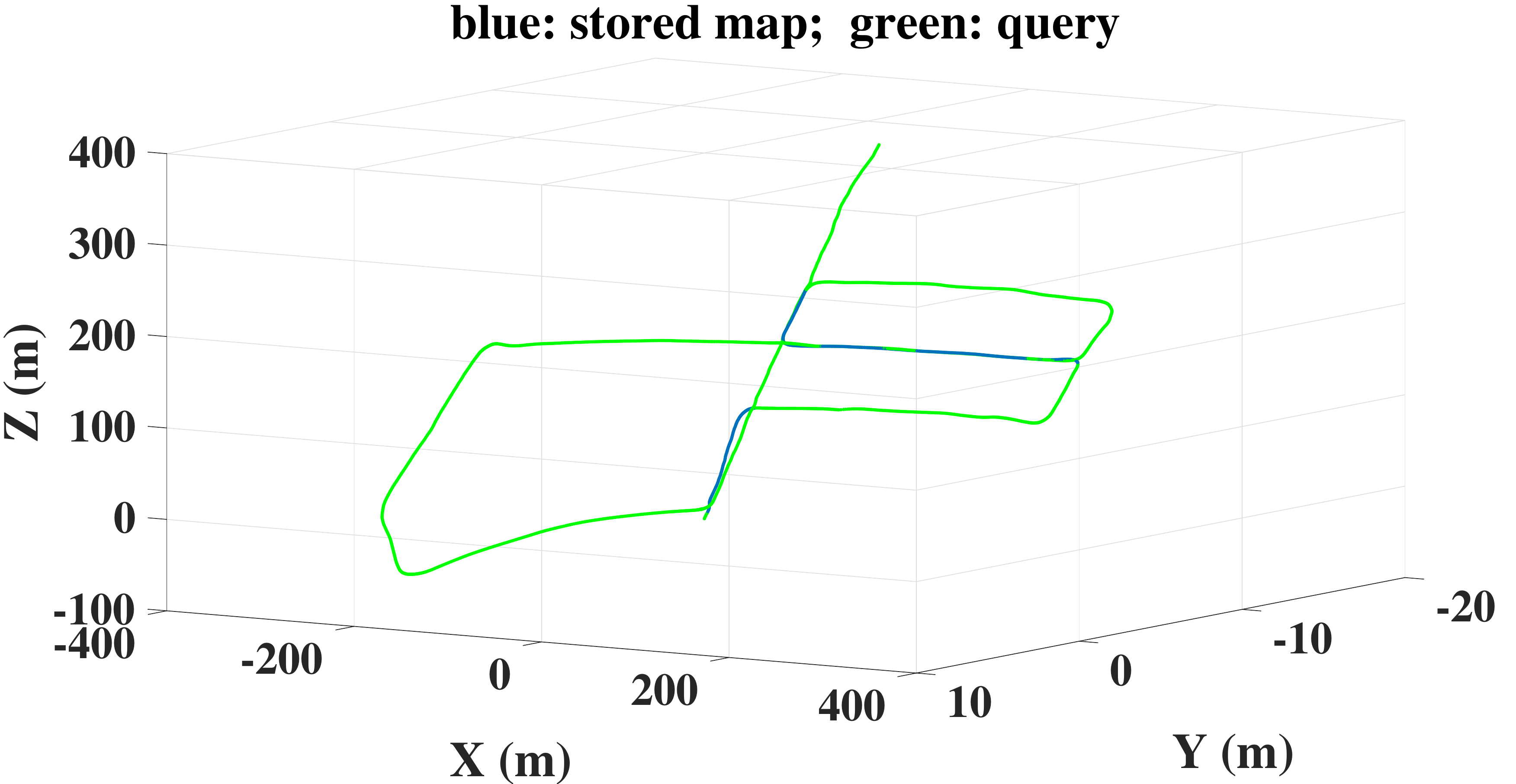}
 	\includegraphics[width=0.32\textwidth,height=0.2\textwidth]{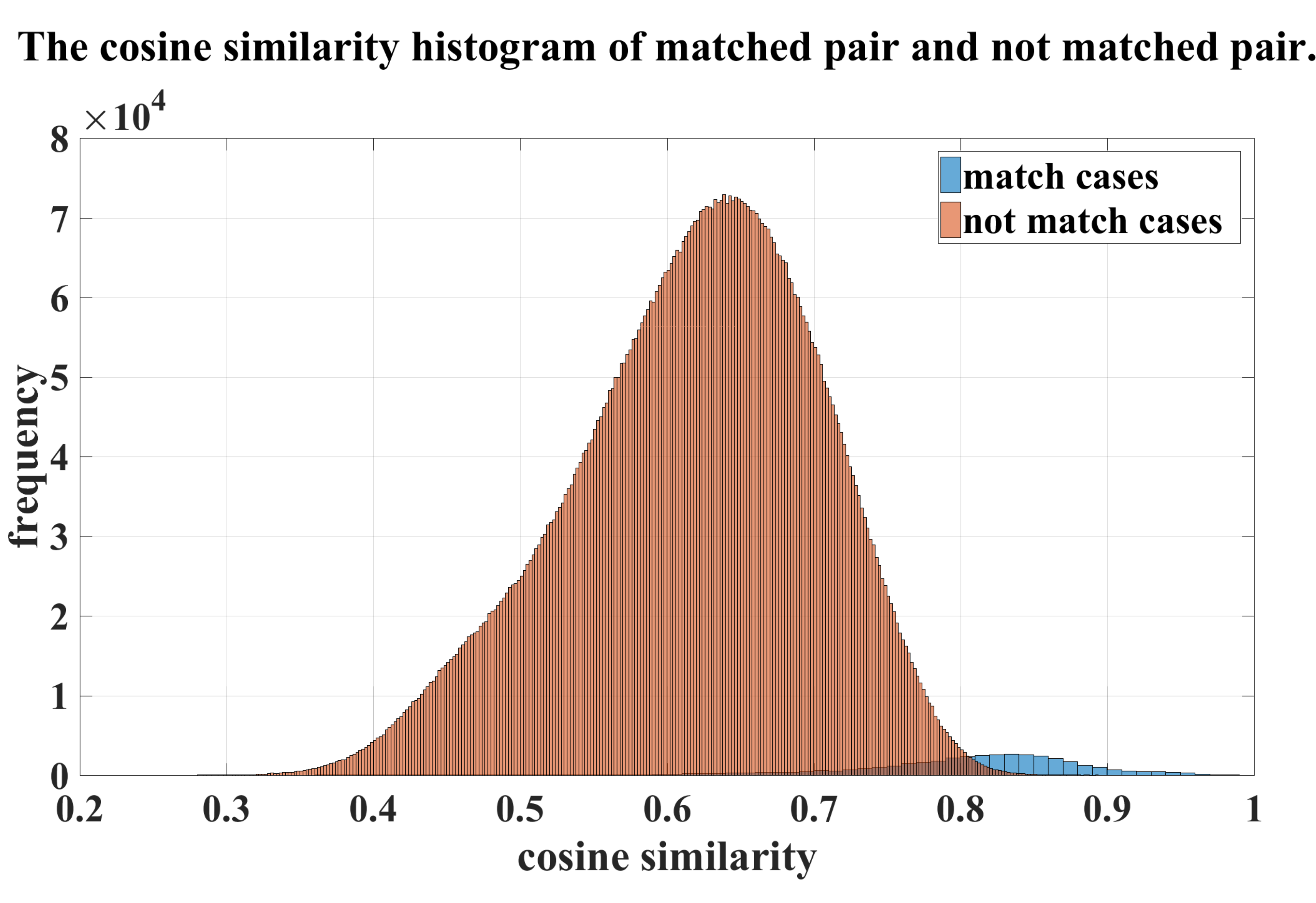}
 	\includegraphics[width=0.32\textwidth,height=0.2\textwidth]{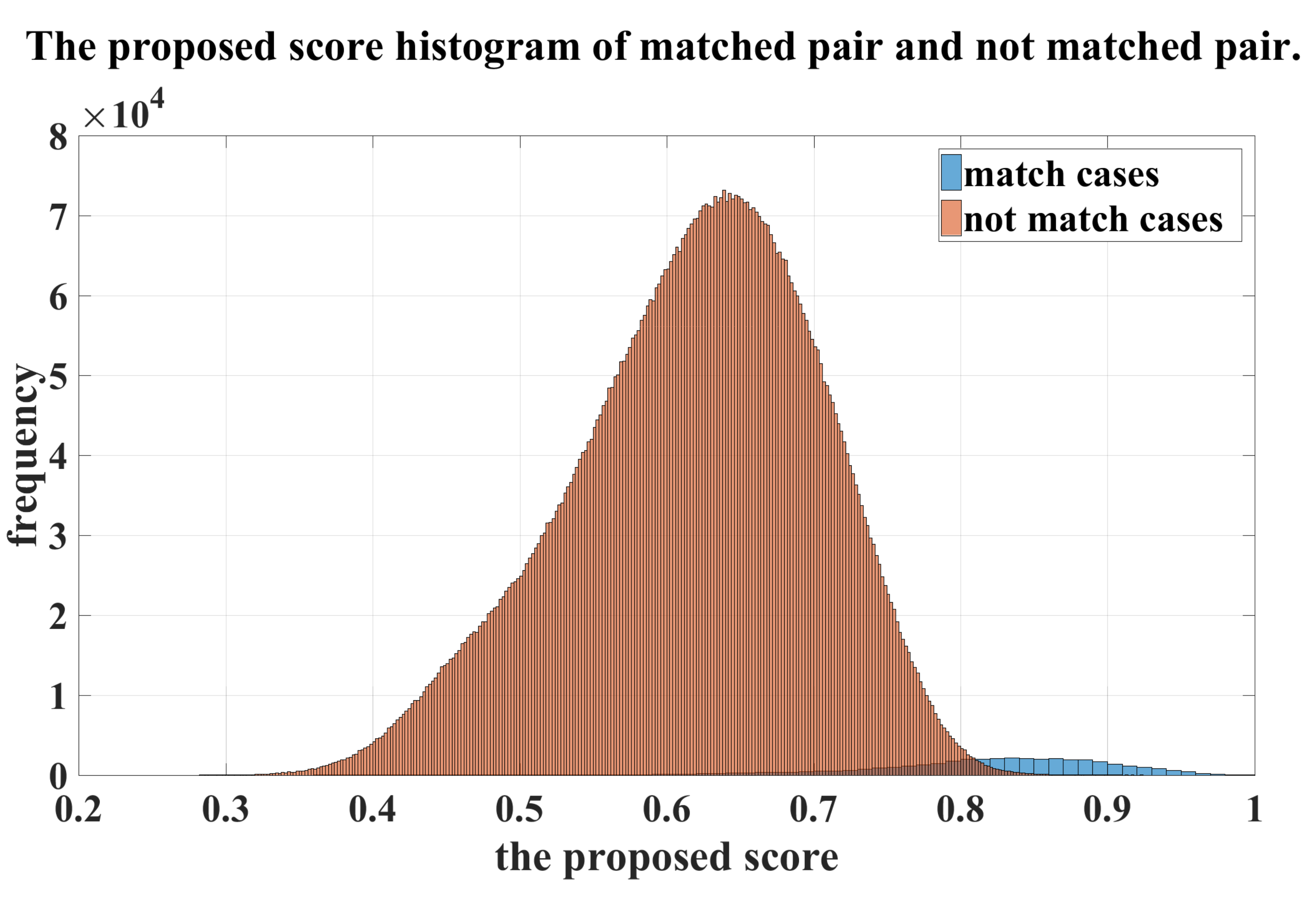}
 	\caption{The effectiveness of proposed score is shown in this figure.  The left subfigure shows sequence 05 from the KITTI dataset \cite{kitti}, where the blue dots are the manually selected stored scans and the green dots are the queries. The middle subfigure plots the histogram of the cosine similarity of all the matched scan pairs in blue, and scans that do not match in orange.  The right subfigure plots the histogram of our proposed score.}
 	\label{fig:score}
 \end{figure*}

 Notice that the score $s =  \mathcal{C}(\mathbf{f}^{query}, \mathbf{f}^{i_1}) \times 2 - \mathcal{C}(\mathbf{f}^{query}, \mathbf{f}^{i_{k}})$ is the sum of two terms: the cosine similarity of the best match $\mathcal{C}(\mathbf{f}^{query}, \mathbf{f}^{i_1})$ and the gap between the best match and the k\textit{th} best match $\mathcal{C}(\mathbf{f}^{query}, \mathbf{f}^{i_1}) - \mathcal{C}(\mathbf{f}^{query}, \mathbf{f}^{i_{k}})$.  The intuition of this design is illustrated in Figure~\ref{fig:top_5_similarity} where two typical retrieval examples in the KITTI dataset \cite{kitti} are shown.  We manually find the revisited trails and use one round as the stored map, colored blue, and leave the rest as queries colored green.  A pair of scans is treated as a true match if all the differences of their 3 coordinates are smaller than 3 meters.  In each example, the query and its top 5 matches, as well as their corresponding cosine similarity, are marked.  The top example shows a case when the required place has a match in the memory, while the bottom example shows a case when there is no match.  After comparison, our key observation is that when a true match exists, it has two properties: 1) it has high similarity with the query and 2) its cosine similarity distinguishes it from the rest by a large margin, which we call the `discrimination gap' in the following text.  In our system, we empirically set k = 4.  We do not use the gap between the 1\textit{st} match and the 2\textit{nd}, considering that it is rare for the vehicle to revisit exactly the same location.  It is more likely to be somewhere between the stored places so there could be a few true matches instead of a unique one.  The effectiveness of this design is shown in Figure~\ref{fig:score}.  The left subfigure shows sequence 05 from the KITTI dataset \cite{kitti}.  The middle subfigure plots the histogram of the cosine similarity of all the matched scan pairs in blue, and those that do not match in orange.  The right subfigure plots the histogram of our proposed score.  Comparing with using cosine similarity only, it can be seen that in the histogram of our proposed score, the blue bars are slightly pushed to the right, while the orange bars have no visible change.  This result is consistent with our observation that the discrimination gap is a property of the true match.  Thus, modifying cosine similarity with the discrimination gap makes the matched scan pairs more distinguishable from the unmatched ones.  This effect is also shown in Subsec.~\ref{subsec:exp_retrieval}.

 \subsection{HKUST dataset}
 \label{subsec:dataset}
\begin{figure}
 	\centering
 	\includegraphics[width=0.4\textwidth]{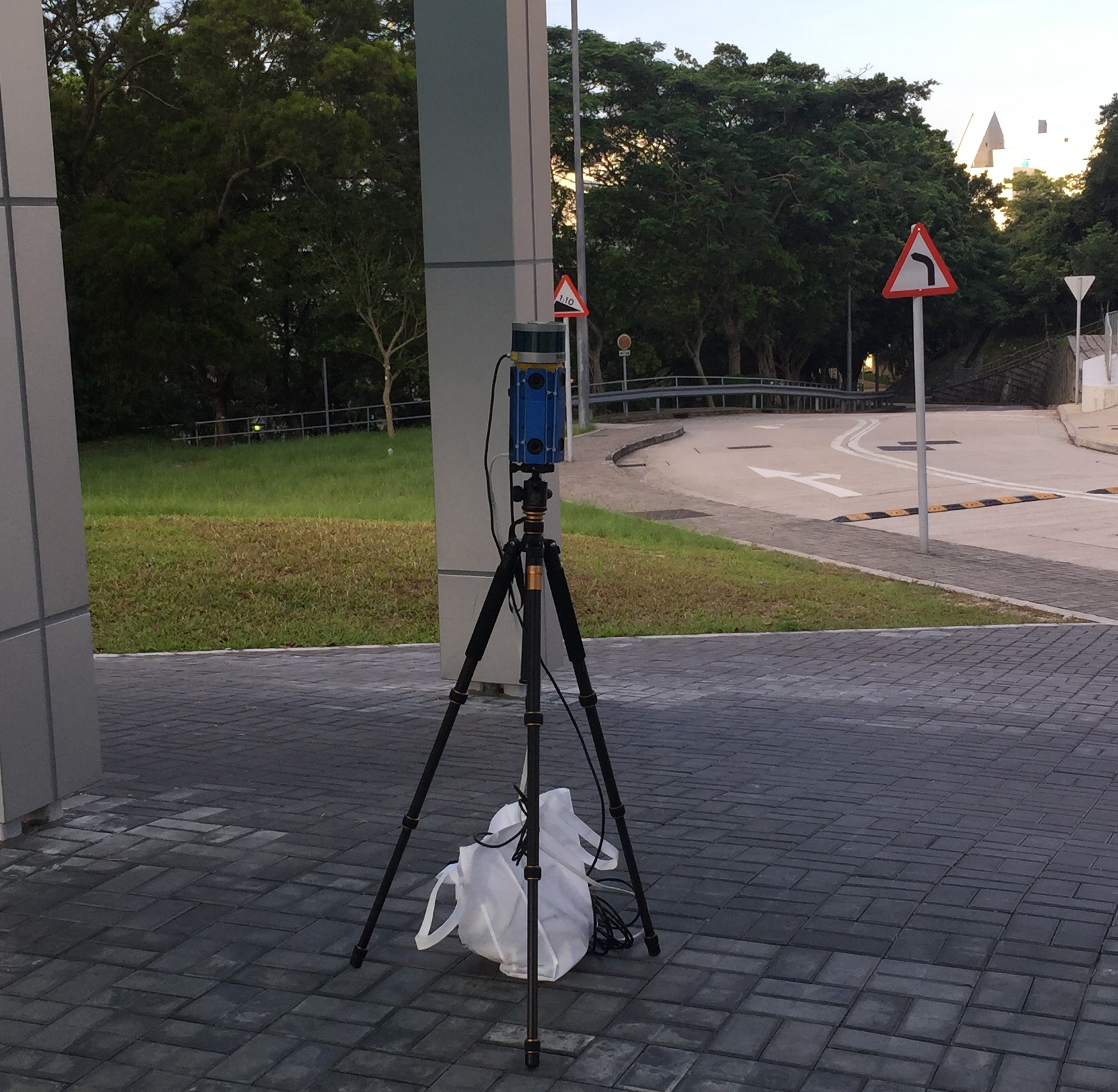}
 	\caption{The equipment used to collect our dataset.  We use an omni stereo camera and a VLP-16 Velodyne LiDAR tied together and placed on a tripod.}
 	\label{fig:equipment}
 \end{figure}

\begin{figure*}
 	\centering
 	\includegraphics[width=0.245\textwidth, height=0.2\textwidth]{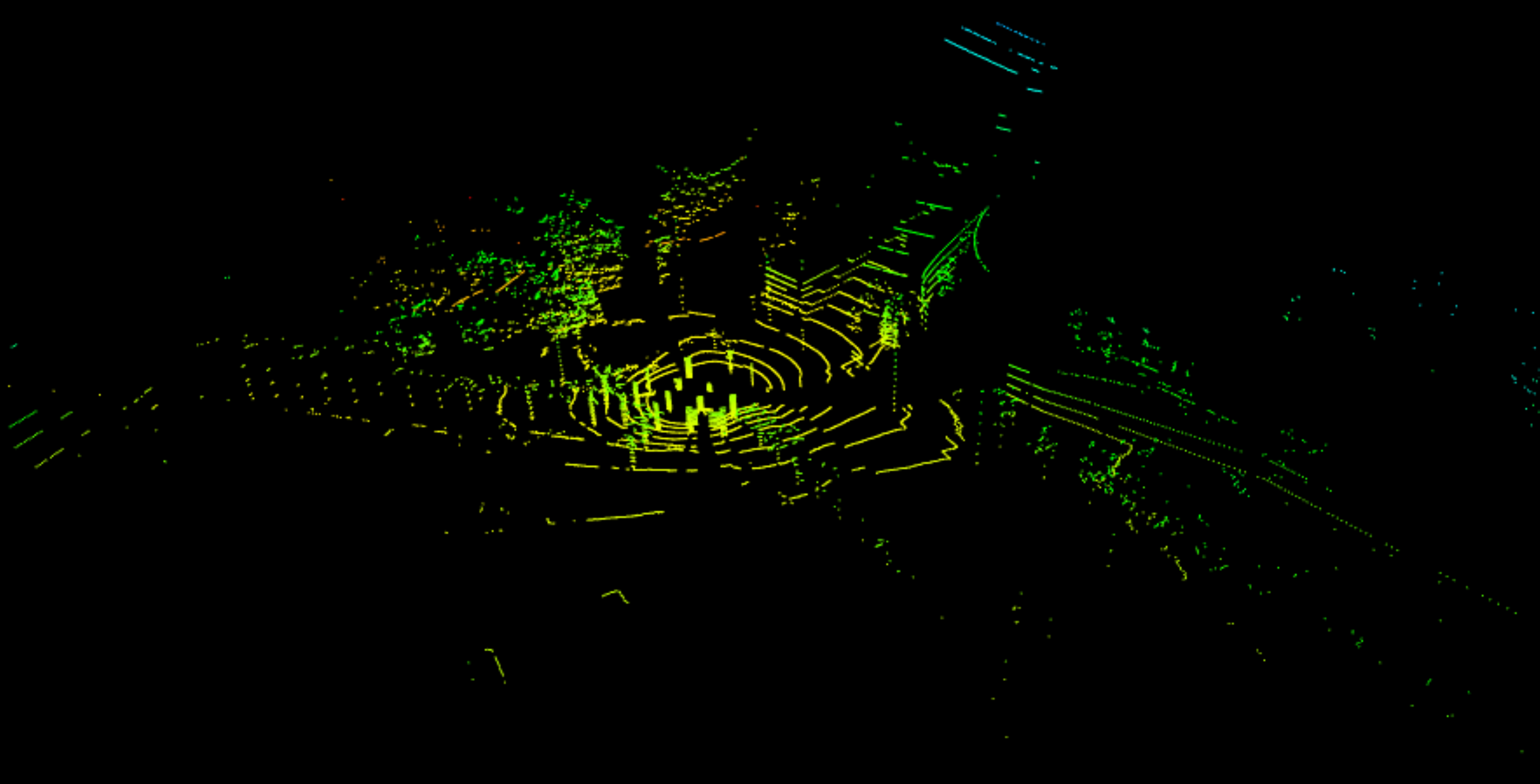}
 	\includegraphics[width=0.245\textwidth, height=0.2\textwidth]{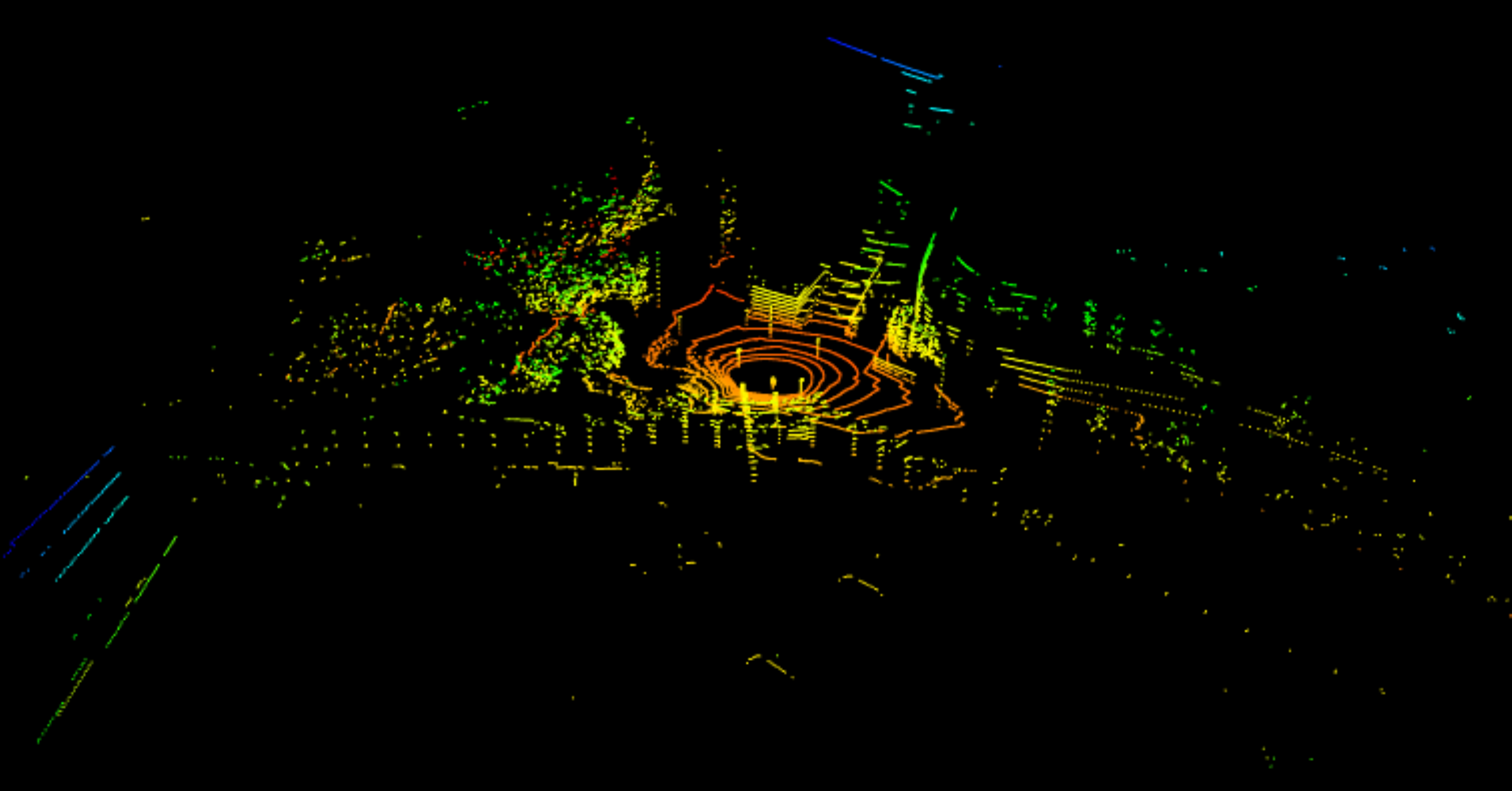}
 	\includegraphics[width=0.245\textwidth, height=0.2\textwidth]{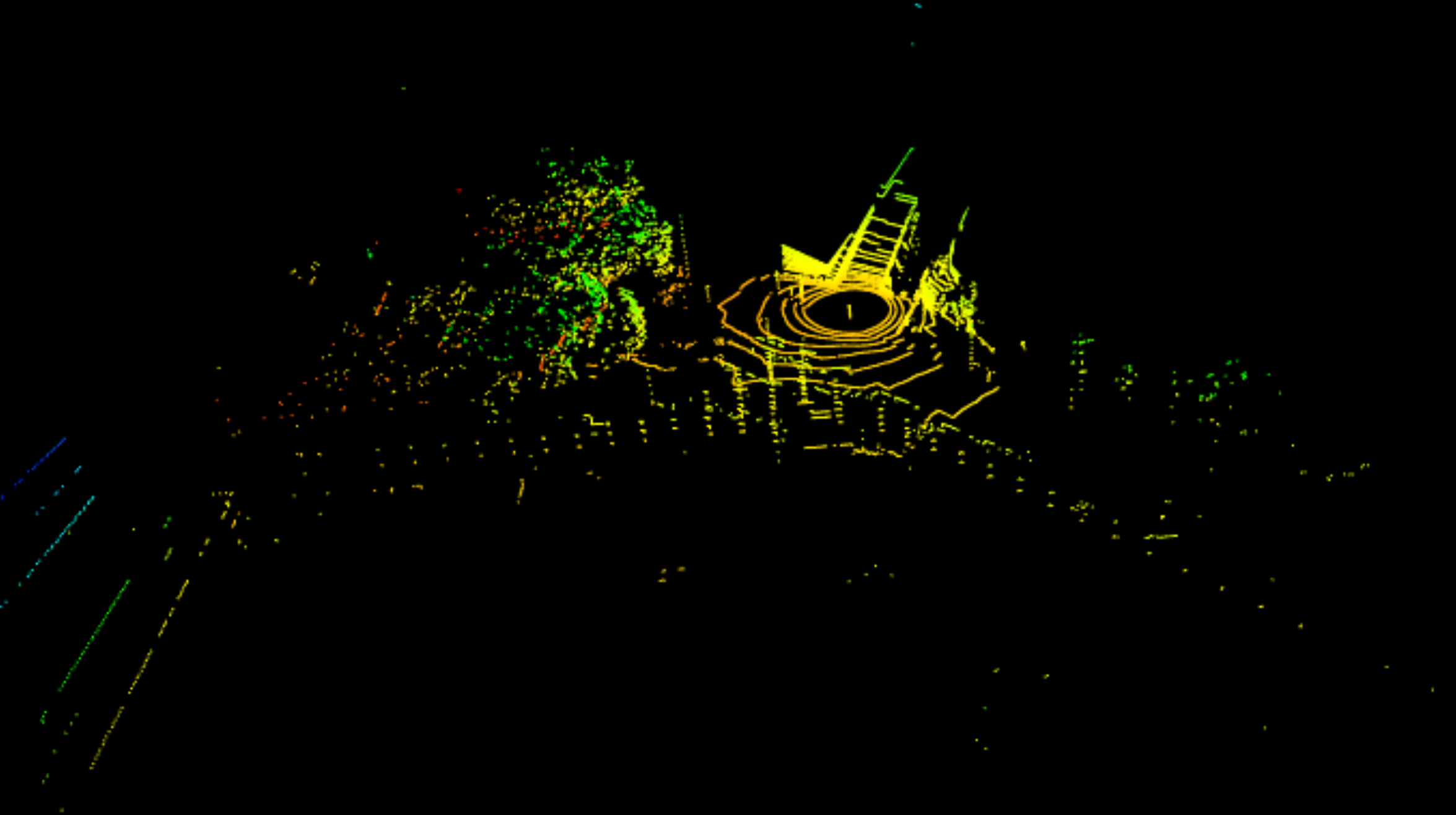}
 	\includegraphics[width=0.245\textwidth, height=0.2\textwidth]{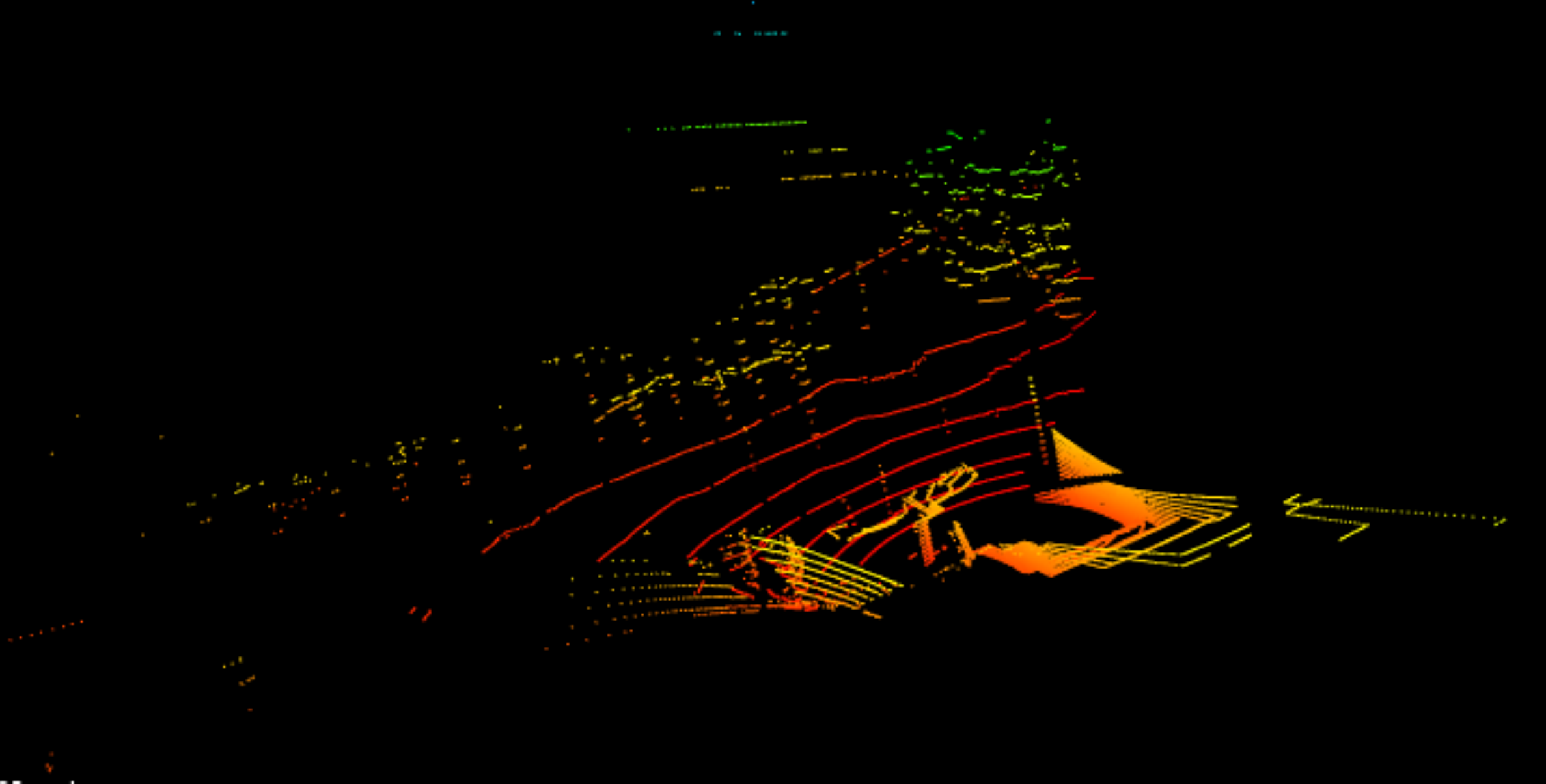}
 	\includegraphics[width=0.245\textwidth, height=0.1\textwidth]{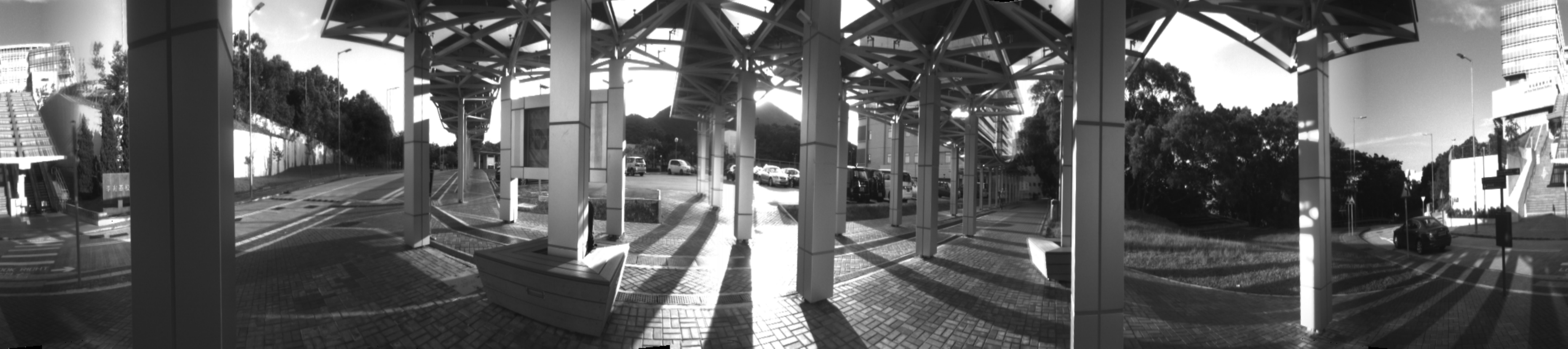}
 	\includegraphics[width=0.245\textwidth, height=0.1\textwidth]{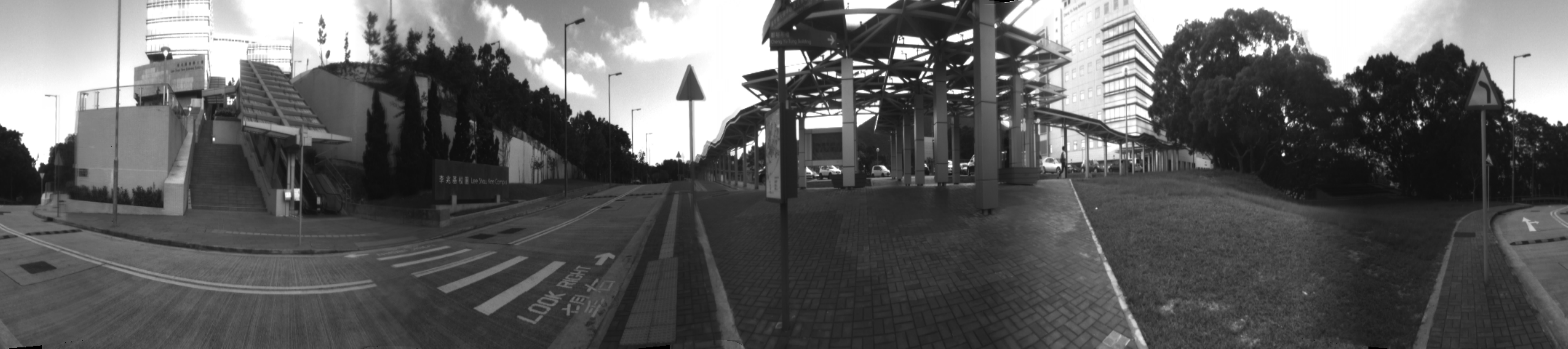}
 	\includegraphics[width=0.245\textwidth, height=0.1\textwidth]{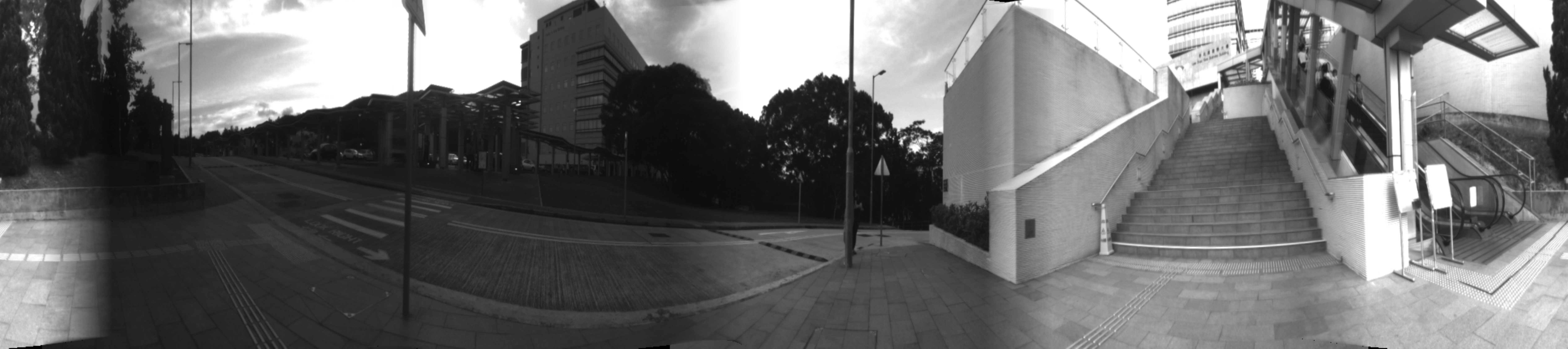}
 	\includegraphics[width=0.245\textwidth, height=0.1\textwidth]{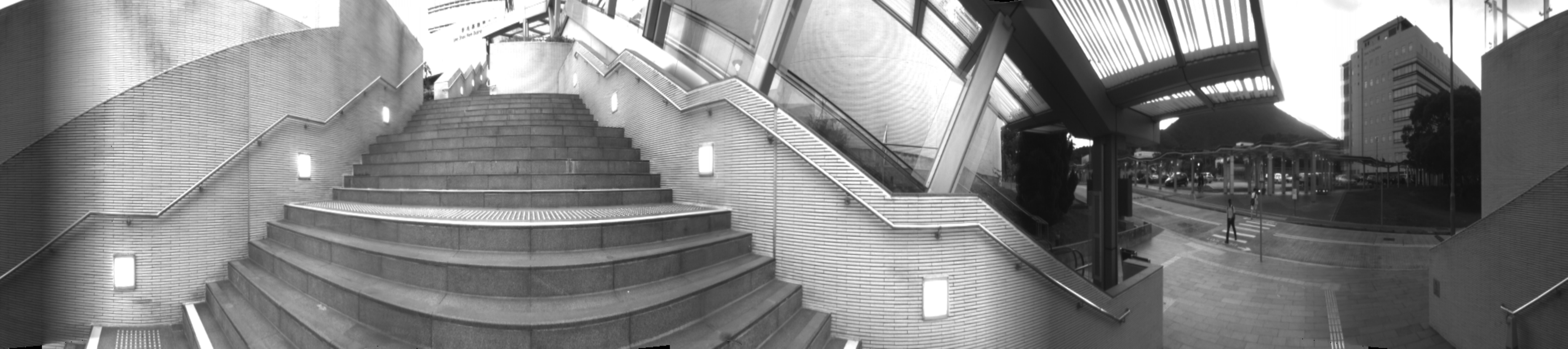}
 	\includegraphics[width=0.245\textwidth, height=0.1\textwidth]{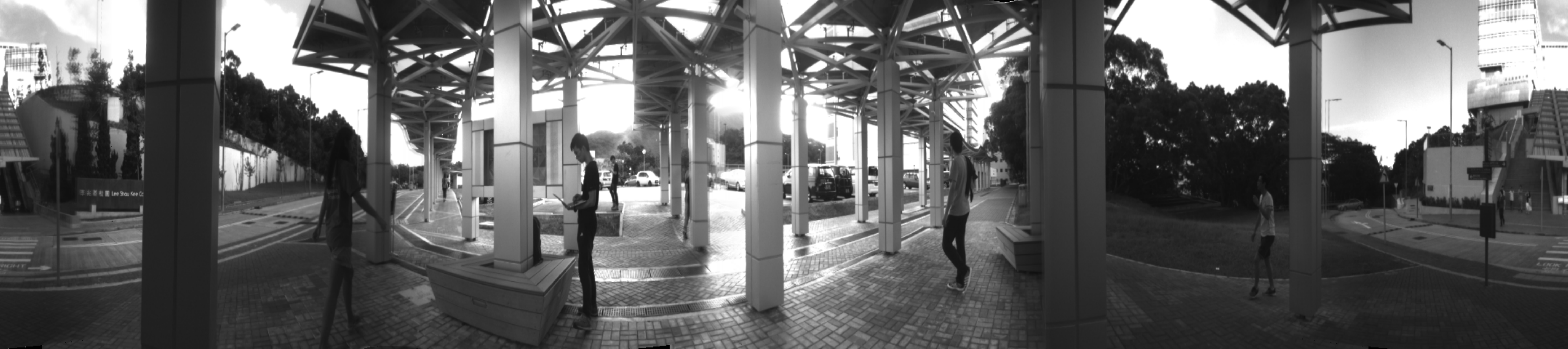}
 	\includegraphics[width=0.245\textwidth, height=0.1\textwidth]{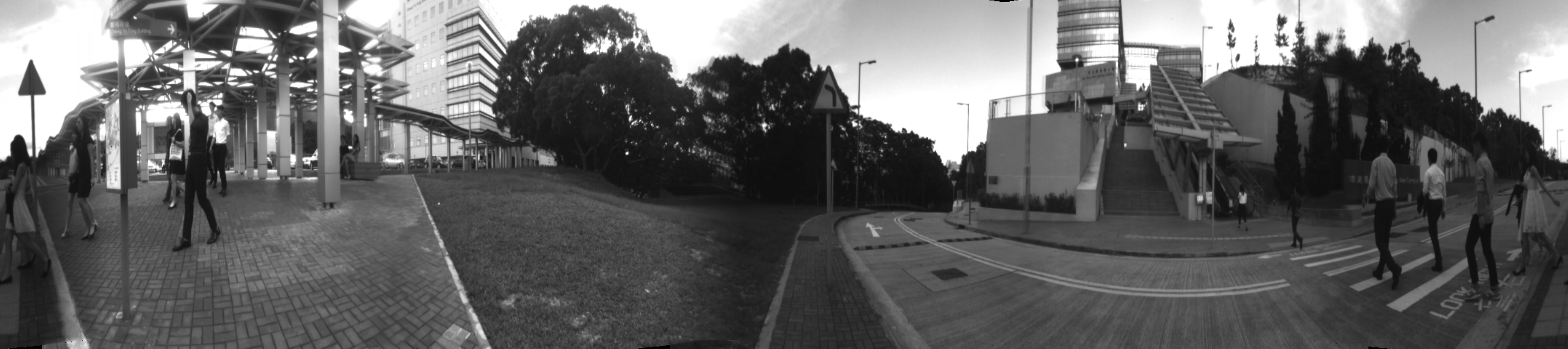}
 	\includegraphics[width=0.245\textwidth, height=0.1\textwidth]{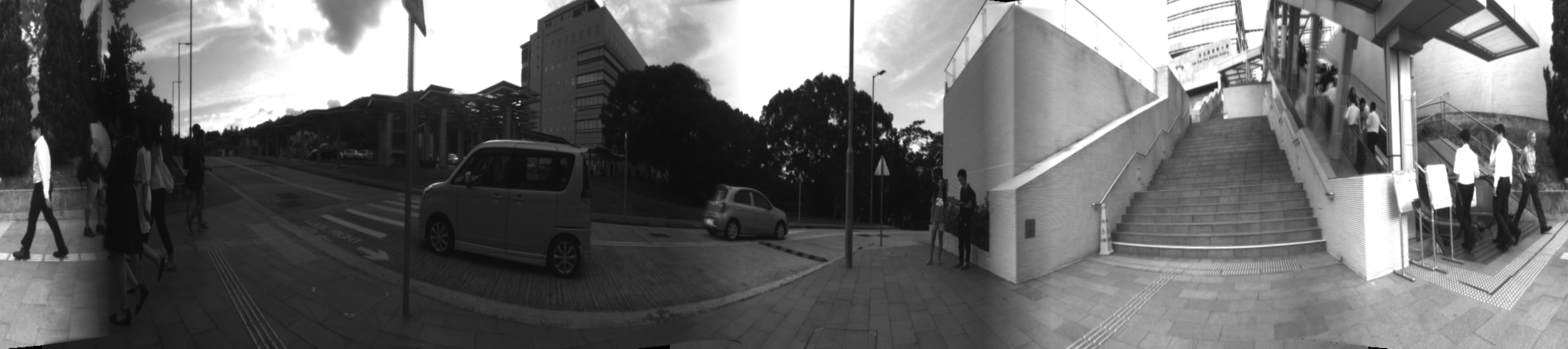}
 	\includegraphics[width=0.245\textwidth, height=0.1\textwidth]{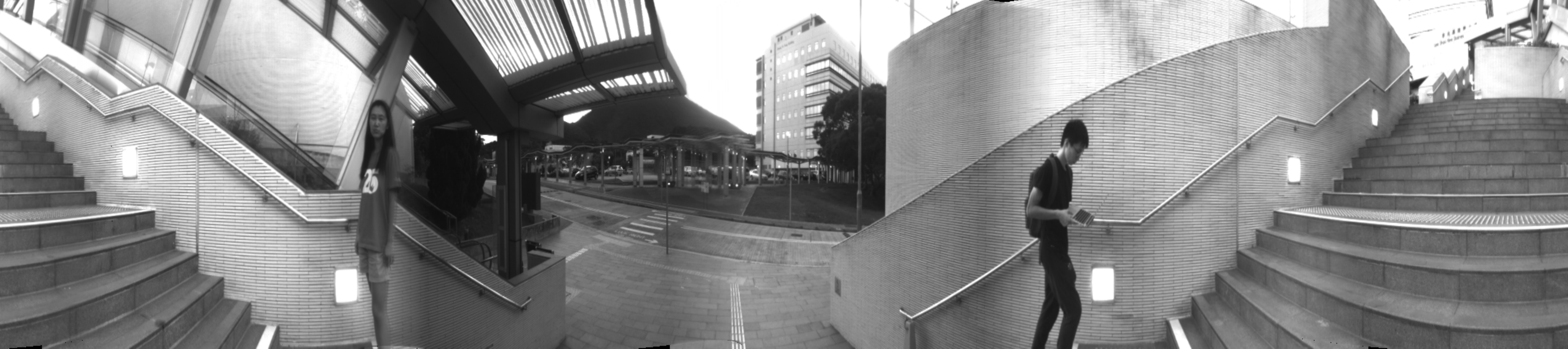}
 	
 	\includegraphics[width=0.245\textwidth, height=0.2\textwidth]{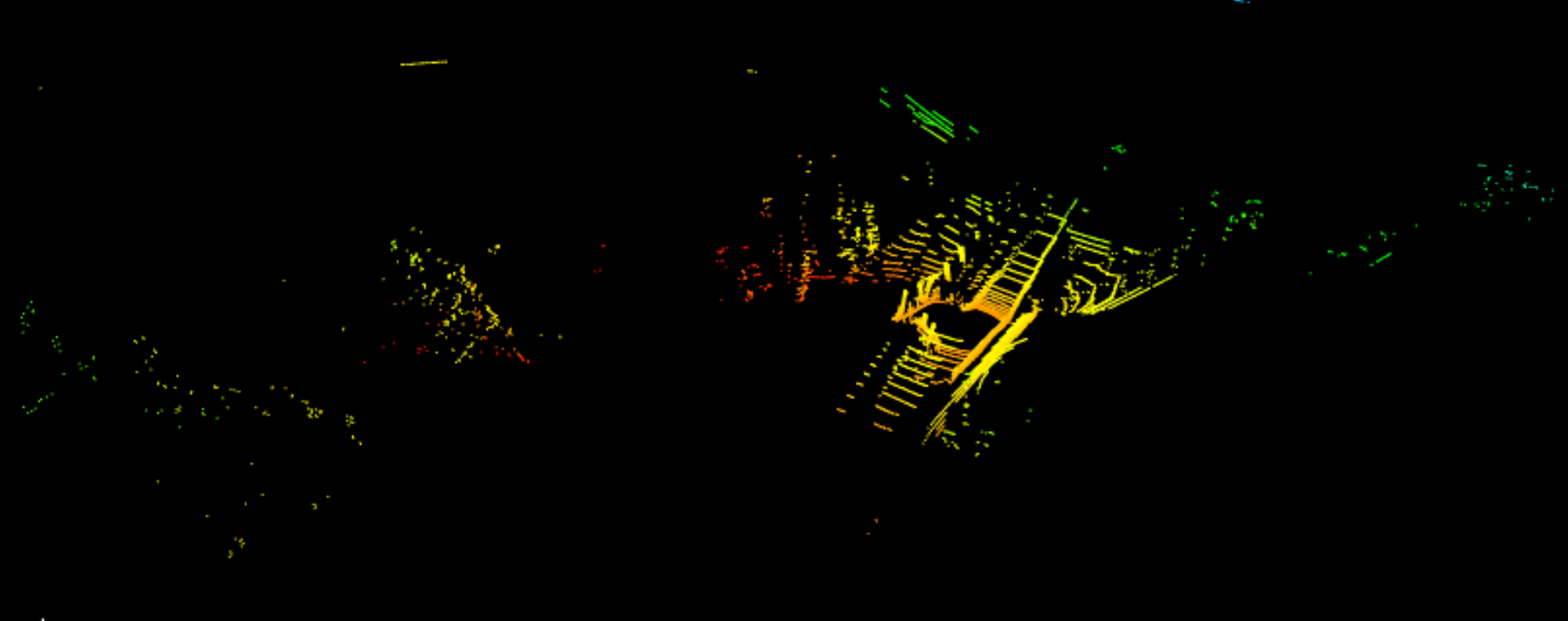}
 	\includegraphics[width=0.245\textwidth, height=0.2\textwidth]{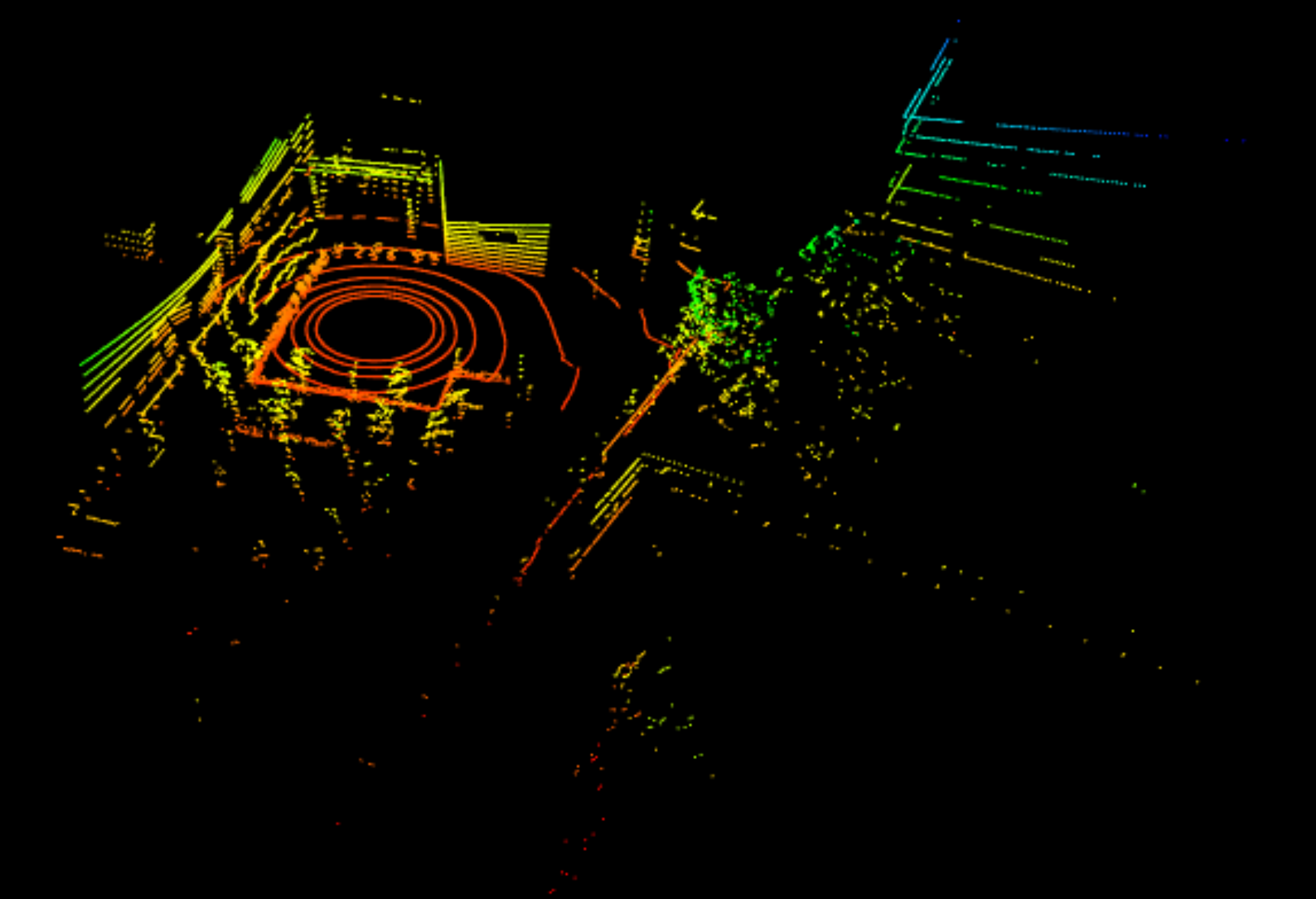}
 	\includegraphics[width=0.245\textwidth, height=0.2\textwidth]{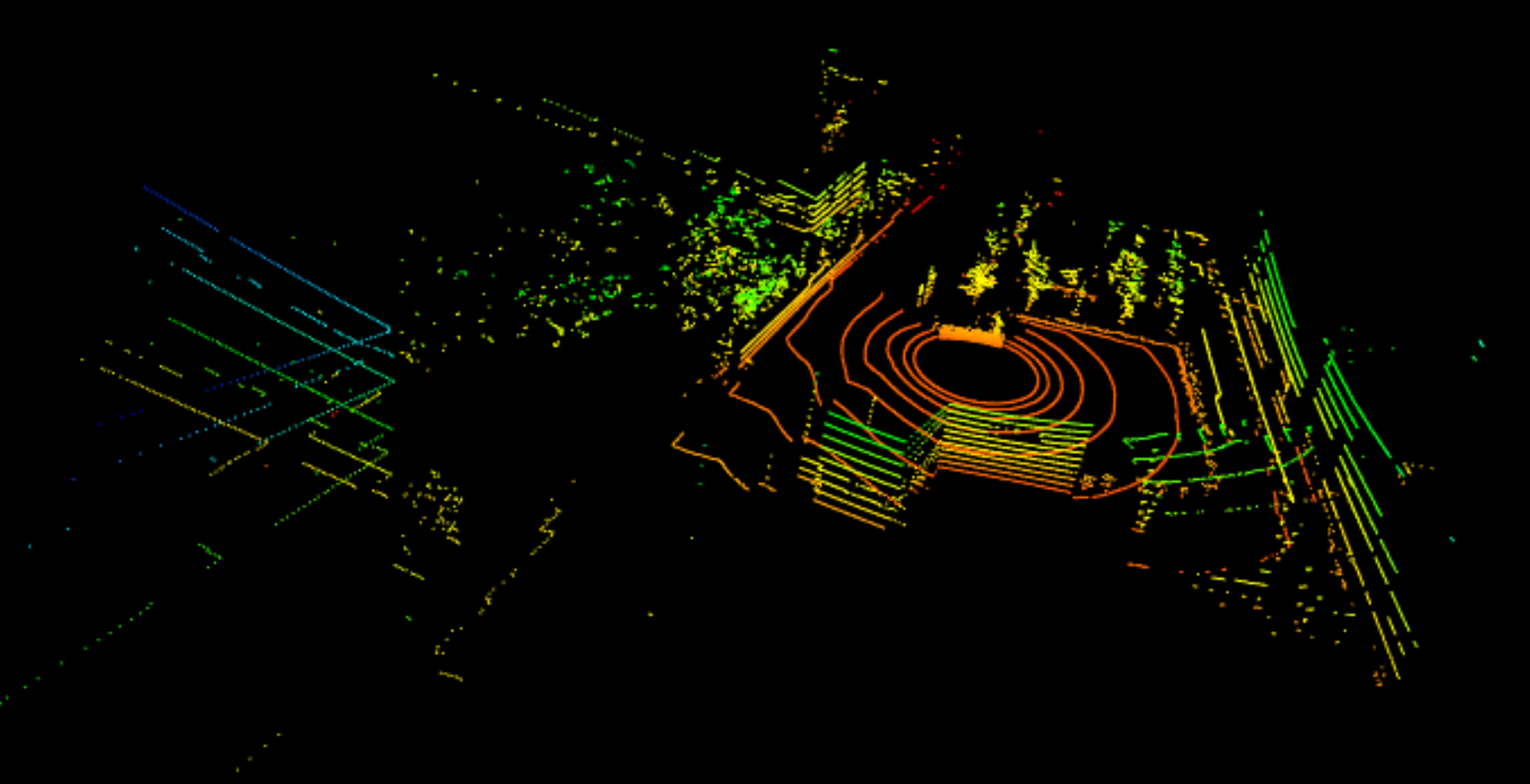} \\
 	\includegraphics[width=0.245\textwidth, height=0.1\textwidth]{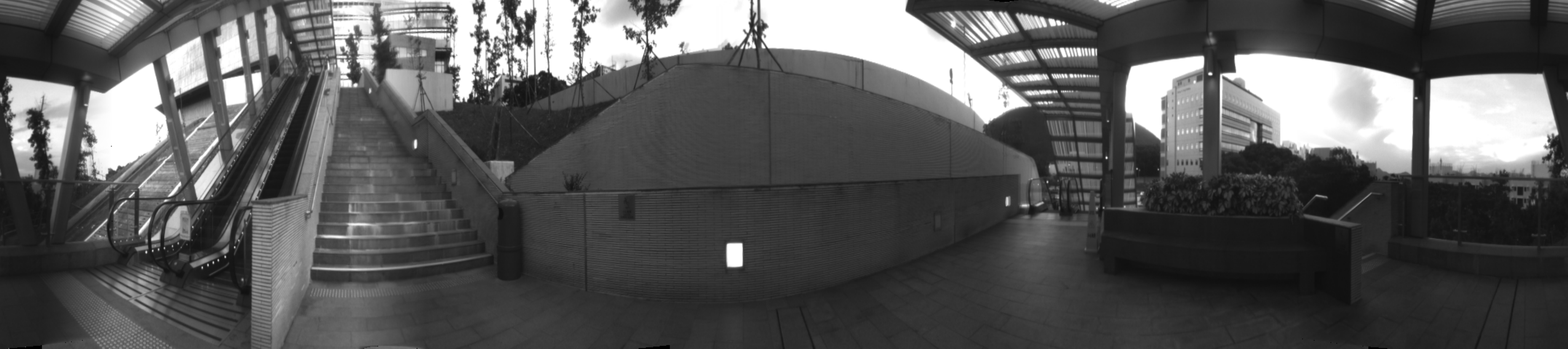}
 	\includegraphics[width=0.245\textwidth, height=0.1\textwidth]{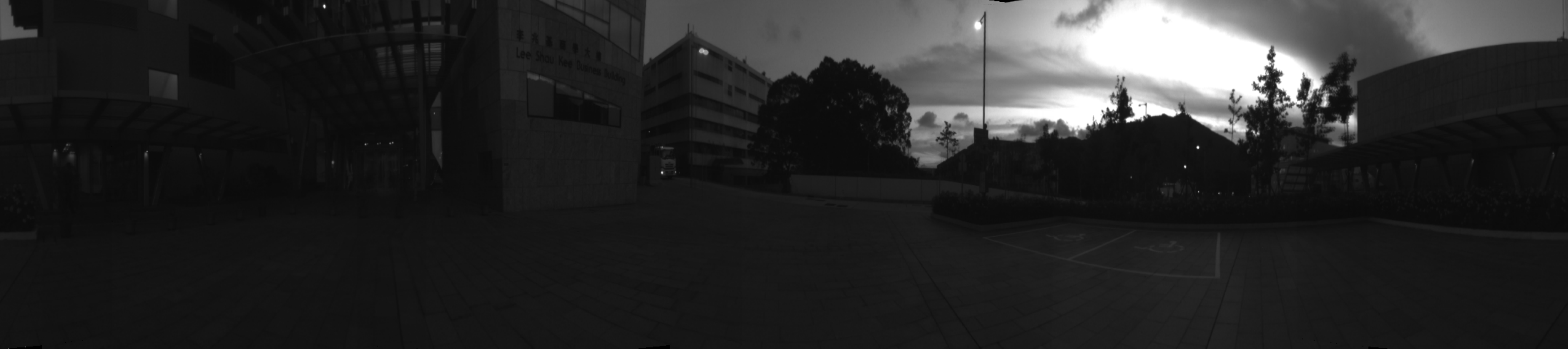}
 	\includegraphics[width=0.245\textwidth, height=0.1\textwidth]{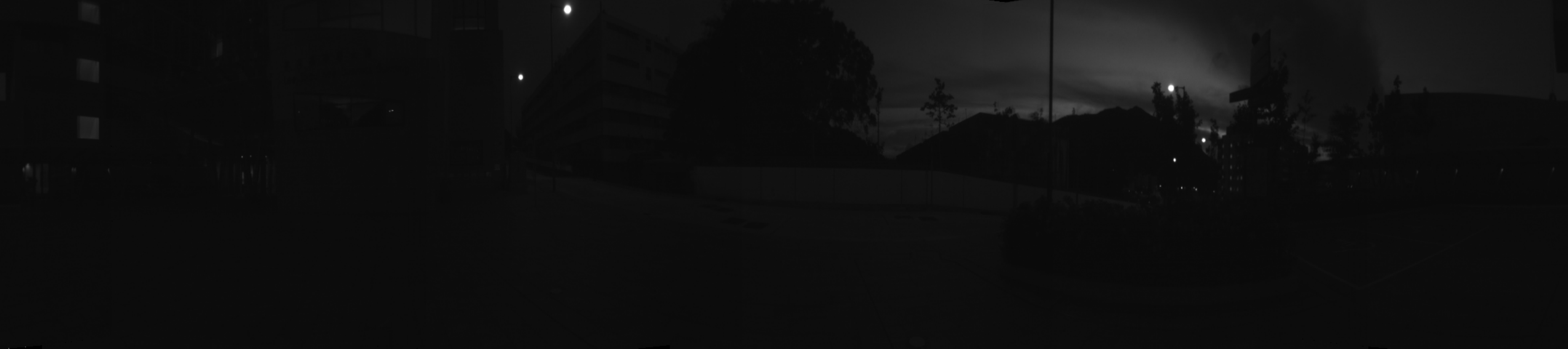} \\
 	\includegraphics[width=0.245\textwidth, height=0.1\textwidth]{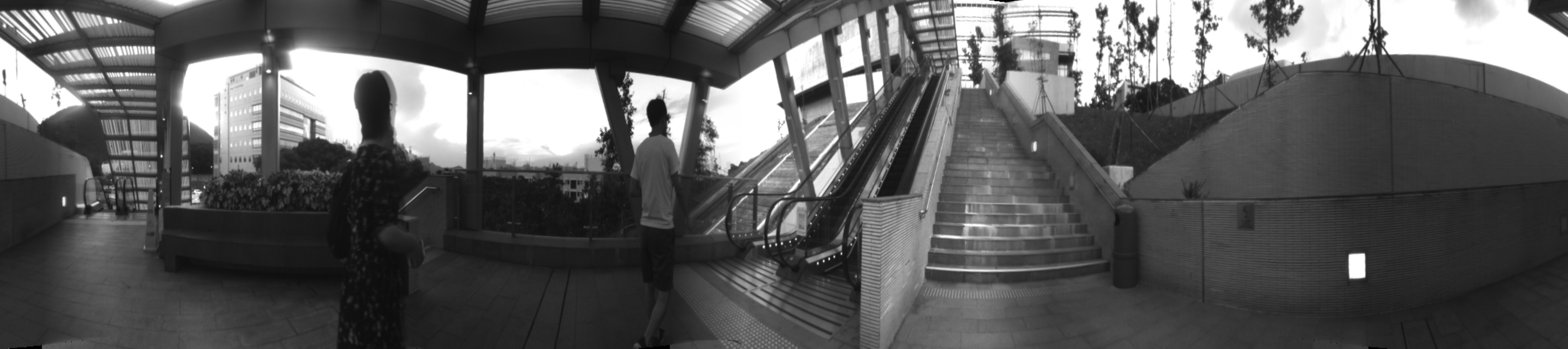}
 	\includegraphics[width=0.245\textwidth, height=0.1\textwidth]{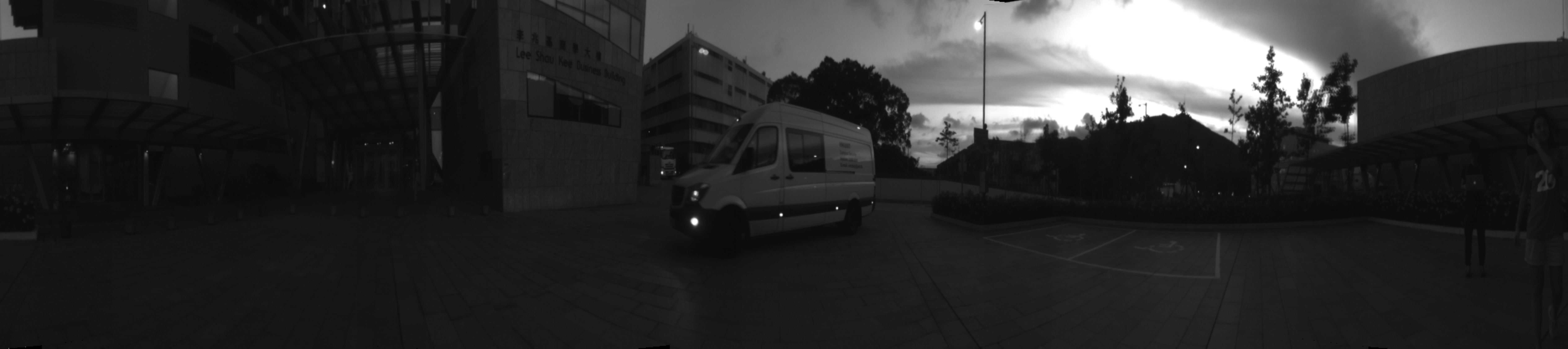}
 	\includegraphics[width=0.245\textwidth, height=0.1\textwidth]{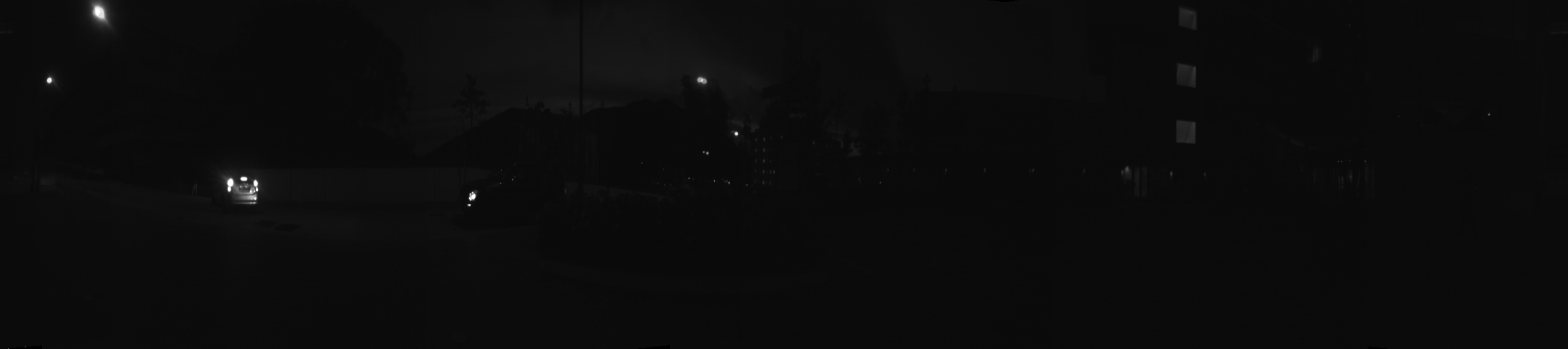}
 	\caption{Examples of the 7 locations in our HKUST dataset.  The top images show the point cloud scans, the middle images show the grayscale images taken at the same location and at the same time, and the bottom images show the same location with unrelated objects passing by.}
 	\label{fig:data_samples}
 \end{figure*}

 We capture our dataset using an omni stereo camera and a VLP-16 Velodyne LiDAR tied together and placed on a tripod, as shown in Figure~\ref{fig:equipment}.  Each pair of shots contain a grayscale image and a point cloud.  Both types of data cover a full $360^\circ$ environmental view and they are synchronized within 1 millisecond.  At each location (i.e. within a 1 meter shift), we collect two sets of data: one for rotation invariance testing, and one for robustness to unrelated (moving) objects testing.  When collecting the first set of data, we turn the tripod about $40^\circ$ for each pair of shots, and repeat this process about 10 times at one location, completing a $360^\circ$ circle; while for the collection of the second set of data, the equipment is fixed, and we take a pair of shots when there are people or cars passing by.  This is also repeated about 10 times at each location.  We collect data in 7 different locations on the Hong Kong University of Science and Technology (HKUST) campus under different lightening conditions, gaining 171 pairs of shots in total.  Among them, 77 pairs belong to the rotation invariance testing set and 94 pairs are for unrelated object testing.  Examples from the 7 locations in our HKUST dataset are shown in Figure~\ref{fig:data_samples}. The top images show the point cloud scans, the middle images show the grayscale images taken at the same location and at the same time, and the bottom images show the same location when unrelated objects are passing by. 
 
\section{Experiments}
\label{sec:Experiments}
\begin{figure*}
	\centering
	\includegraphics[width=0.24\textwidth,height=0.2\textwidth]{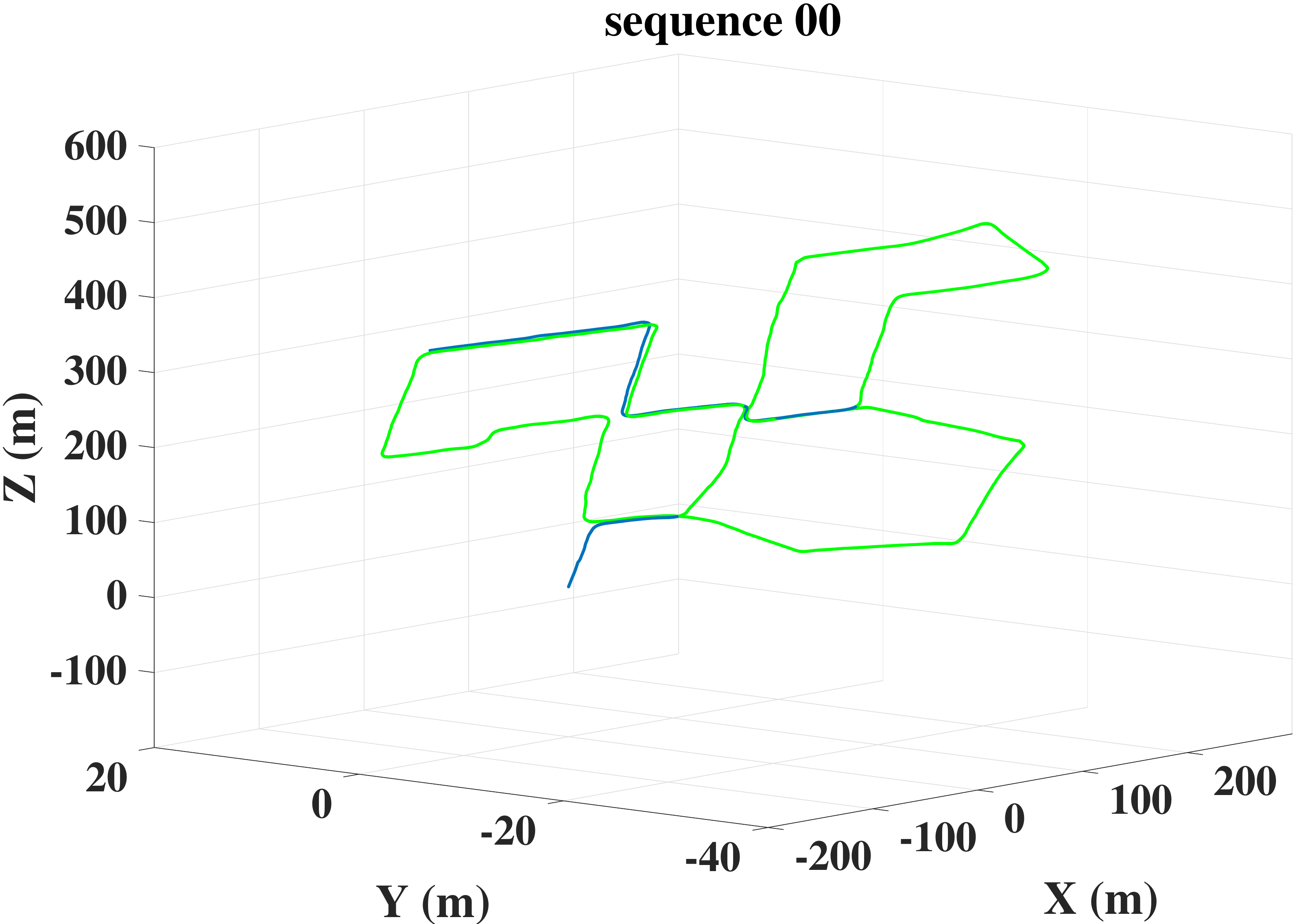}
	\includegraphics[width=0.24\textwidth,height=0.2\textwidth]{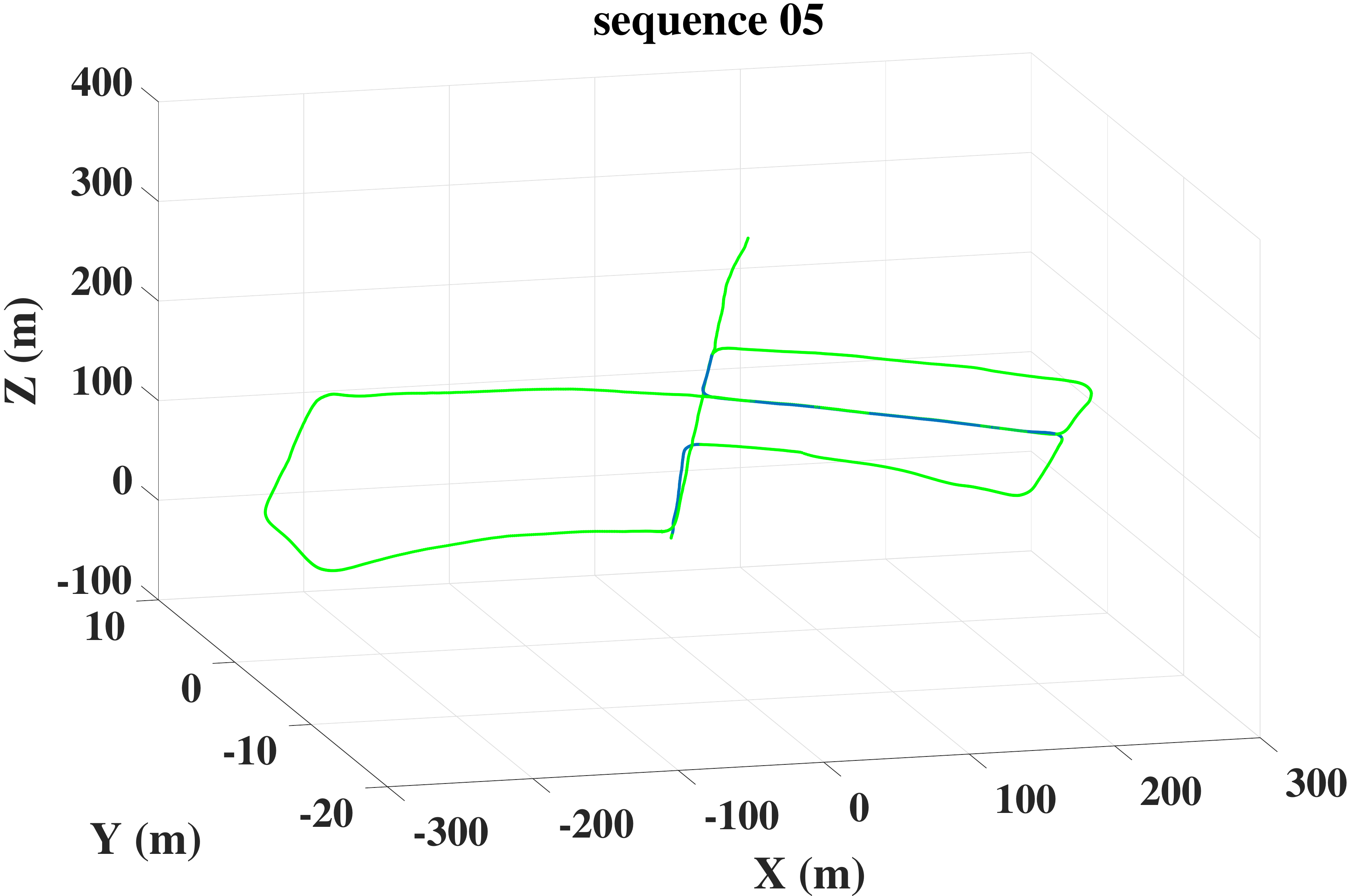}
	\includegraphics[width=0.24\textwidth,height=0.2\textwidth]{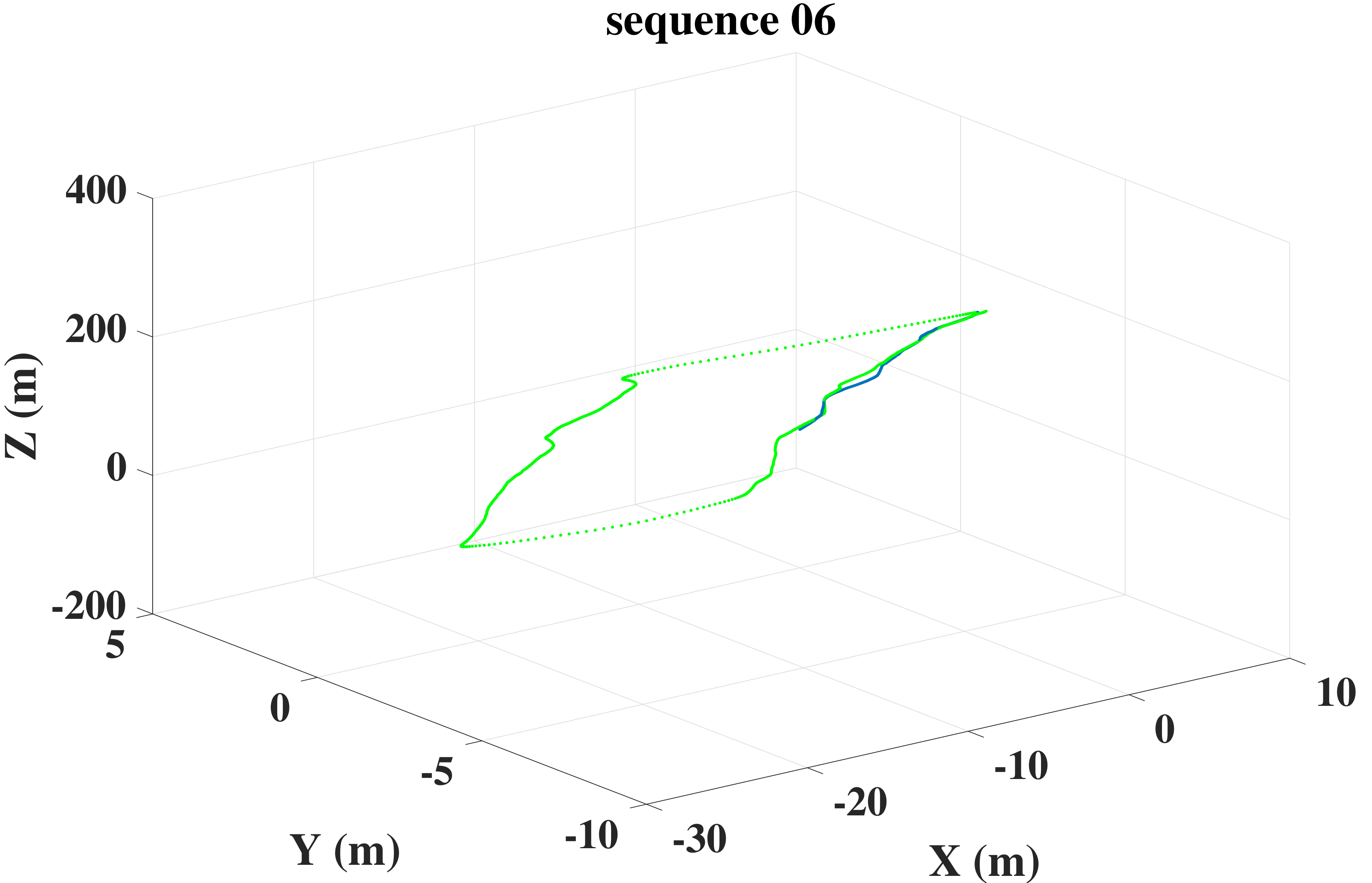}
	\includegraphics[width=0.24\textwidth,height=0.2\textwidth]{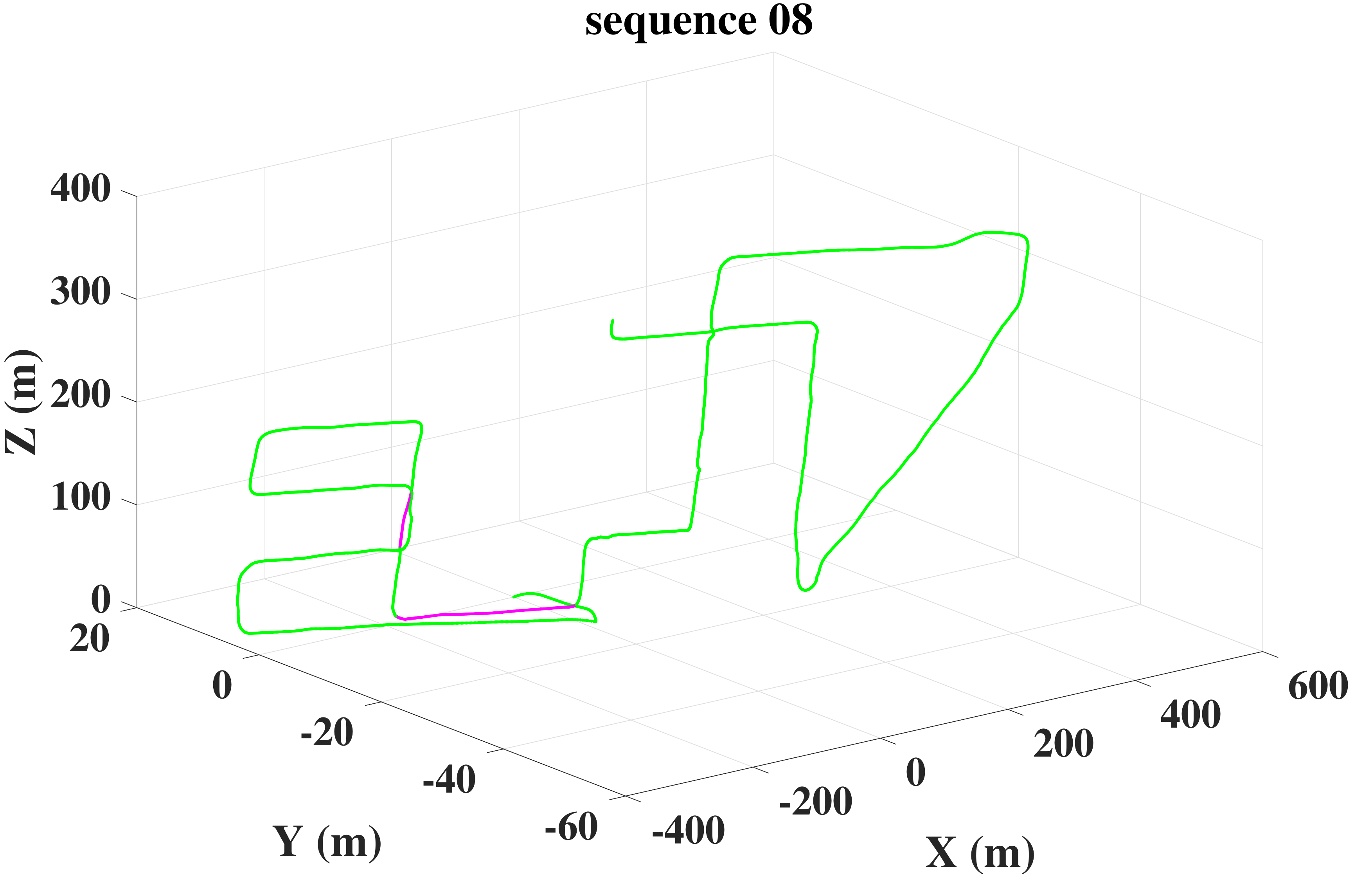}
	\caption{The manually selected revisited paths in each sequence are shown in this figure. The unidirectional loop closure trails are blue and the bidirectional loop closure trails are pink.  The precise blue/pink scans' indexes are listed in Table~\ref{tab:index}. All the green scans are queries.}
	\label{fig:show_path}
\end{figure*}

We thoroughly test our proposed system in this section.  The performance of features extracted from different CNNs and different layers are shown first, then the effectiveness of each module in our system is tested using the best performing layers.  We test our system mainly on the KITTI dataset \cite{kitti}, which offers the ground truth locations for 11 sequences numbered from 00 to 10.  Among the 11 sequences, sequences 01, 03, 04, 07, 09 and 10 do not contain a significant (i.e. less than 100 scans) revisited path; sequences 00, 05 and 06 only contain paths revisited from the same direction i.e. unidirectional loop closures; sequence 08 only contains paths revisited from the opposite directions i.e. bidirectional loop closures; and sequence 02 has revisited paths from both directions.  In order to clearly demonstrate the performance of our system, we use sequences 00, 05, 06 and 08.  We classify our place recognition tasks into 3 classes according to their difficulty.  In the easiest task, the stored scans are uniformly sampled from the sequences, so most of the true matches are only different from the queries by a small location shift and rotation.  The tasks with median difficulty use one trail of the manually-selected unidirectional loop closure paths as stored scans, so most of the queries do not have a true match and the existing matches differ from the queries by a large location shift and rotation.  The difficult tasks are to recognize the revisited places in bidirectional loop closures.  Many works are tested on the easy tasks, the median difficulty ones are more practical, and few works can handle the difficult tasks. (Notice that in order to recognize revisited places under a bidirectional loop closure situation, the input sensor must cover a full $360^\circ$ degree environmental view.)  We found that for unidirectional loop closures, only using the alignment that satisfies (\ref{equ:cond1}) is sufficient to achieve good performance.  We then show in Subsec.~\ref{subsec:bidirectional} that bidirectional loop closures require considering both (\ref{equ:cond1}) and (\ref{equ:cond2}), then we pick the one with higher matching score.  The revisited paths in each sequence are manually selected and are shown in figure~\ref{fig:show_path}, and the stored scans' indexes are listed in Table~\ref{tab:index}

\begin{table}[h]
	\begin{center}
		{\small
			\begin{tabular}{|c|c c|}
				\hline
				\multicolumn{1}{|c|}{sequence}    & loop closure type &  stored scans' index  \\
				\hline\hline
				00             &  unidirectional       & 1-200, 3280-3840     \\
				05             &  unidirectional       & 13-150, 530-885           \\
				06             &  unidirectional       & 1-276 \\
				08             &  bidirectional        & 1420-1503, 1620-1840 \\
				\hline
			\end{tabular}
		}
	\end{center}
	\caption{Manually selected stored scans' indexes.}
	\label{tab:index}
\end{table}

We run our C++ code on a desktop with 4 cores @ 3.20 GHz without multi-threading for CPU implementation, and adopt Caffe \cite{Caffe} for CNN feature extraction. We also use GPU version of Caffe \cite{Caffe} on a desktop with 4 cores @ 3.07 GHz with a NVIDIA GTX 980.

\subsection{Performance of features from different layers of different CNNs}
\begin{figure*}
	\centering
	\includegraphics[width=0.32\textwidth,height=0.2\textwidth]{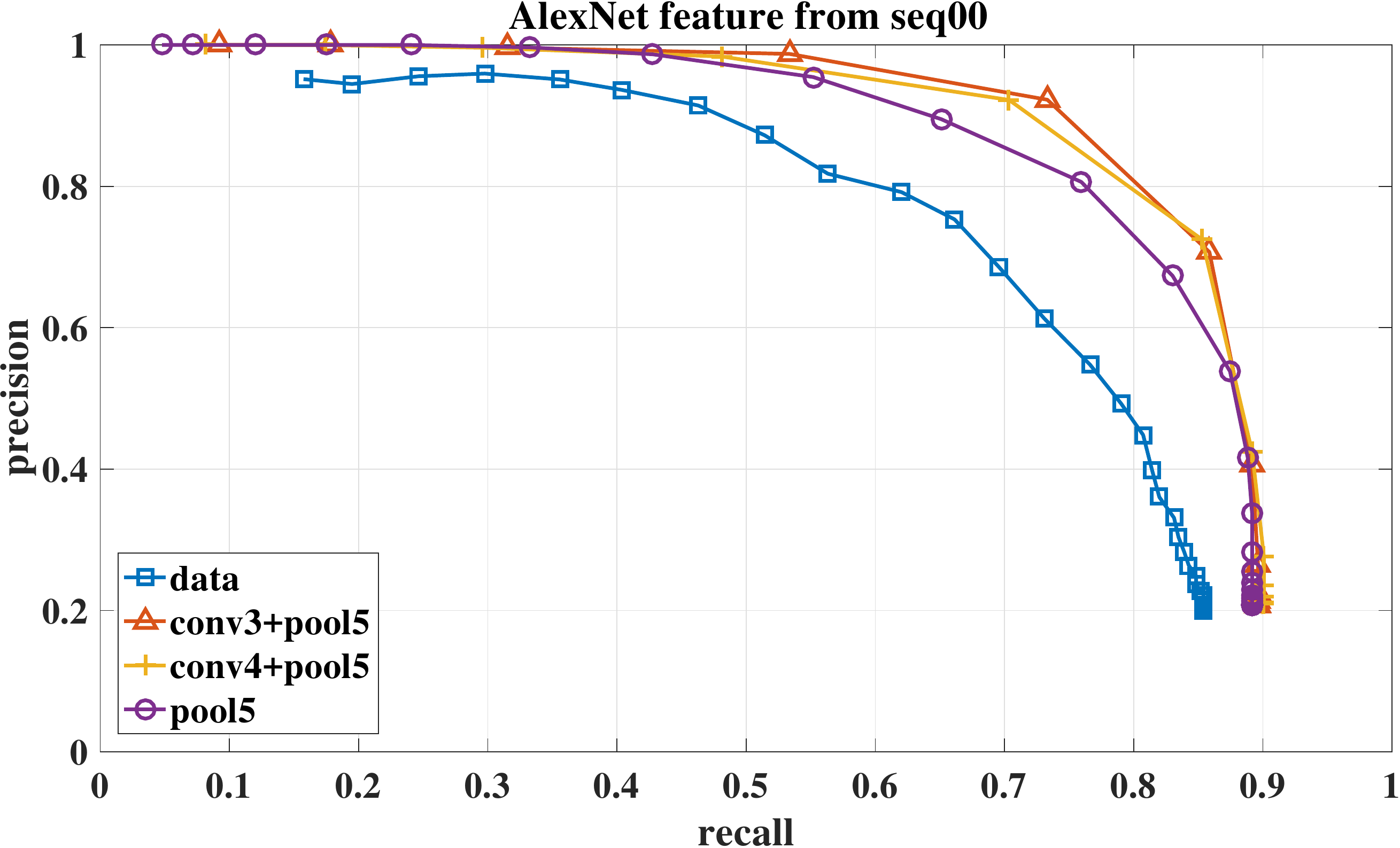}
	\includegraphics[width=0.32\textwidth,height=0.2\textwidth]{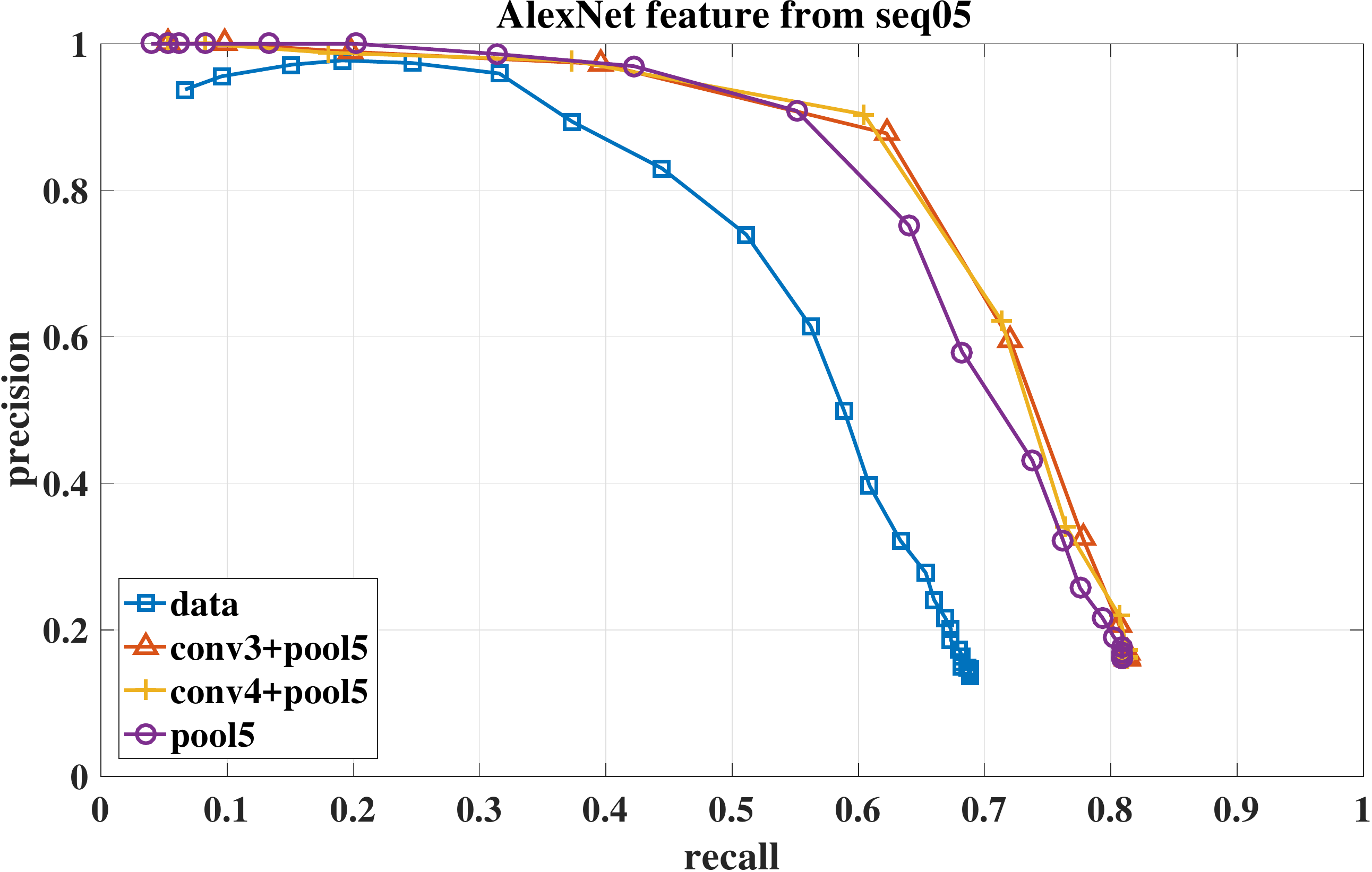}
	\includegraphics[width=0.32\textwidth,height=0.2\textwidth]{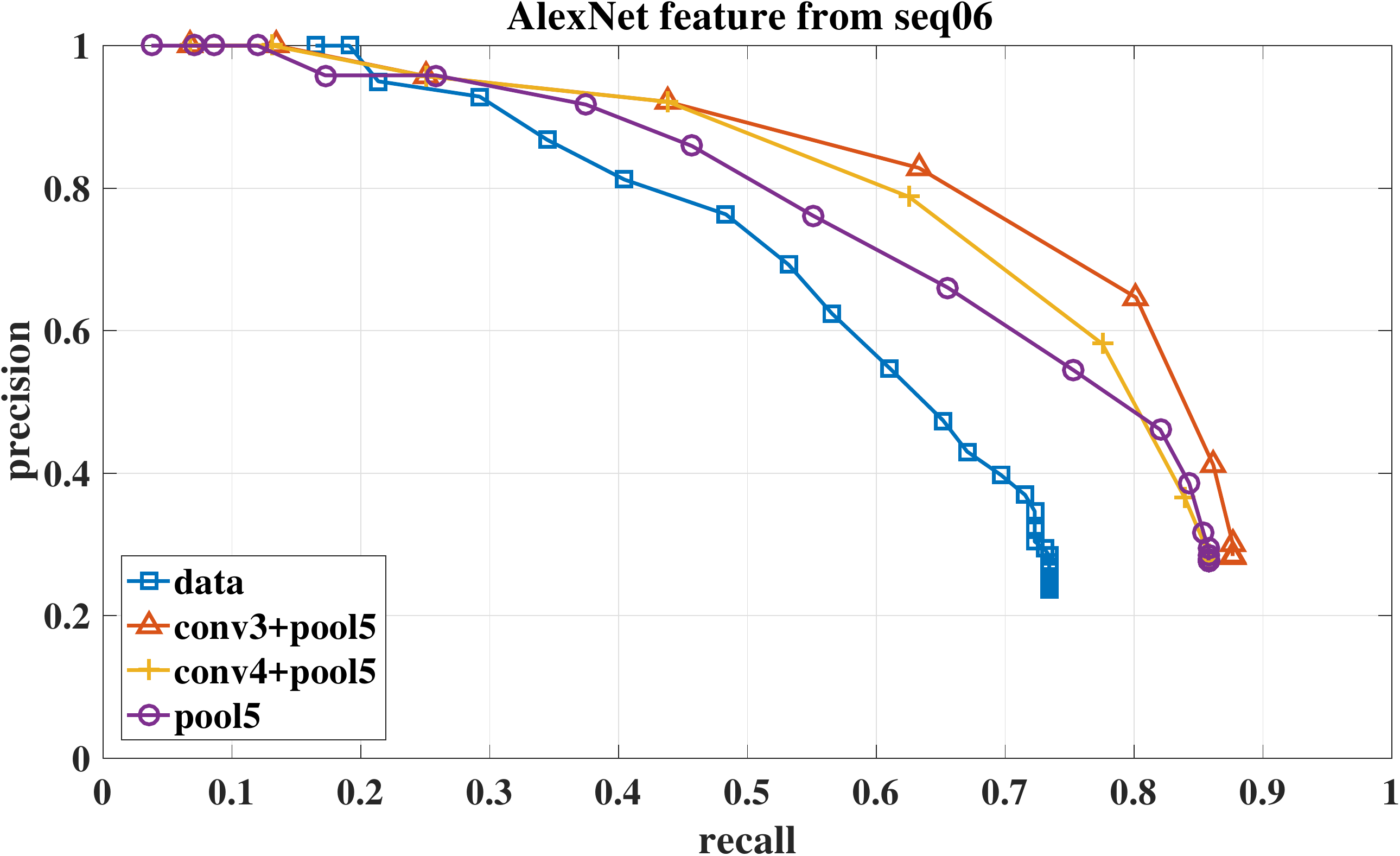}
	\includegraphics[width=0.32\textwidth,height=0.2\textwidth]{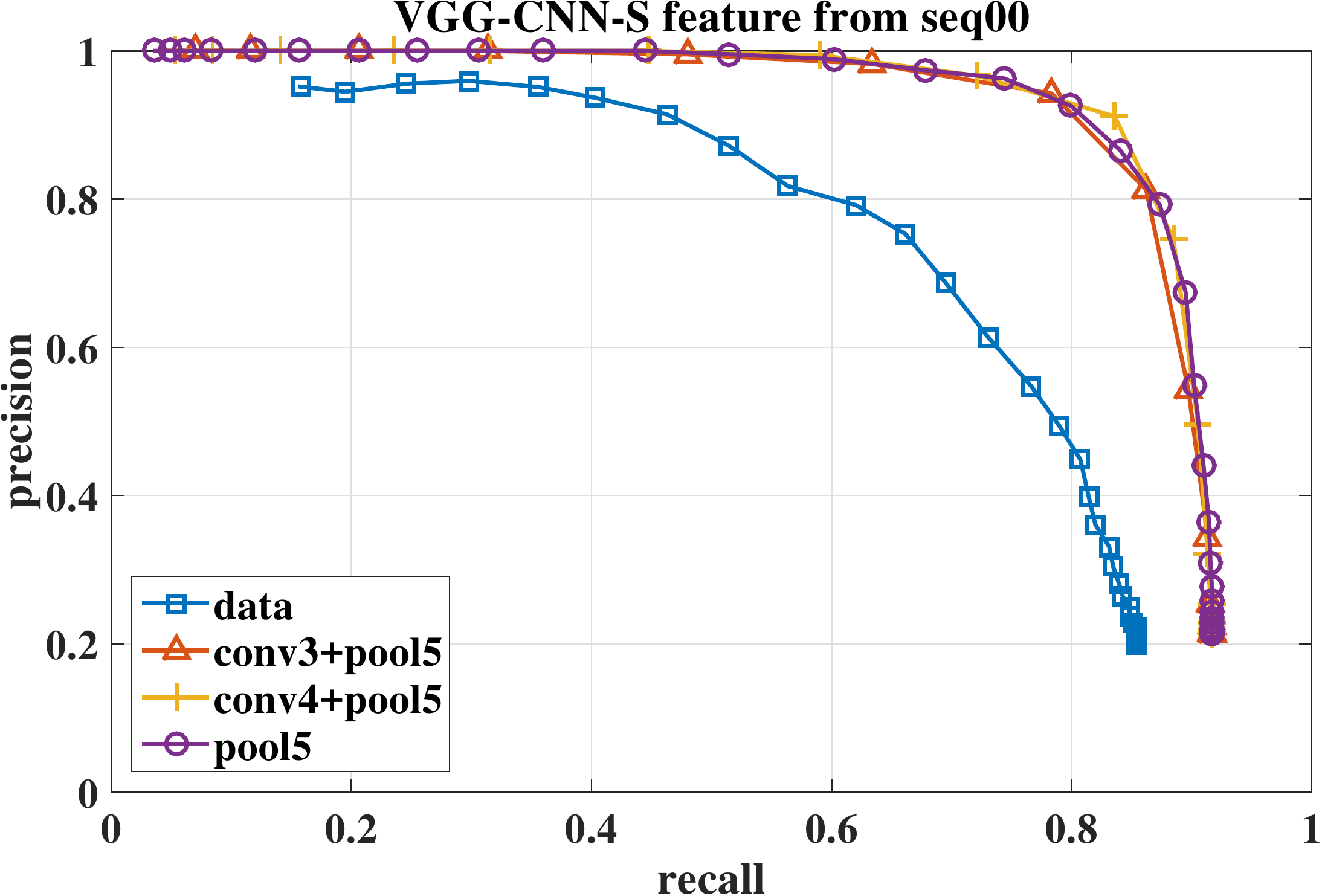}
	\includegraphics[width=0.32\textwidth,height=0.2\textwidth]{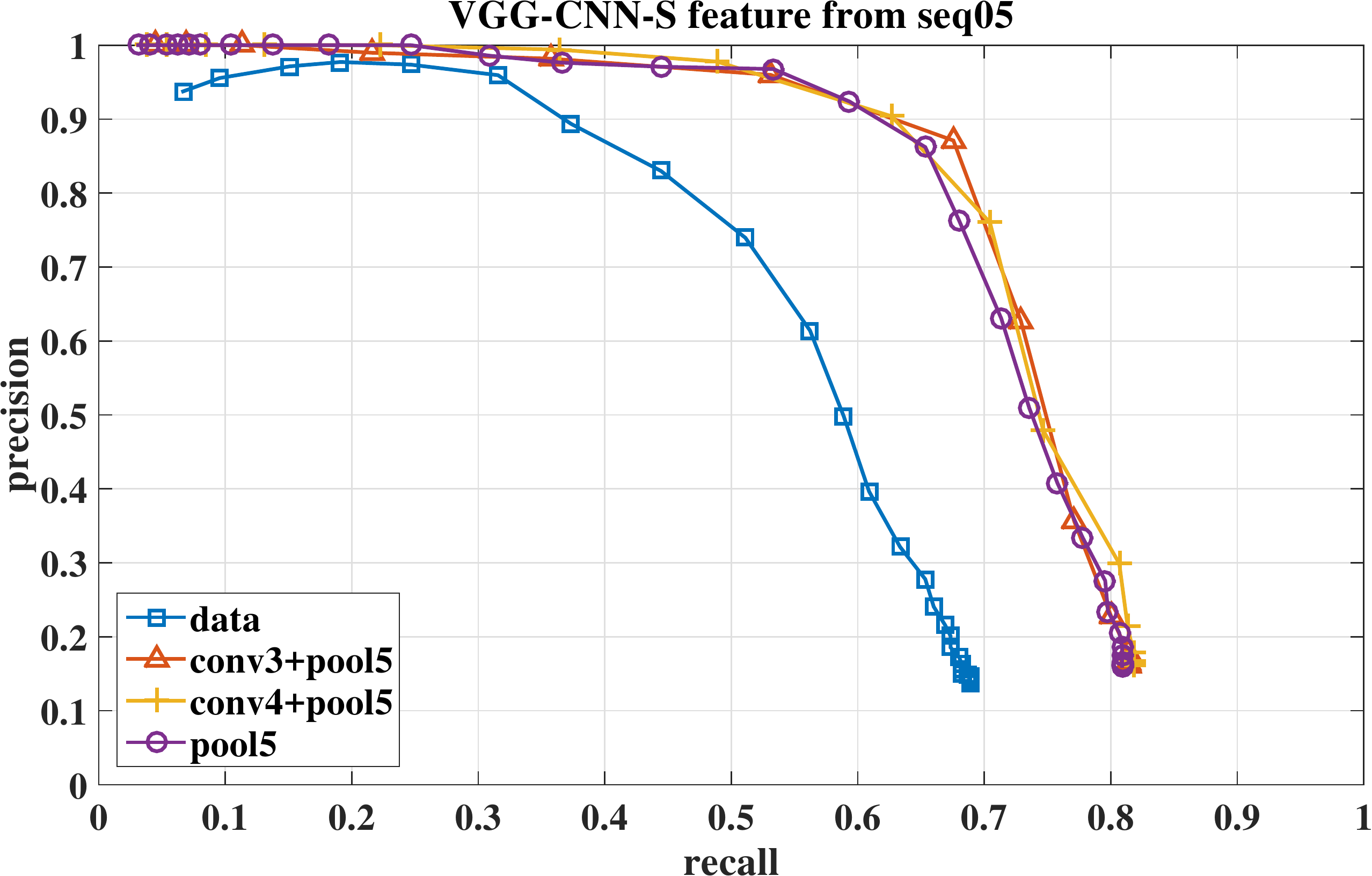}
	\includegraphics[width=0.32\textwidth,height=0.2\textwidth]{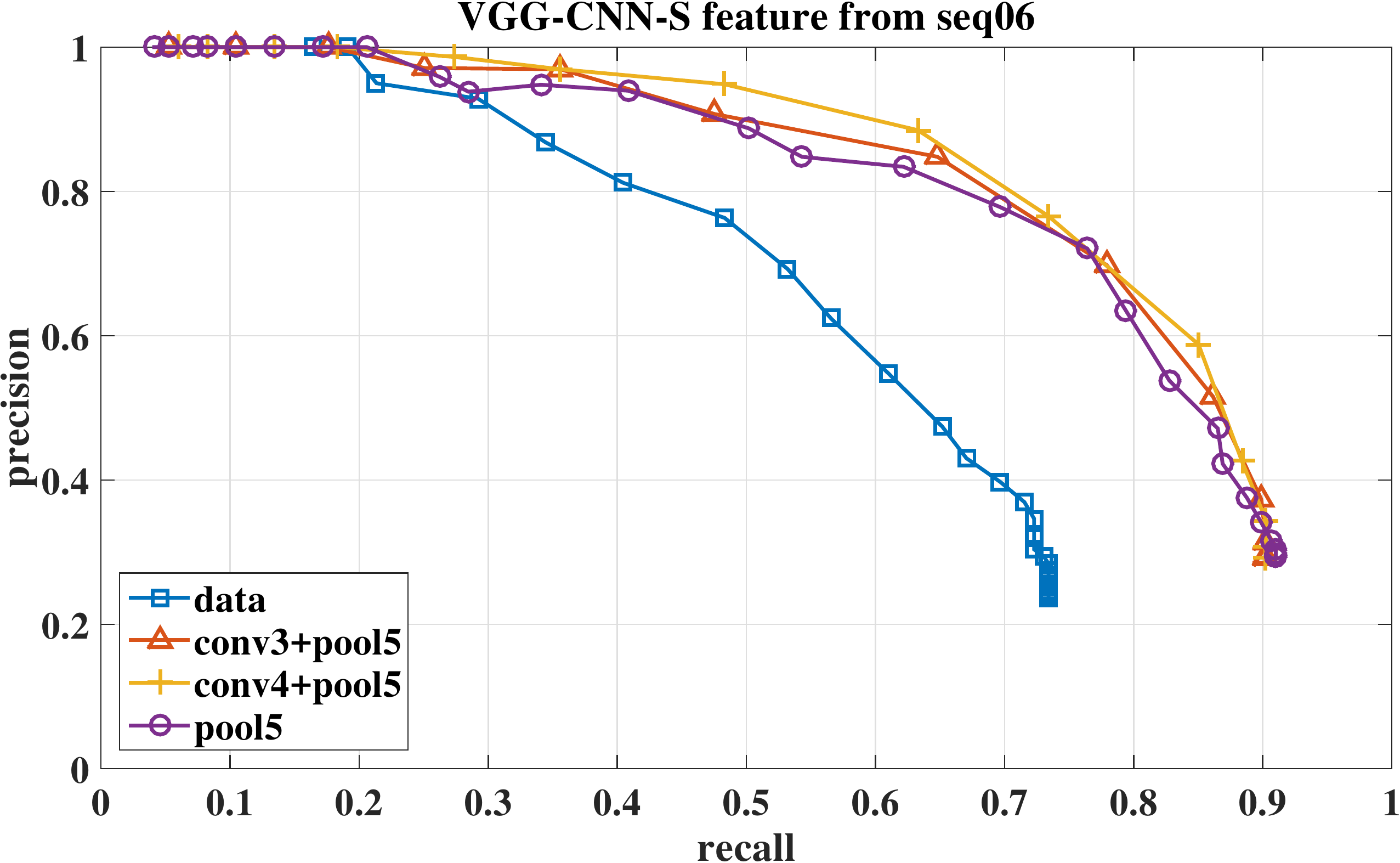}
	\includegraphics[width=0.32\textwidth,height=0.2\textwidth]{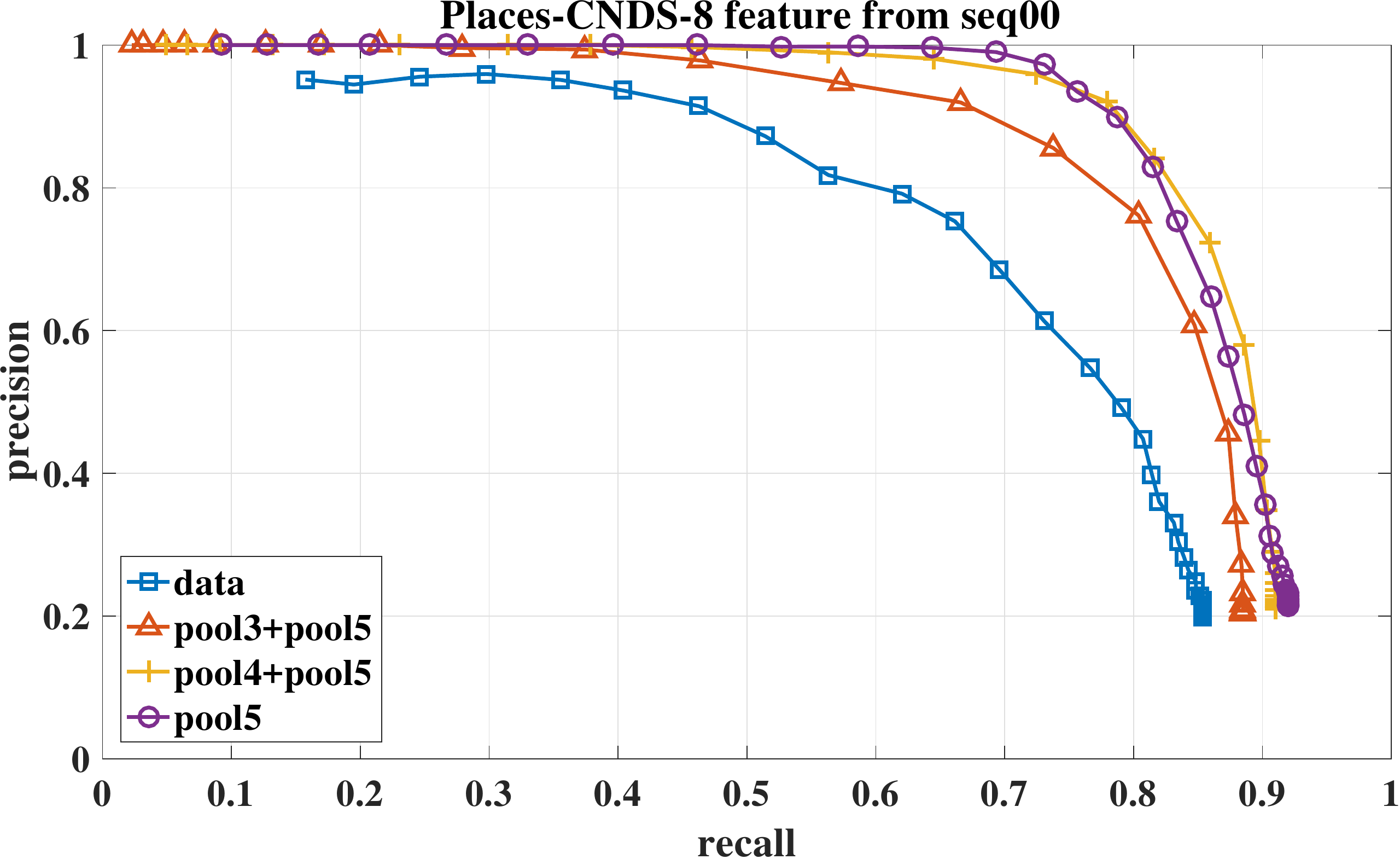}
	\includegraphics[width=0.32\textwidth,height=0.2\textwidth]{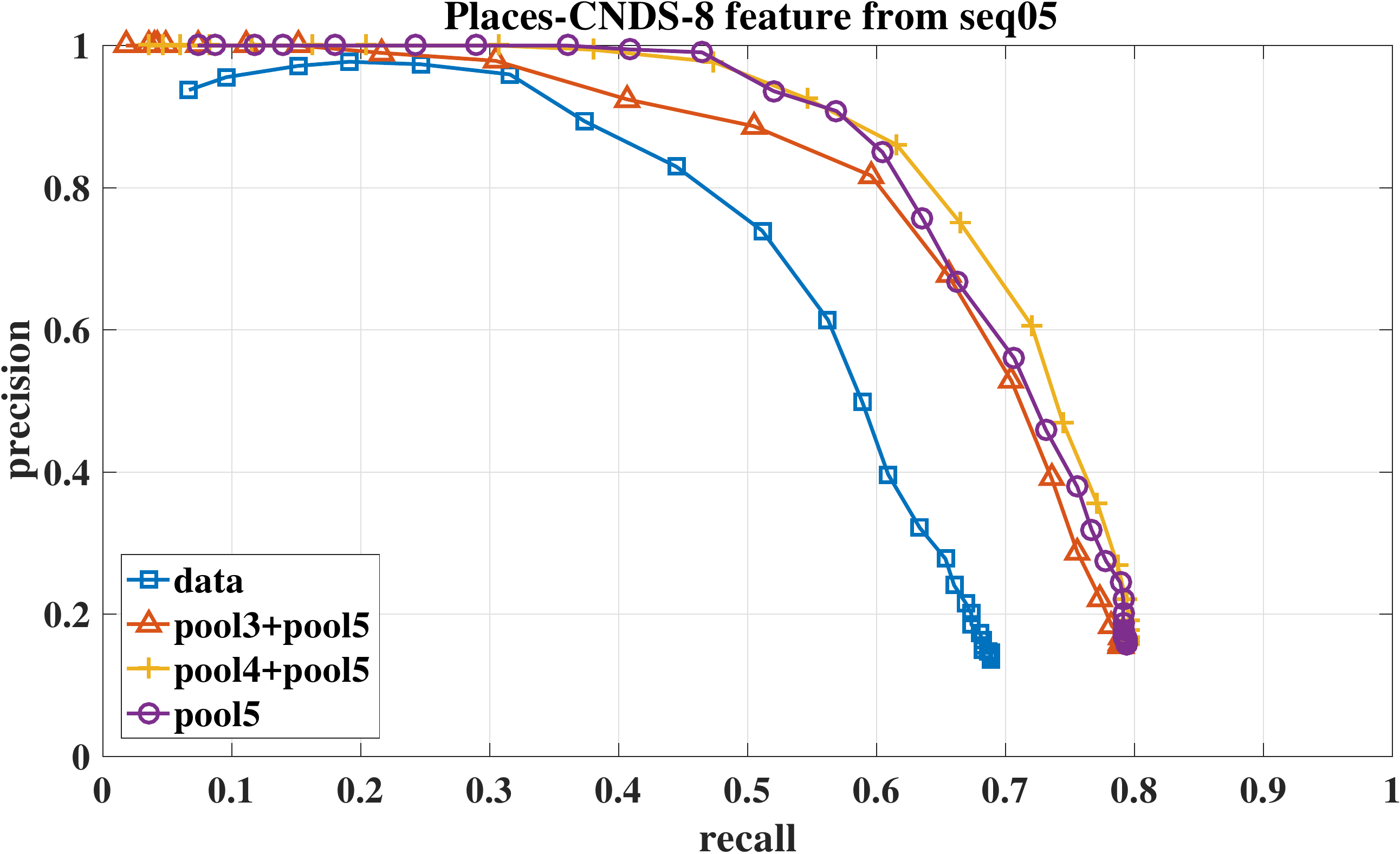}
	\includegraphics[width=0.32\textwidth,height=0.2\textwidth]{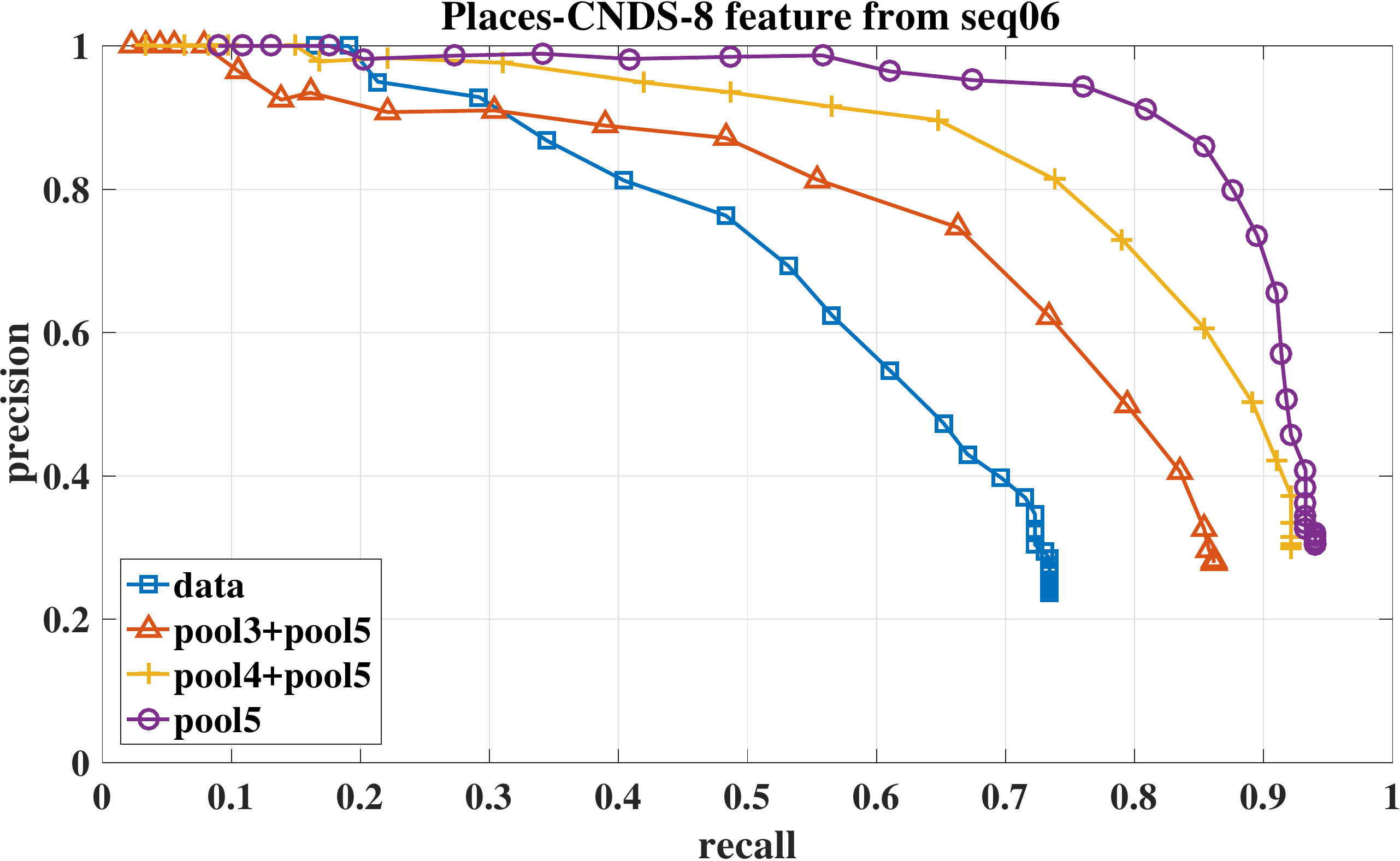}
	\caption{Precision-vs-recall performance of features extracted from different CNNs and different layers. `data' means directly using the vectorized range image as a feature, `conv3+pool5' means extracting the feature from the `conv3' layer, and using a pooling layer with the same structure as the `pool5' layer for preliminary dimension reduction.  Similarly for `conv4+pool5', `pool3+pool5' and `pool4+pool5'.}
	\label{fig:cnns_performance}
\end{figure*}

As mentioned in Subsec~\ref{subsec:CNNs}, we test our proposed method on AlexNet \cite{Imagenet}, VGG-CNN-S \cite{return-devil} and Places-CNDS-8 \cite{caffenet}.  VGG-CNN-S \cite{return-devil} is modified from AlexNet \cite{Imagenet} by increasing the channels of the last three convolutional layers, and Places-CNDS-8 \cite{caffenet} has a deeper network structure. For AlexNet \cite{Imagenet}, we test the feature from \textit{conv3}, \textit{conv4} and \textit{pool5}.  When extracting the features of \textit{conv3} and \textit{conv4}, we insert a pooling layer after them that has the same parameters as \textit{pool5} for preliminary dimension reduction, and name the resultant features \textit{conv3+pool5} and \textit{conv4+pool5} respectively.  Similarly, we use the features \textit{conv3+pool5}, \textit{conv4+pool5} and \textit{pool5} from VGG-CNN-S \cite{return-devil}; and \textit{pool3+pool5}, \textit{pool4+pool5} and \textit{pool5} from Places-CNDS-8 \cite{caffenet}.  We also show the performance of directly using the vectorized range image as the global descriptor for comparison.  As shown in Figure~\ref{fig:cnns_performance}, all the CNN features outperform the vectorized range images.  Both the wider VGG-CNN-S \cite{return-devil} and deeper Places-CNDS-8 \cite{caffenet} have a better performance than AlexNet \cite{Imagenet}.  The best performing layers differ from network to network.  In AlexNet \cite{Imagenet},  \textit{conv3+pool5} offers the best features, in VGG-CNN-S \cite{return-devil} it is \textit{conv4+pool5} and in Places-CNDS-8 \cite{caffenet} it is \textit{pool5}.

\subsection{Easy task vs median difficulty task}
\begin{figure*}
	\centering
	\includegraphics[width=0.32\textwidth,height=0.2\textwidth]{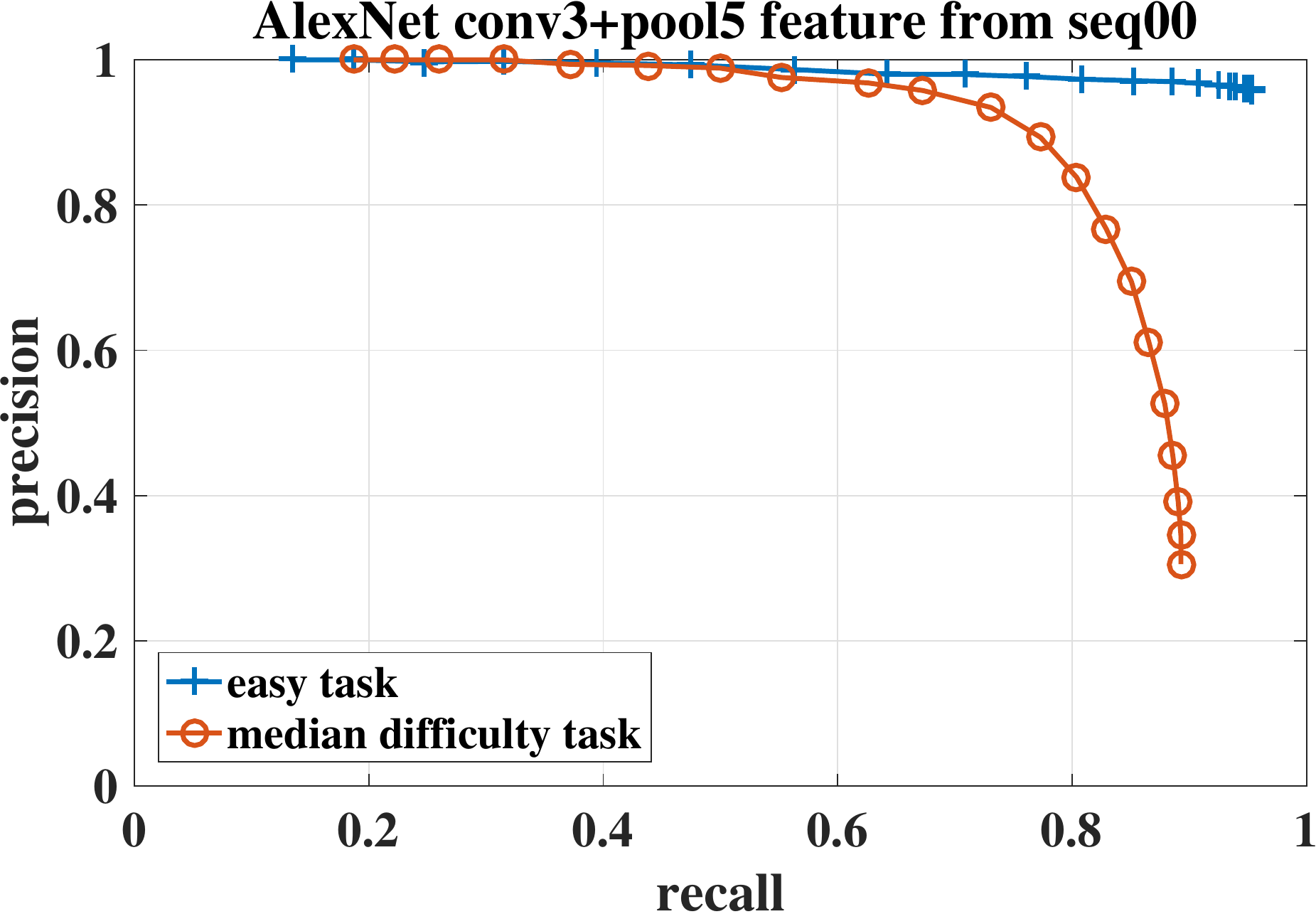}
	\includegraphics[width=0.32\textwidth,height=0.2\textwidth]{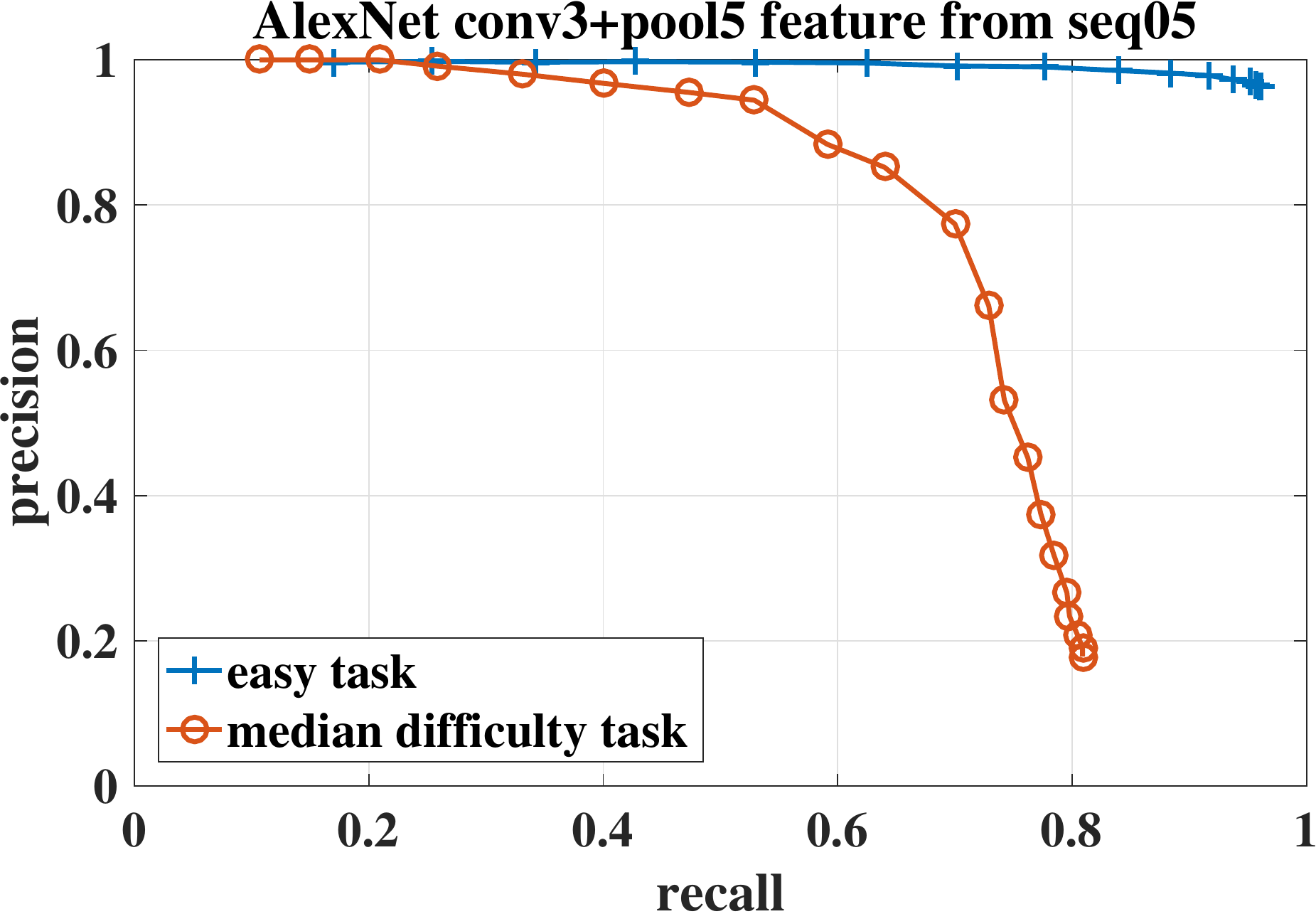}
	\includegraphics[width=0.32\textwidth,height=0.2\textwidth]{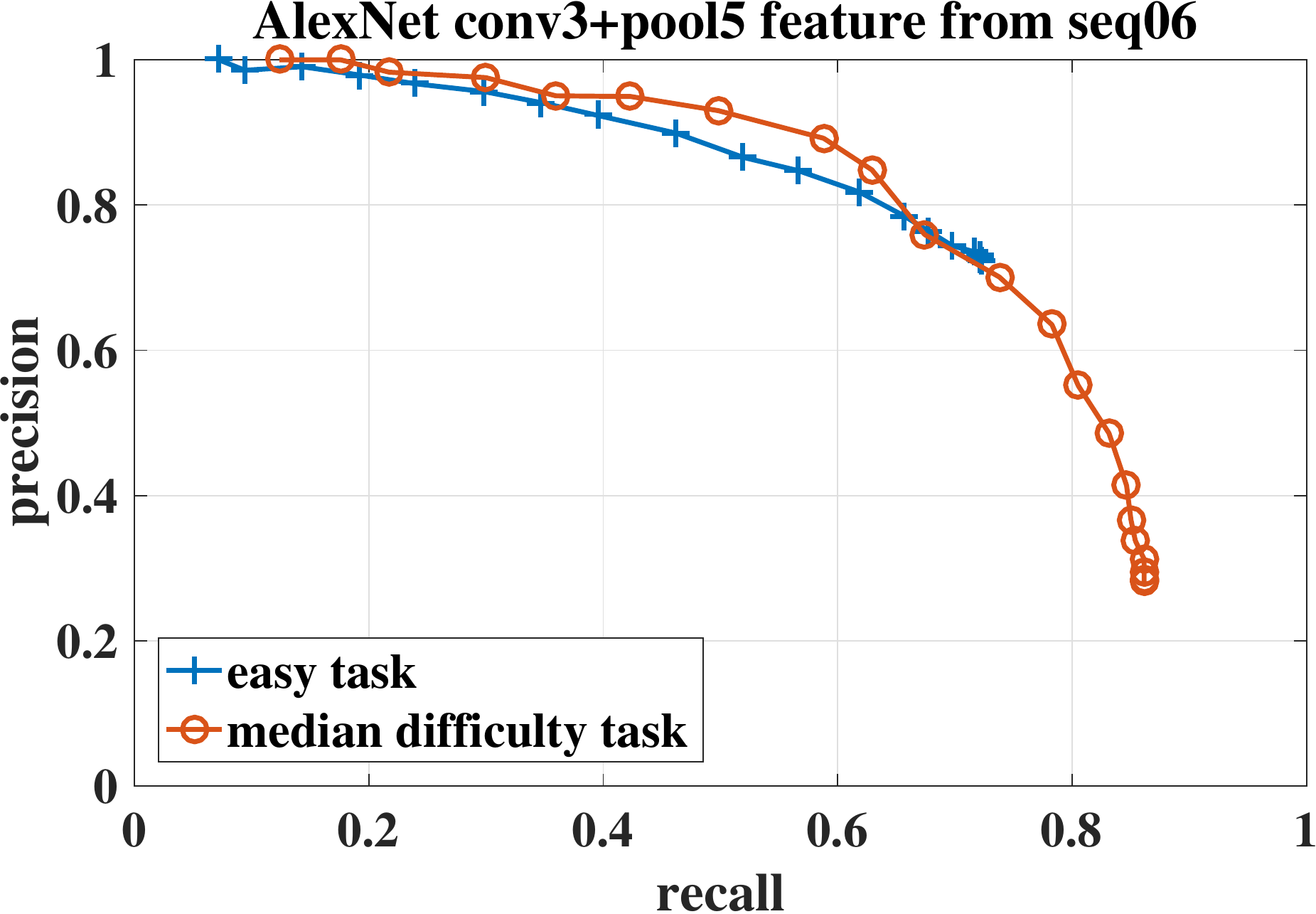}
	\includegraphics[width=0.32\textwidth,height=0.2\textwidth]{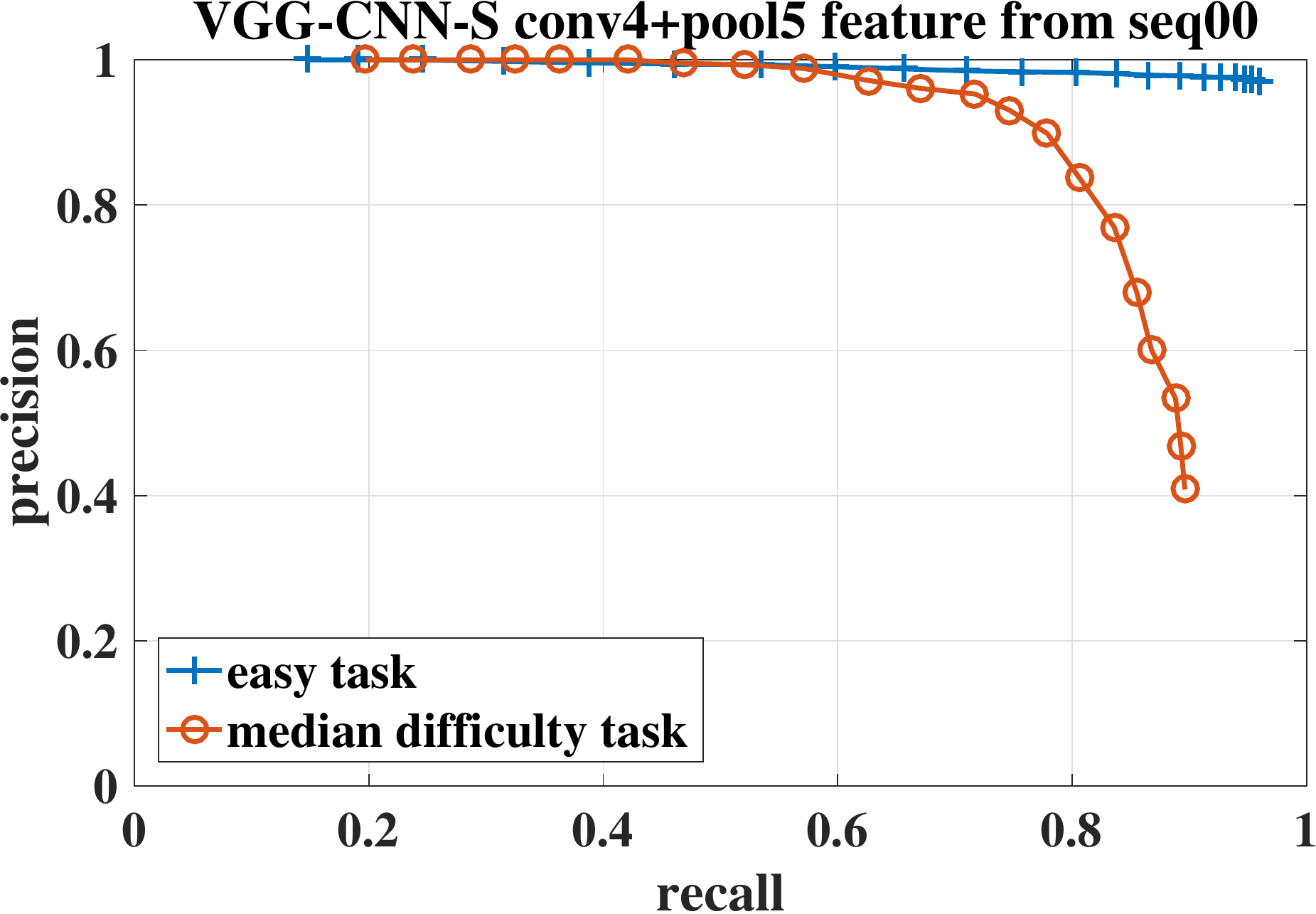}
	\includegraphics[width=0.32\textwidth,height=0.2\textwidth]{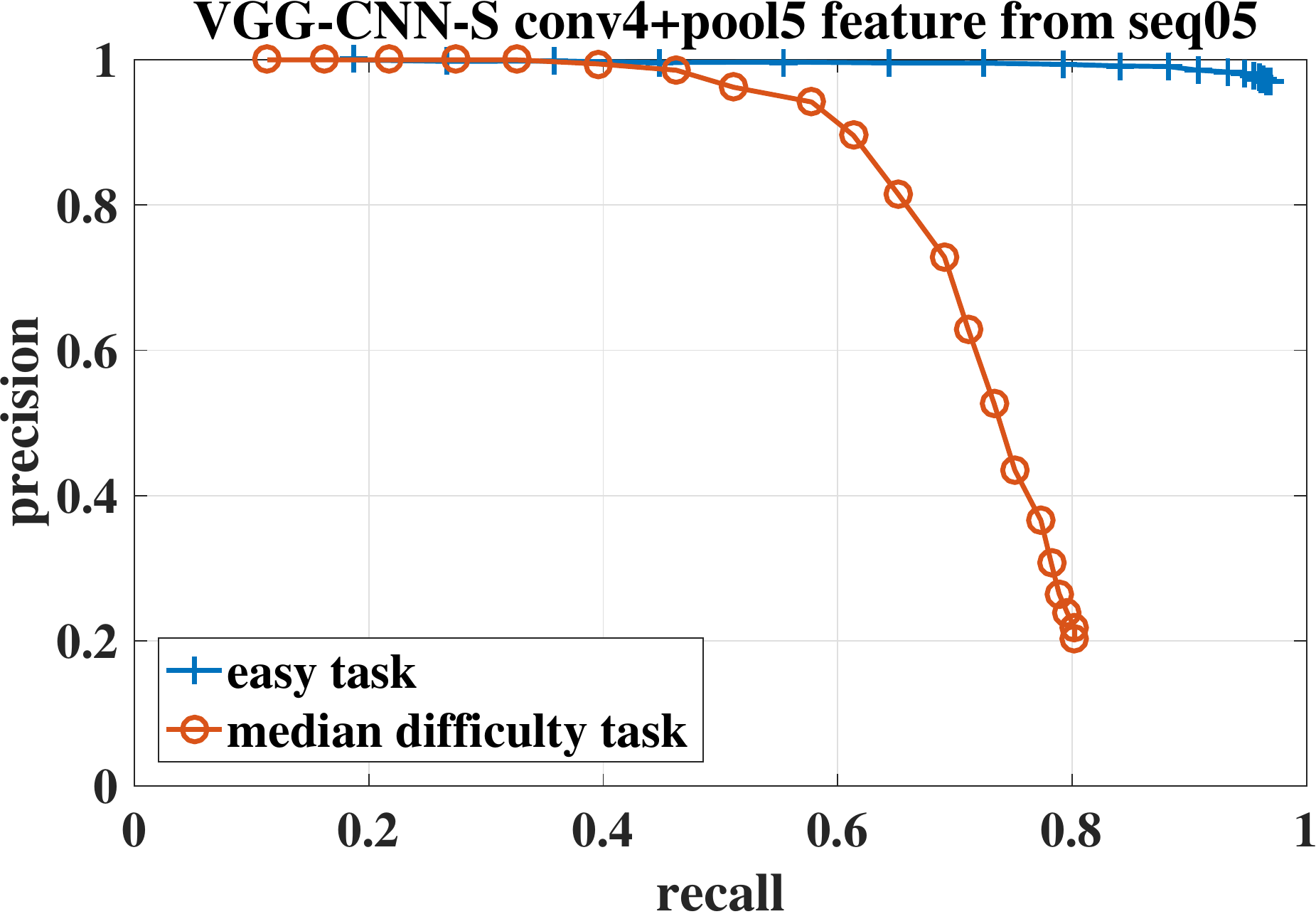}
	\includegraphics[width=0.32\textwidth,height=0.2\textwidth]{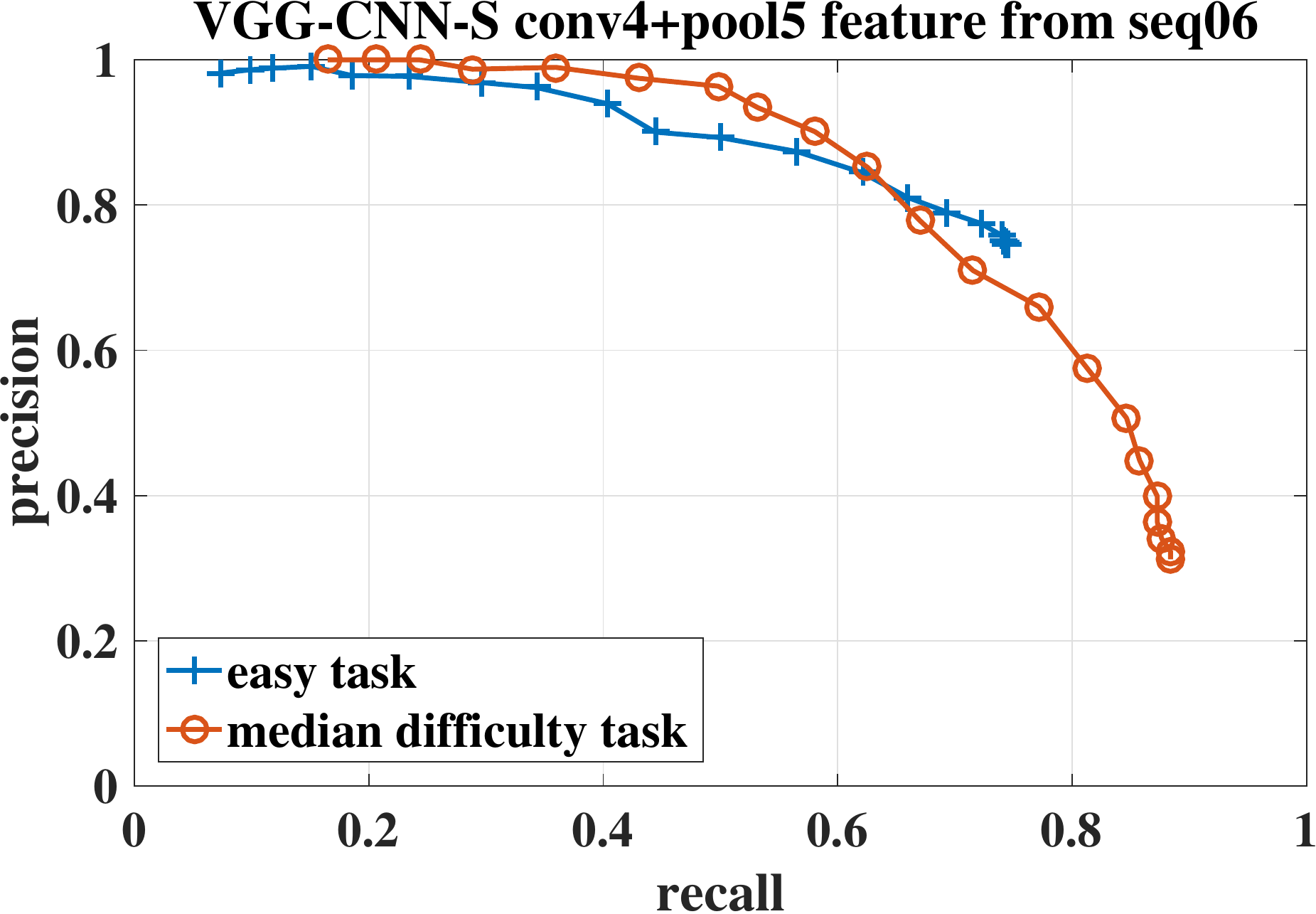}
	\includegraphics[width=0.32\textwidth,height=0.2\textwidth]{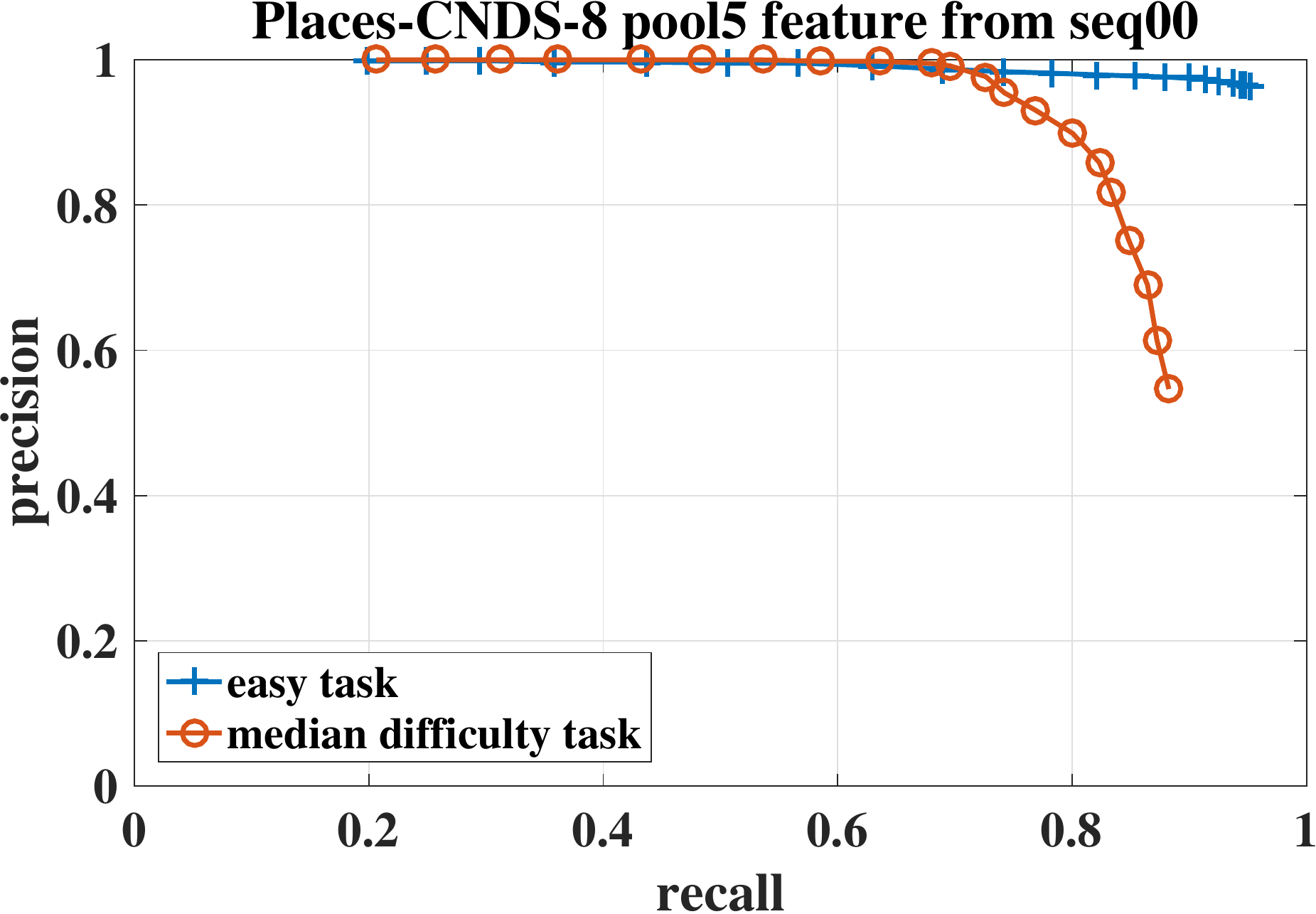}
	\includegraphics[width=0.32\textwidth,height=0.2\textwidth]{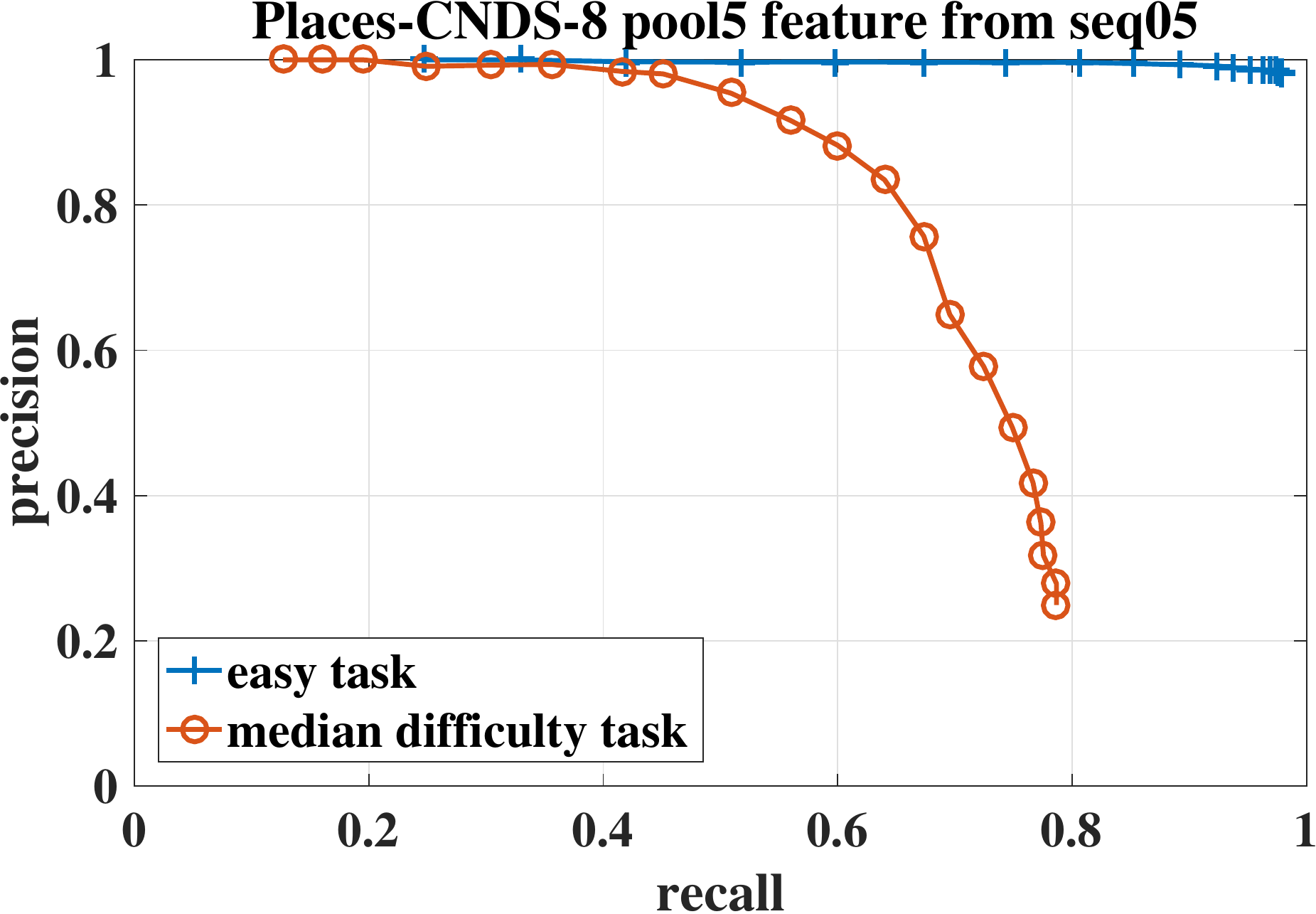}
	\includegraphics[width=0.32\textwidth,height=0.2\textwidth]{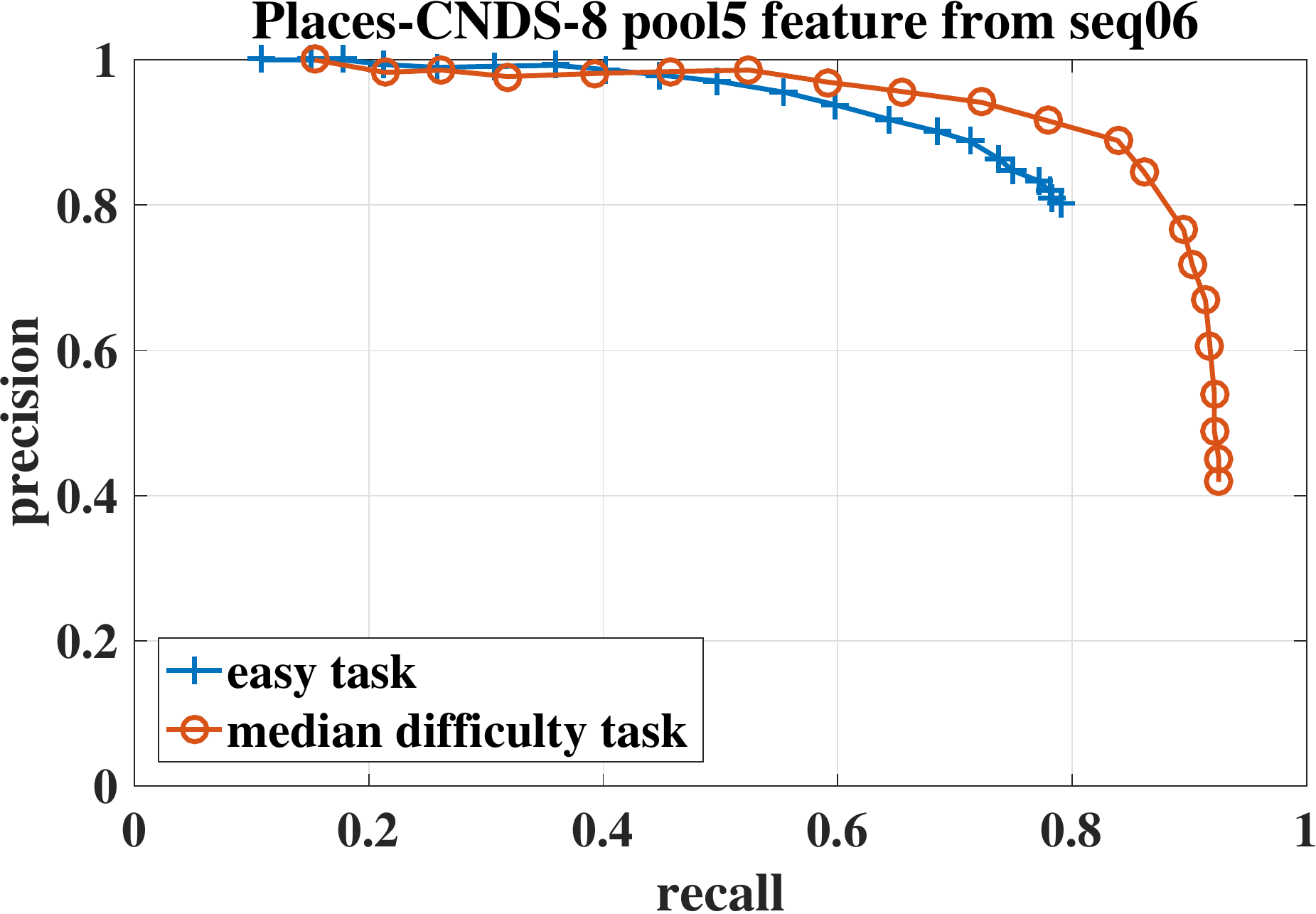}
	\caption{Precision-vs-recall curves of easy task and median difficulty task are plotted together in this figure.  A detailed description can be found in the text.}
	\label{fig:easy_median}
\end{figure*}

In the easy task, we uniformly sample every 3\textit{rd} scna in a sequence as a stored map, and leave the rest as queries; while the median difficulty task stores one path in the manually selected unidirectional loop cloures.  The results of two types of tasks are plotted together in Figure~\ref{fig:easy_median}.  It can be seen that for sequence 00 and sequence 05, the easy task achieves much better performance than the median difficulty task, especially in the high recall region.  Sequence 06 roughly shapes like a rectangle (see Figure~\ref{fig:show_path}), and the car passes two of the edges at high speed and the other two at relatively low speed.  The unidirectional loop closure is completely contained in the `low-speed edge', where the scans densely cover the road.  Using uniform sampling, we actually create a lot of true matches along the `high-speed edge', where even the consecutive scans have a large position change, and we also create a lot of unmatched but close pairs at the same time.  Thus, the performance of `easy task' in sequence 06 drops.

\subsection{Effectiveness of PCA alignment}
\begin{figure*}
	\centering
	\includegraphics[width=0.32\textwidth,height=0.2\textwidth]{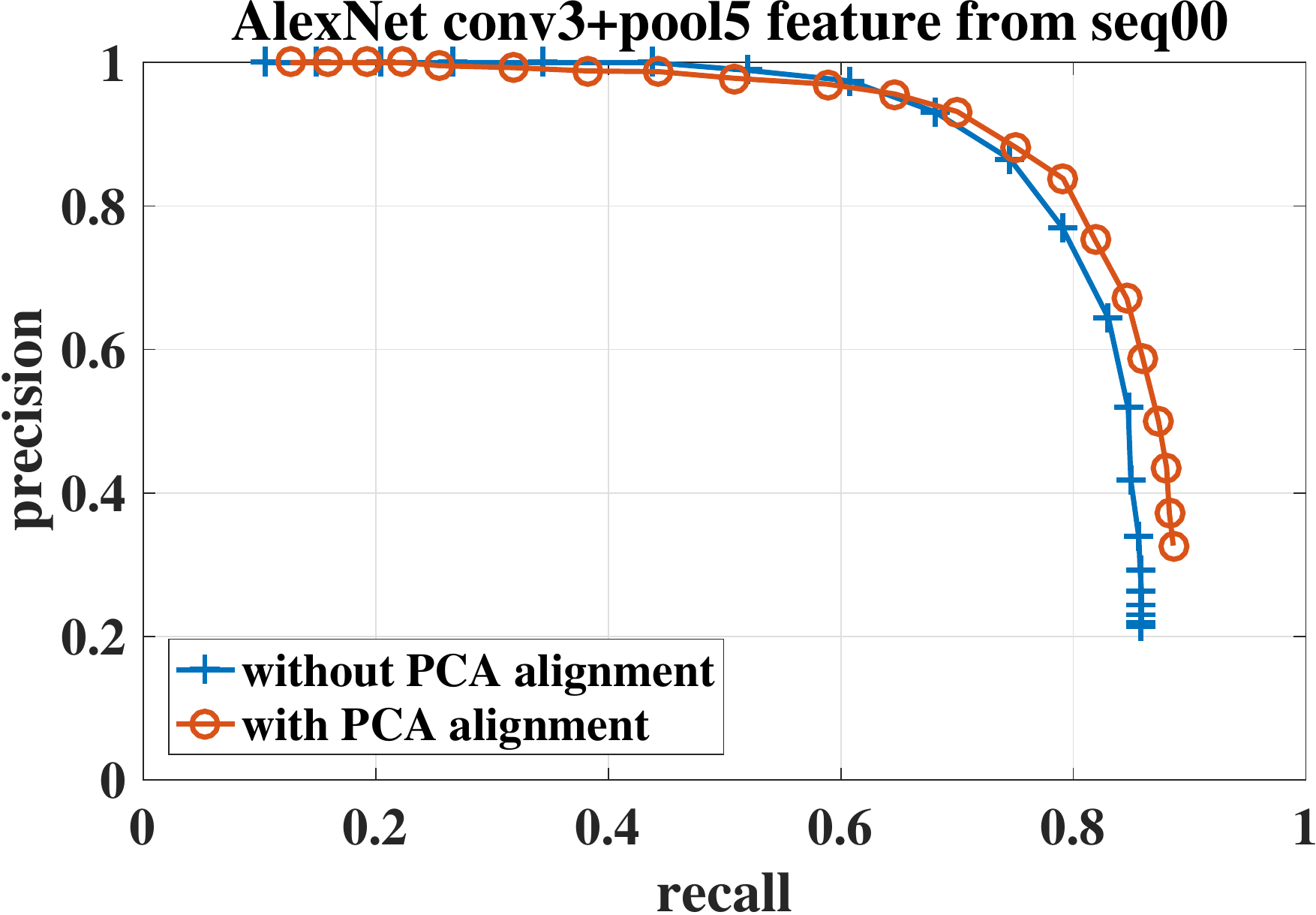}
	\includegraphics[width=0.32\textwidth,height=0.2\textwidth]{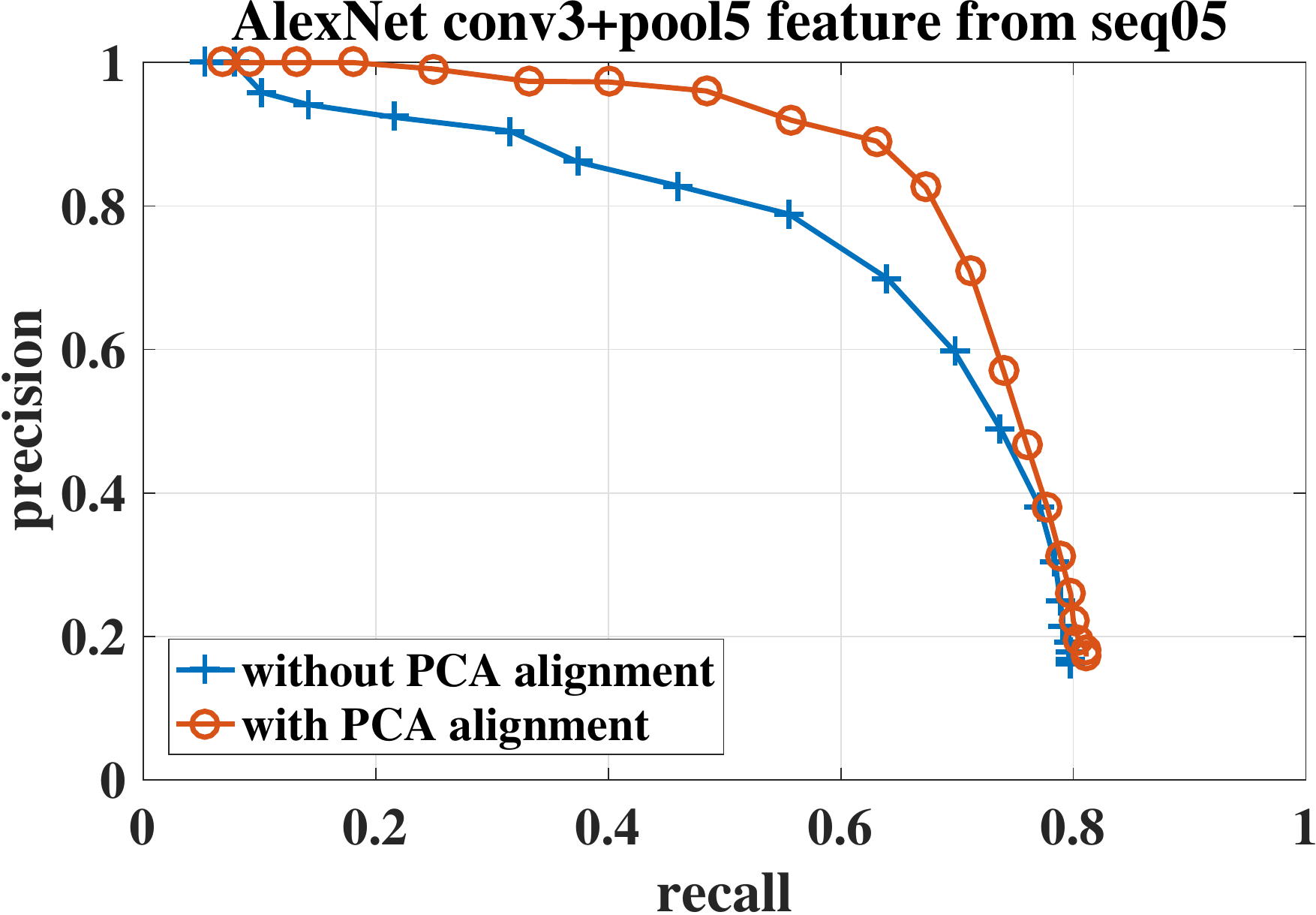}
	\includegraphics[width=0.32\textwidth,height=0.2\textwidth]{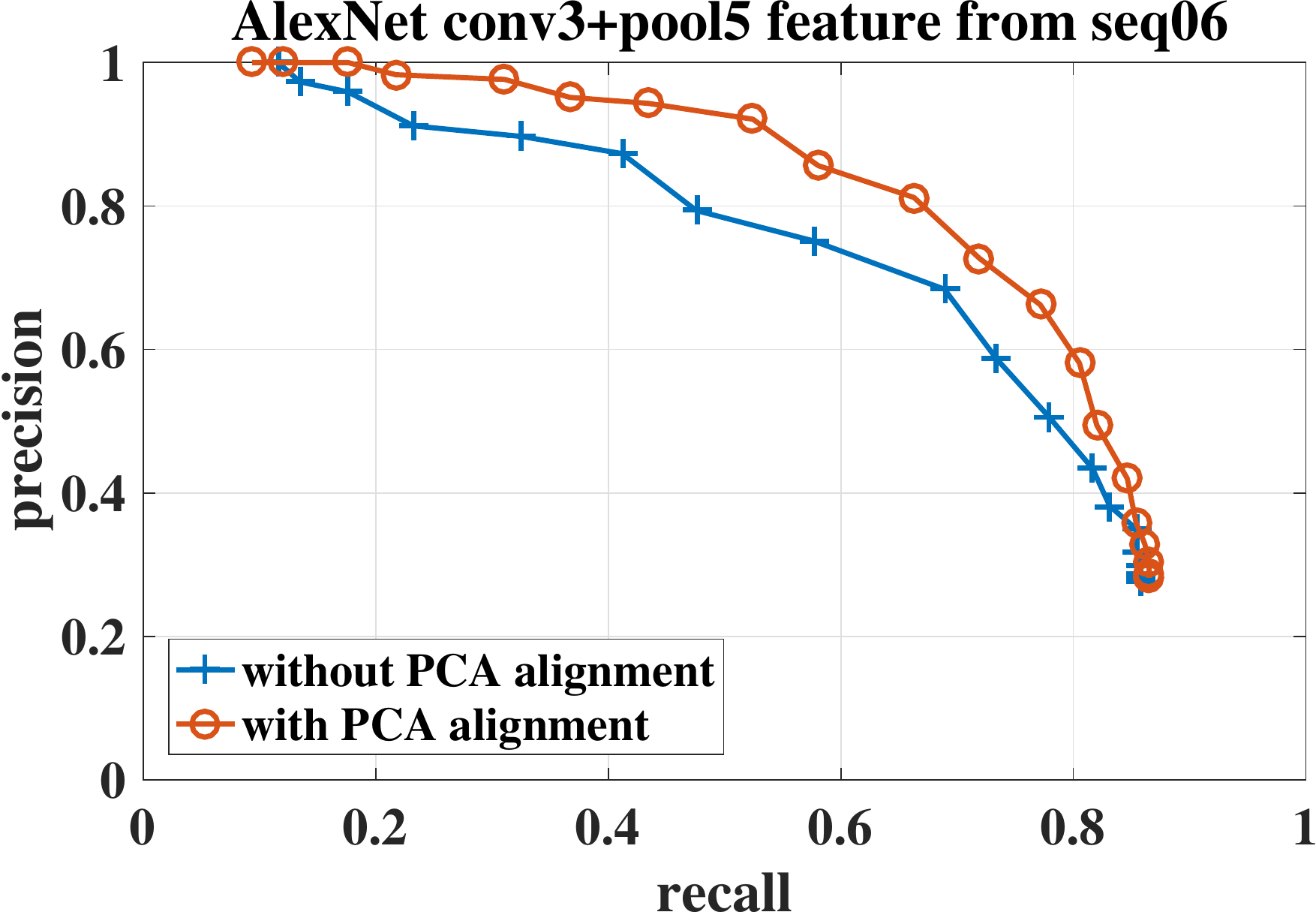}
	\includegraphics[width=0.32\textwidth,height=0.2\textwidth]{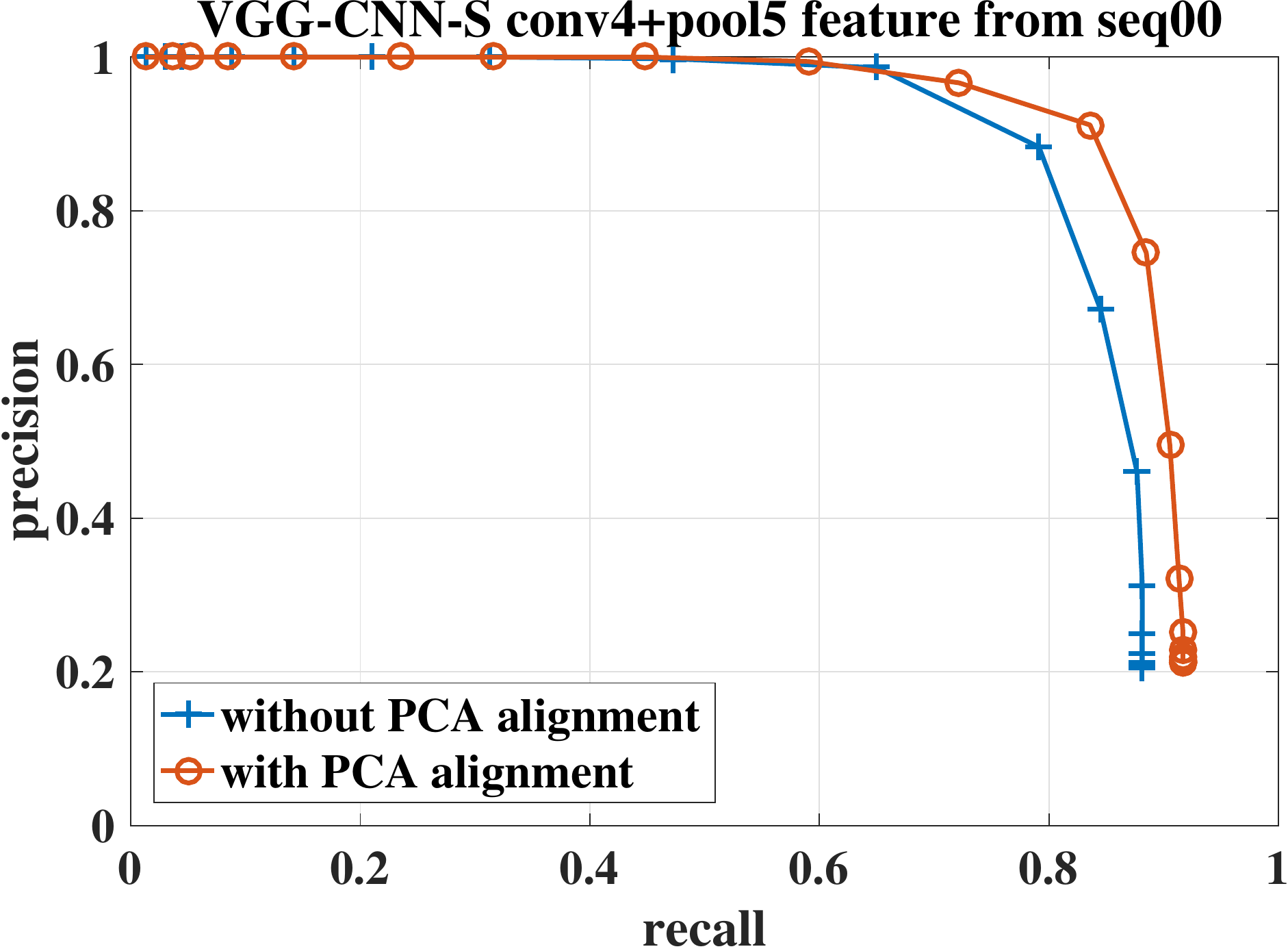}
	\includegraphics[width=0.32\textwidth,height=0.2\textwidth]{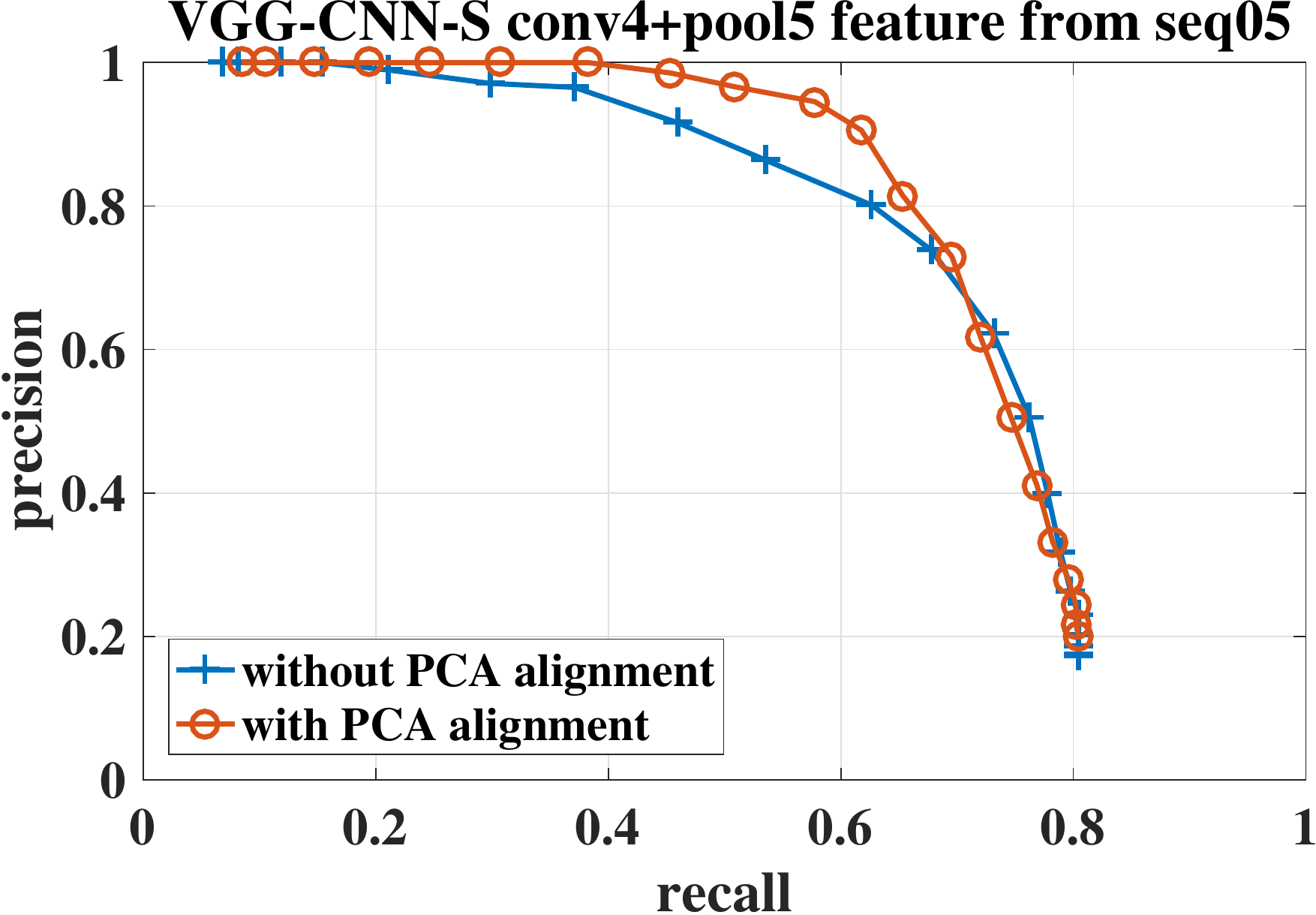}
	\includegraphics[width=0.32\textwidth,height=0.2\textwidth]{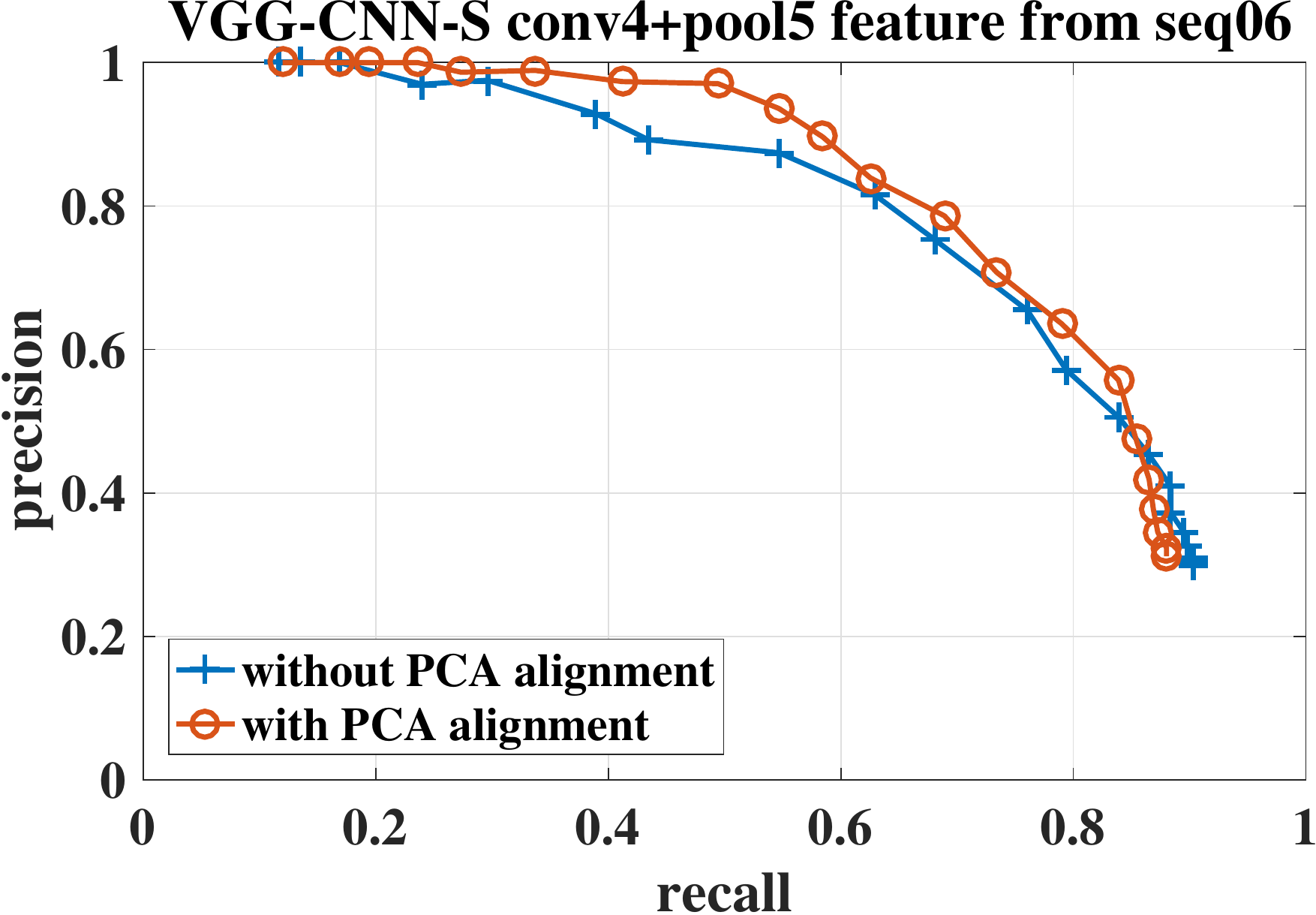}
	\includegraphics[width=0.32\textwidth,height=0.2\textwidth]{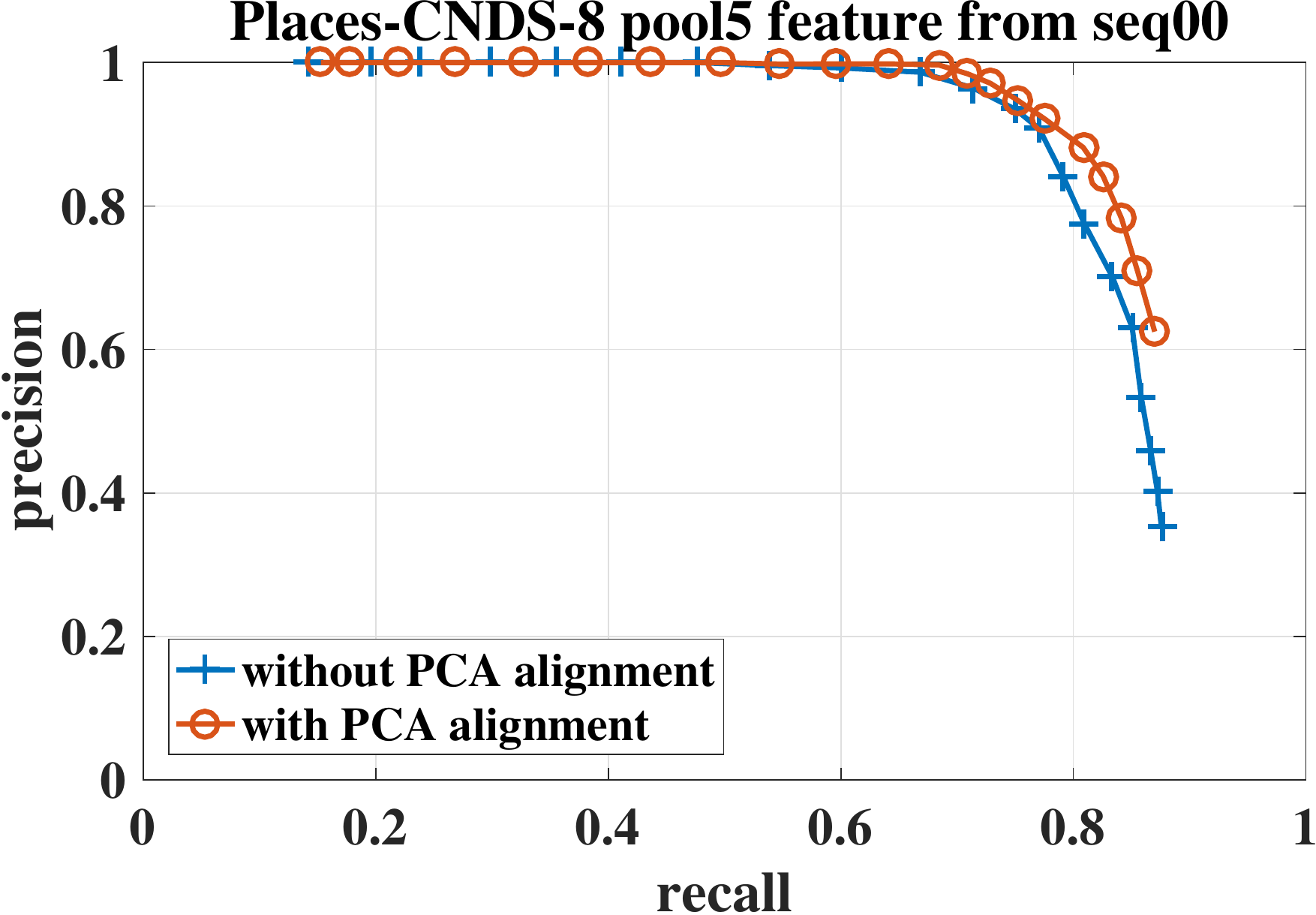}
	\includegraphics[width=0.32\textwidth,height=0.2\textwidth]{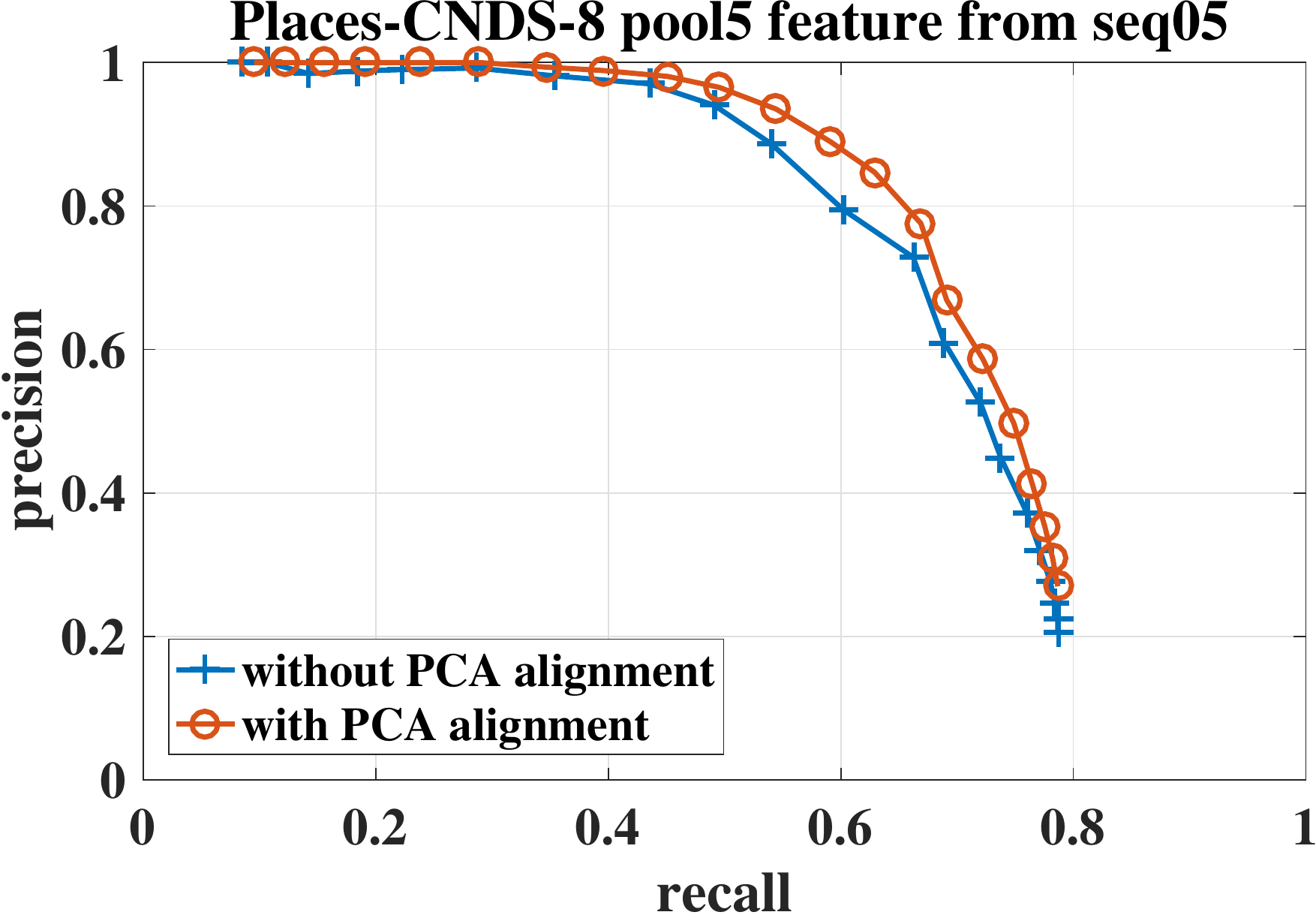}
	\includegraphics[width=0.32\textwidth,height=0.2\textwidth]{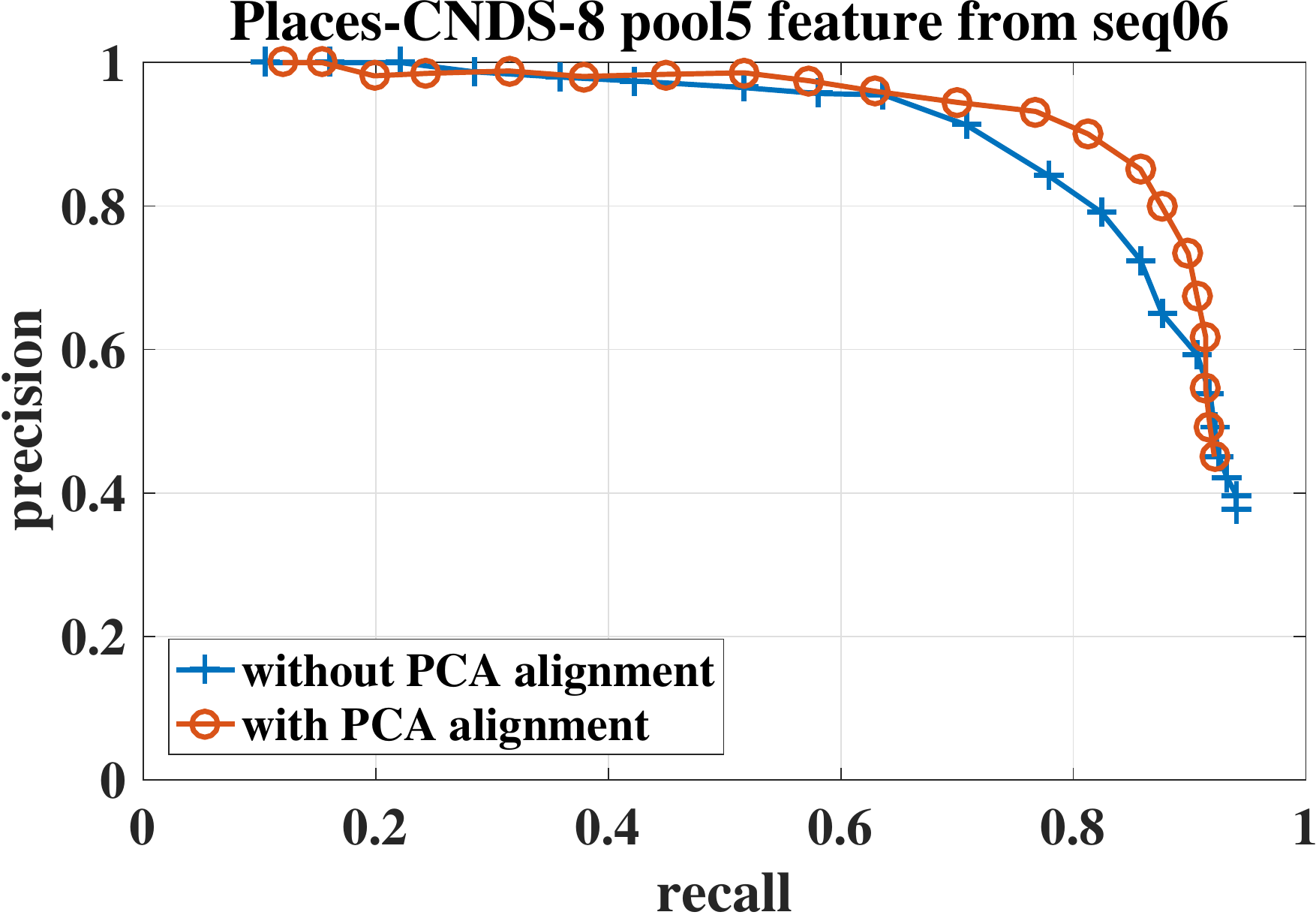}
	\caption{The precision-vs-recall curves with and without PCA alignment mentioned in Subsec.~\ref{subsec:Preprocessing} are plotted together in this figure.}
	\label{fig:pca_alignment}
\end{figure*}

The precision-vs-recall curves with and without the PCA alignment mentioned in Subsec.~\ref{subsec:Preprocessing} are plotted together in Figure~\ref{fig:pca_alignment}, where each row contains the performance of the same feature of different test sequences.  It can be seen that PCA alignment in our preprocessing module significantly helps to promote the performance in all cases.  Another observation is that which regions on the curves show more increase depends on the sequences.  Except sequence 00, the performance of both sequence 05 and 06 is promoted obviously in the high and middle part of the curves, but tends to merge in the low precision region.  This is a common situation when the similarity between true match pairs is decreased by the direction misalignment, and our PCA alignment module eliminates this effect.  Sequence 00 contains the most sharp turnings, and it can be seen that without PCA alignment, some of the true match pairs have such a low score that they fail to stand out among unmatched pairs even when the threshold is lowed, leaving the gap between two curves in the low precision region.

\subsection{Retrieval using cosine similarity only \textit{vs} using proposed score}
\label{subsec:exp_retrieval}
\begin{figure*}
	\centering
	\includegraphics[width=0.32\textwidth,height=0.2\textwidth]{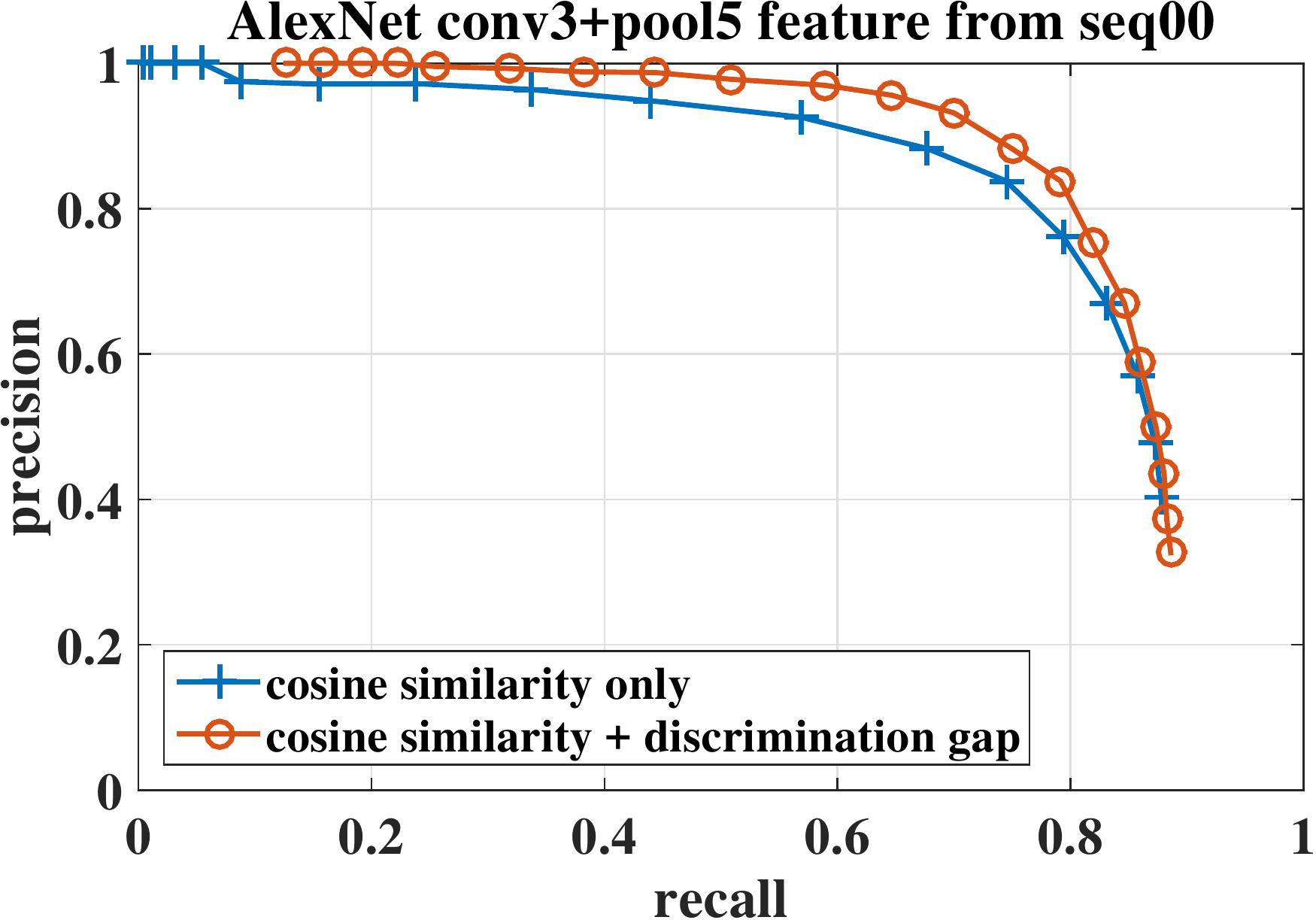}
	\includegraphics[width=0.32\textwidth,height=0.2\textwidth]{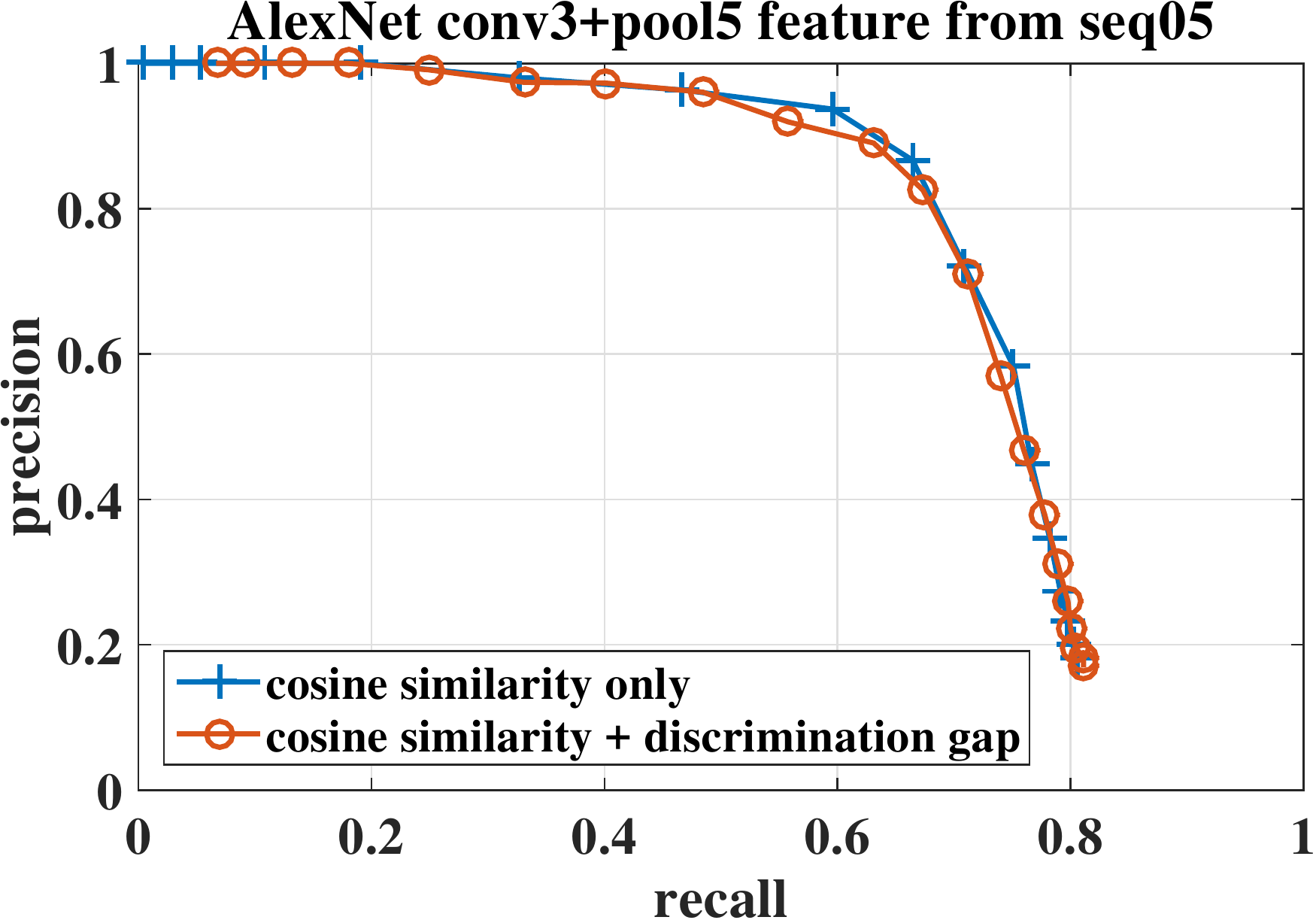}
	\includegraphics[width=0.32\textwidth,height=0.2\textwidth]{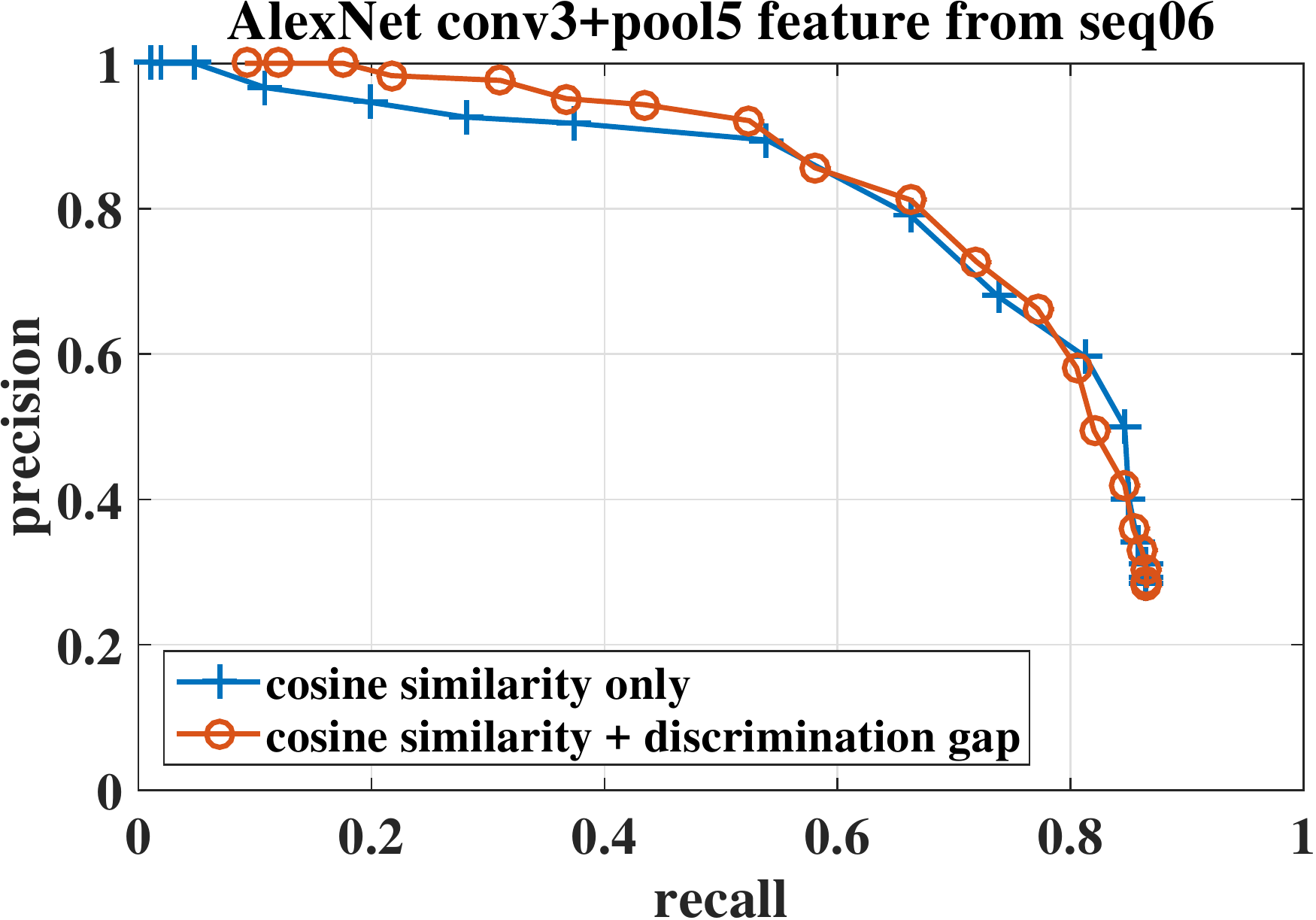}
	\includegraphics[width=0.32\textwidth,height=0.2\textwidth]{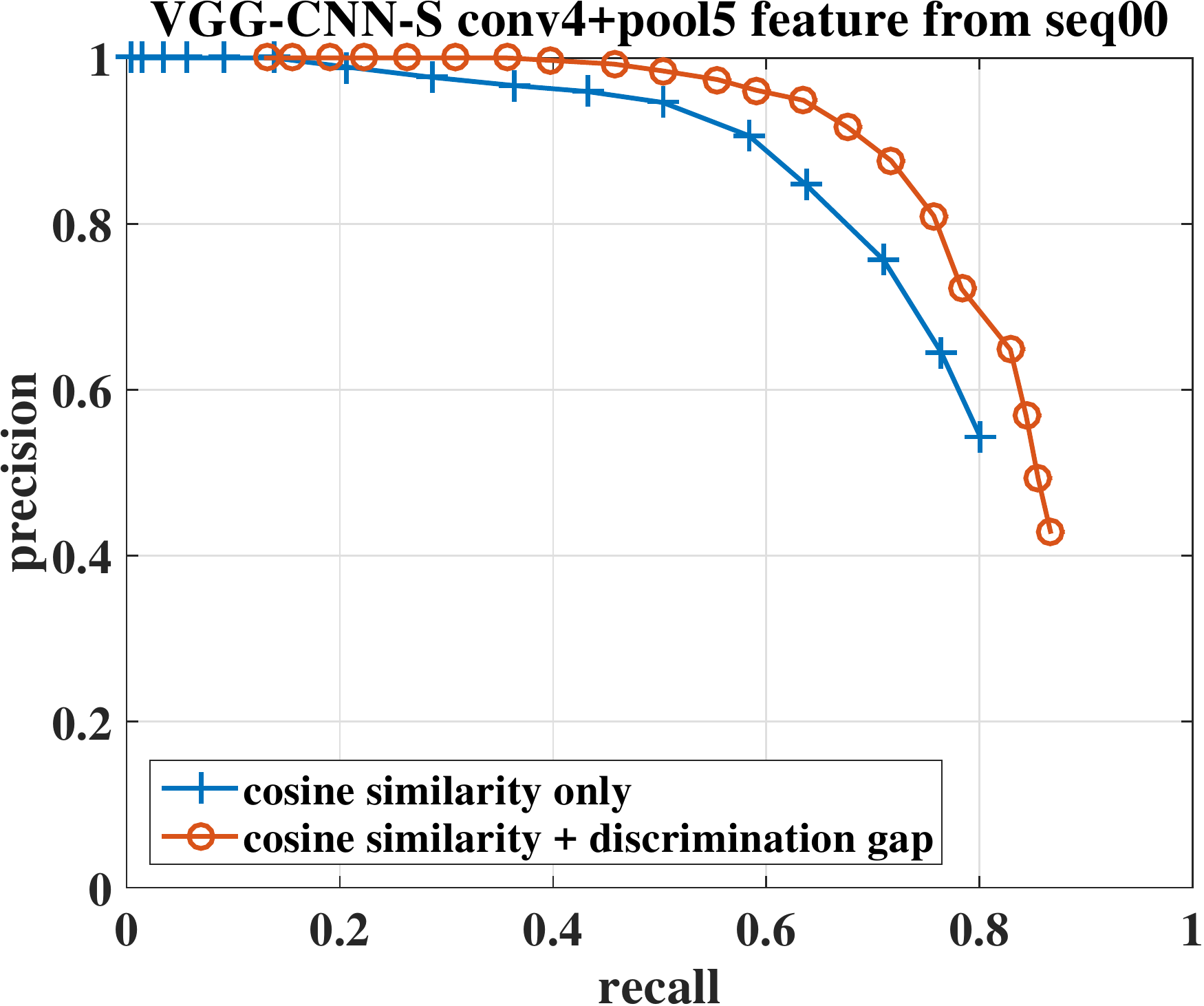}
	\includegraphics[width=0.32\textwidth,height=0.2\textwidth]{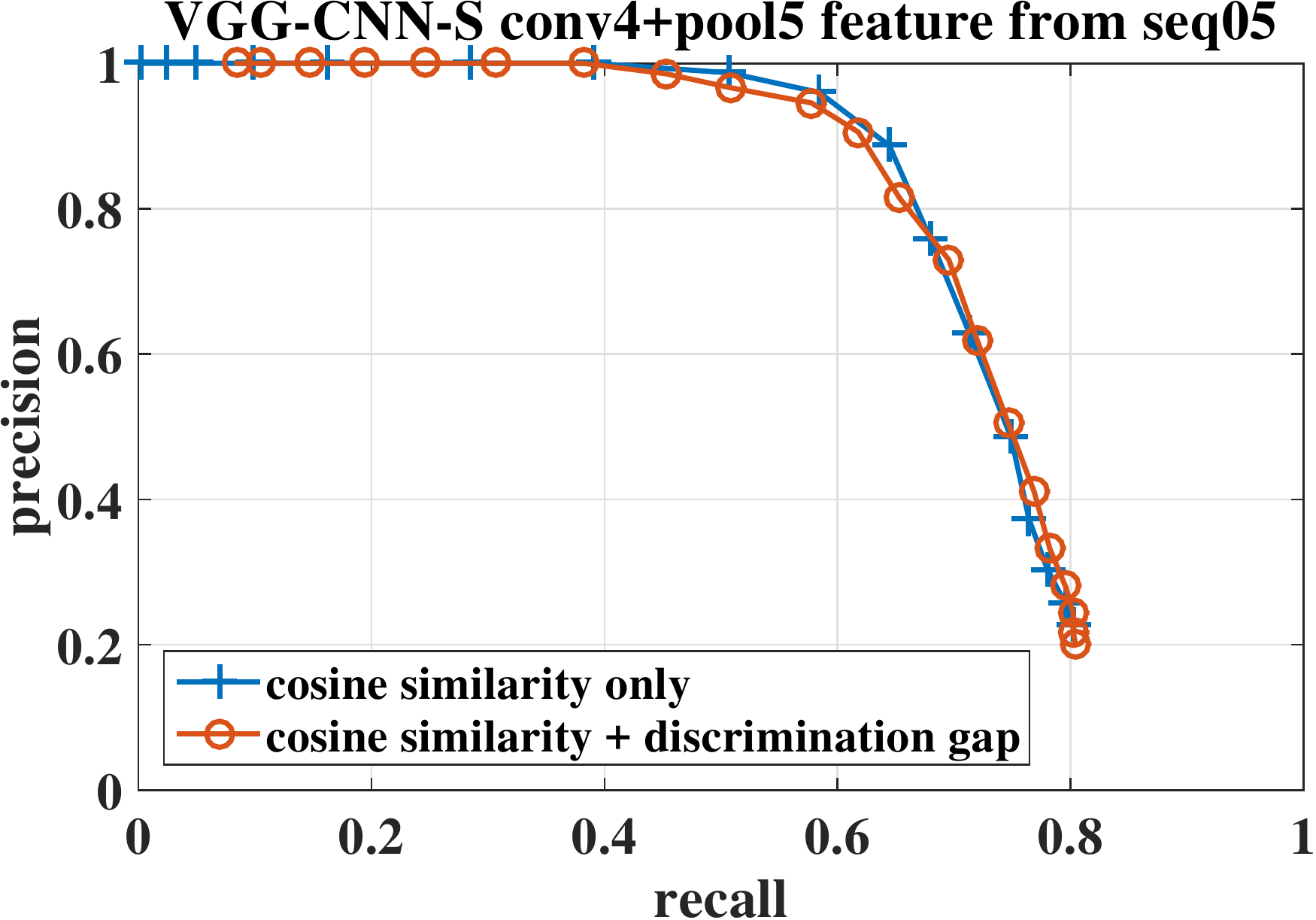}
	\includegraphics[width=0.32\textwidth,height=0.2\textwidth]{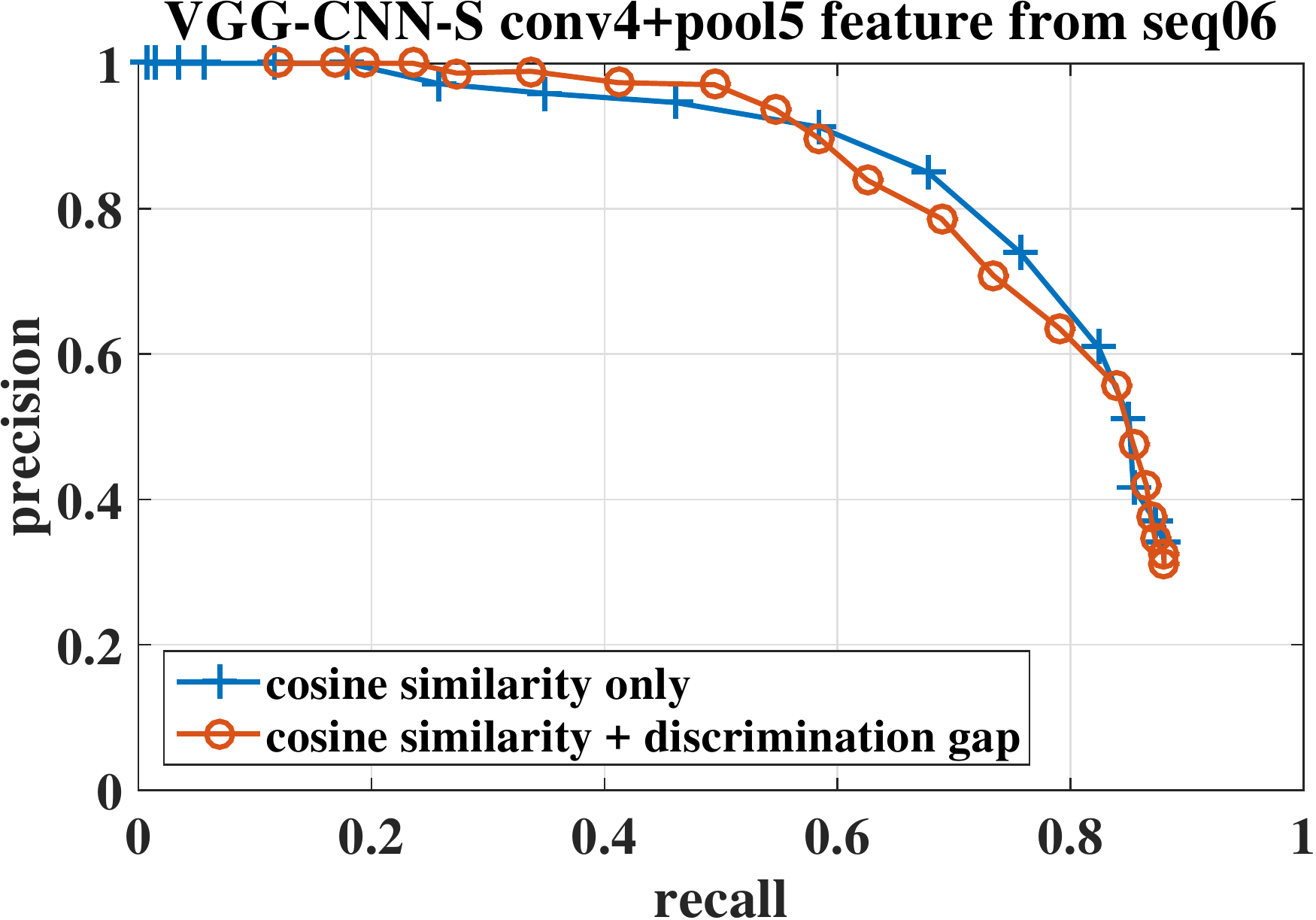}
	\includegraphics[width=0.32\textwidth,height=0.2\textwidth]{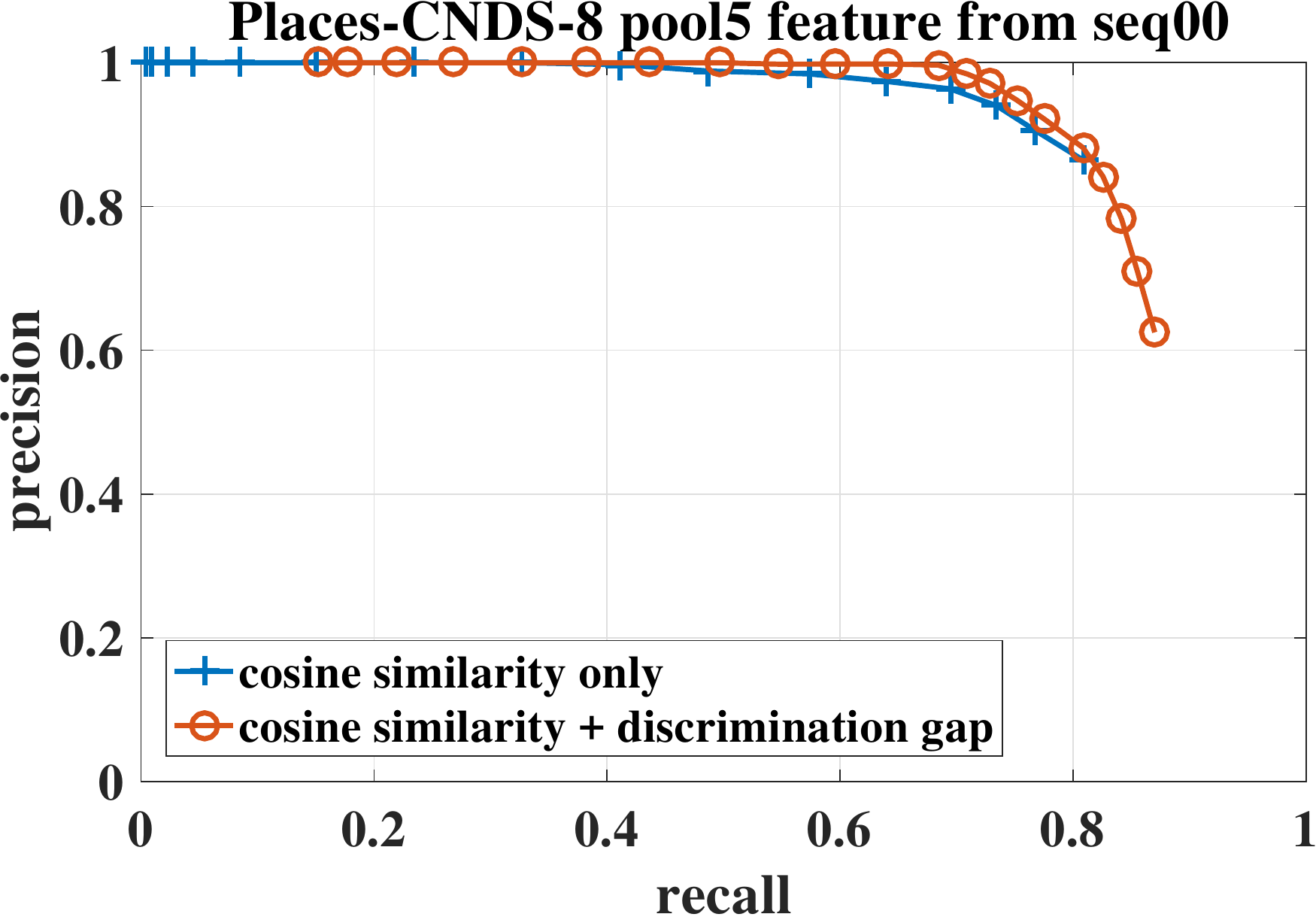}
	\includegraphics[width=0.32\textwidth,height=0.2\textwidth]{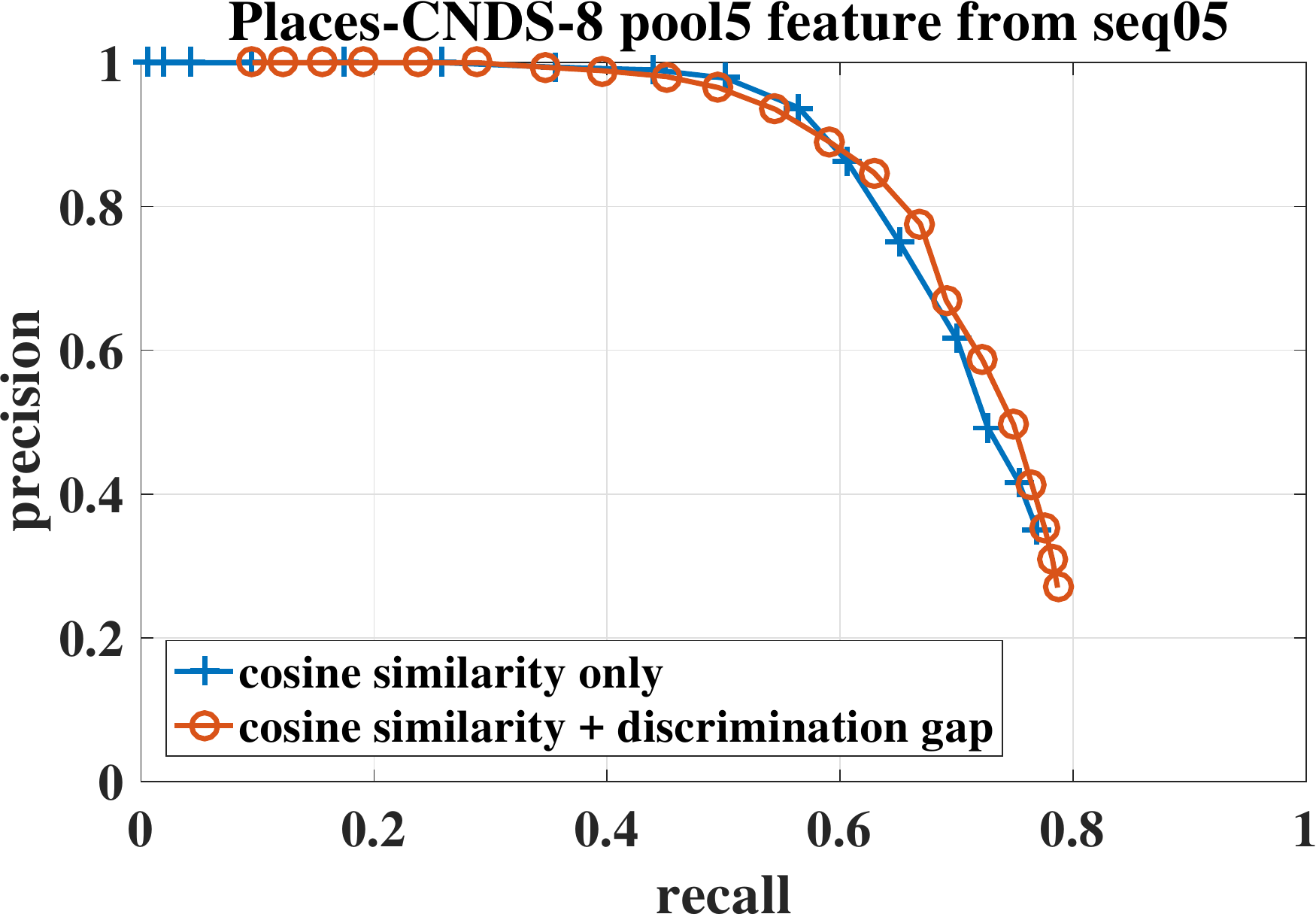}
	\includegraphics[width=0.32\textwidth,height=0.2\textwidth]{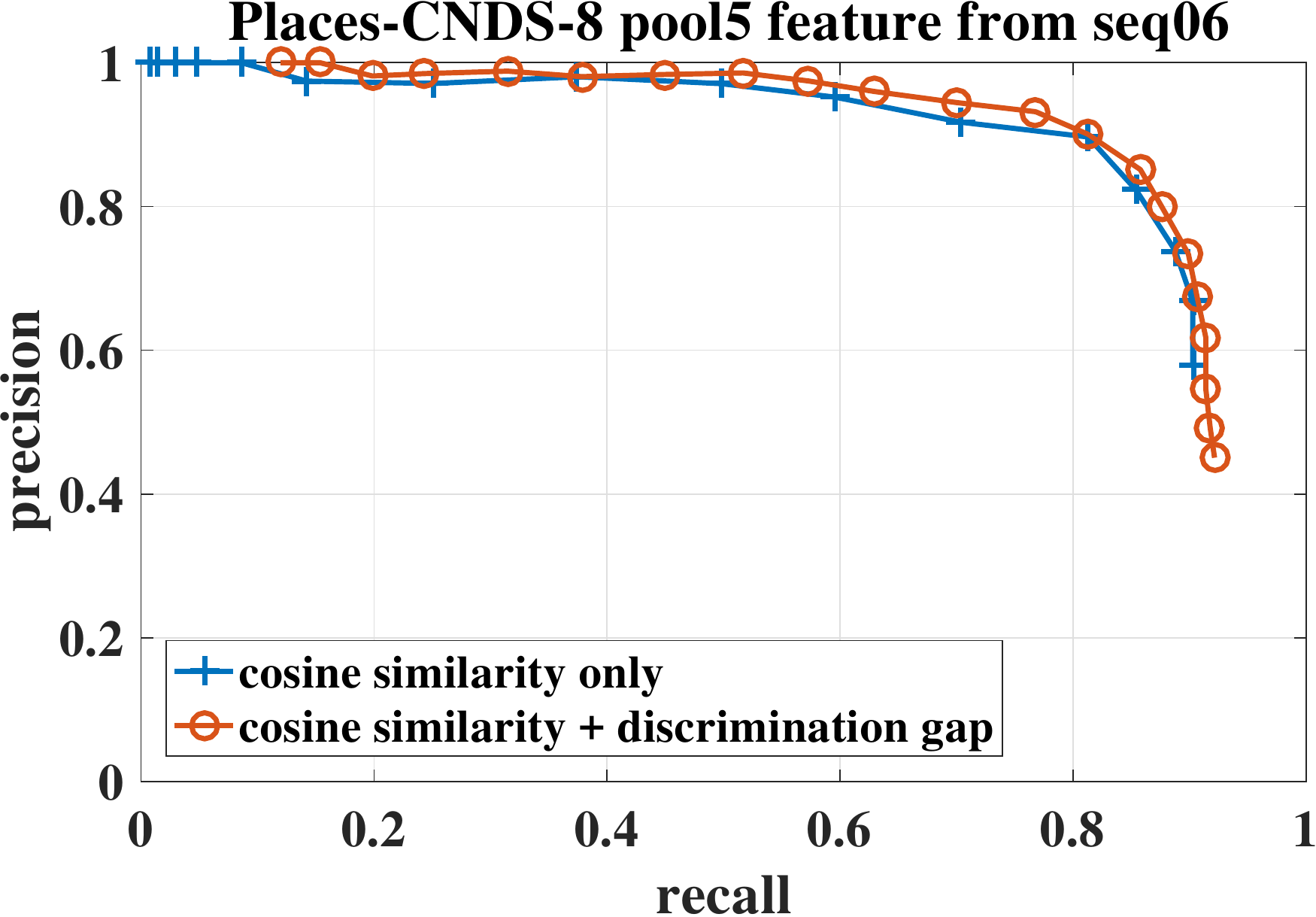}
	\caption{Retrieval using cosine similarity only \textit{vs} using our proposed score = cosine similarity + discrimination gap.}
	\label{fig:precision-recall-score}
\end{figure*}

As discussed in Subsec.~\ref{subsec:retrieval}, during place retrieval, we jointly consider the top 4 best matches according to cosine similarity of the features, and threshold a score modified from cosine similarity by adding a discrimination gap.  The effectiveness of this design is shown in Figure~\ref{fig:precision-recall-score}.  It can be seen that using the proposed score leads to slightly better performance in most of the cases, especially in the high precision region.

\subsection{PCA dimension reduction}
\begin{figure*}
	\centering
	\includegraphics[width=0.32\textwidth,height=0.2\textwidth]{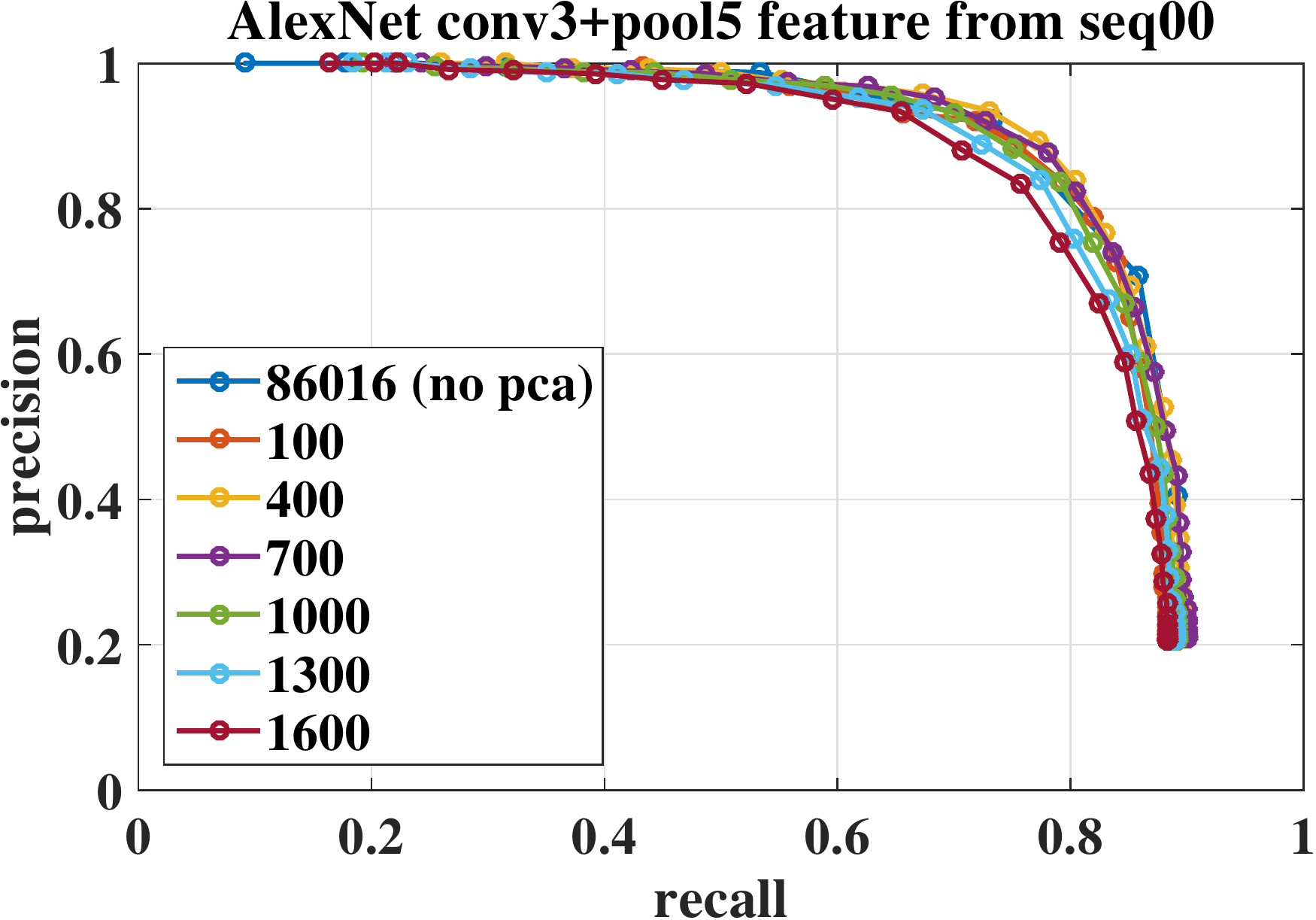}
	\includegraphics[width=0.32\textwidth,height=0.2\textwidth]{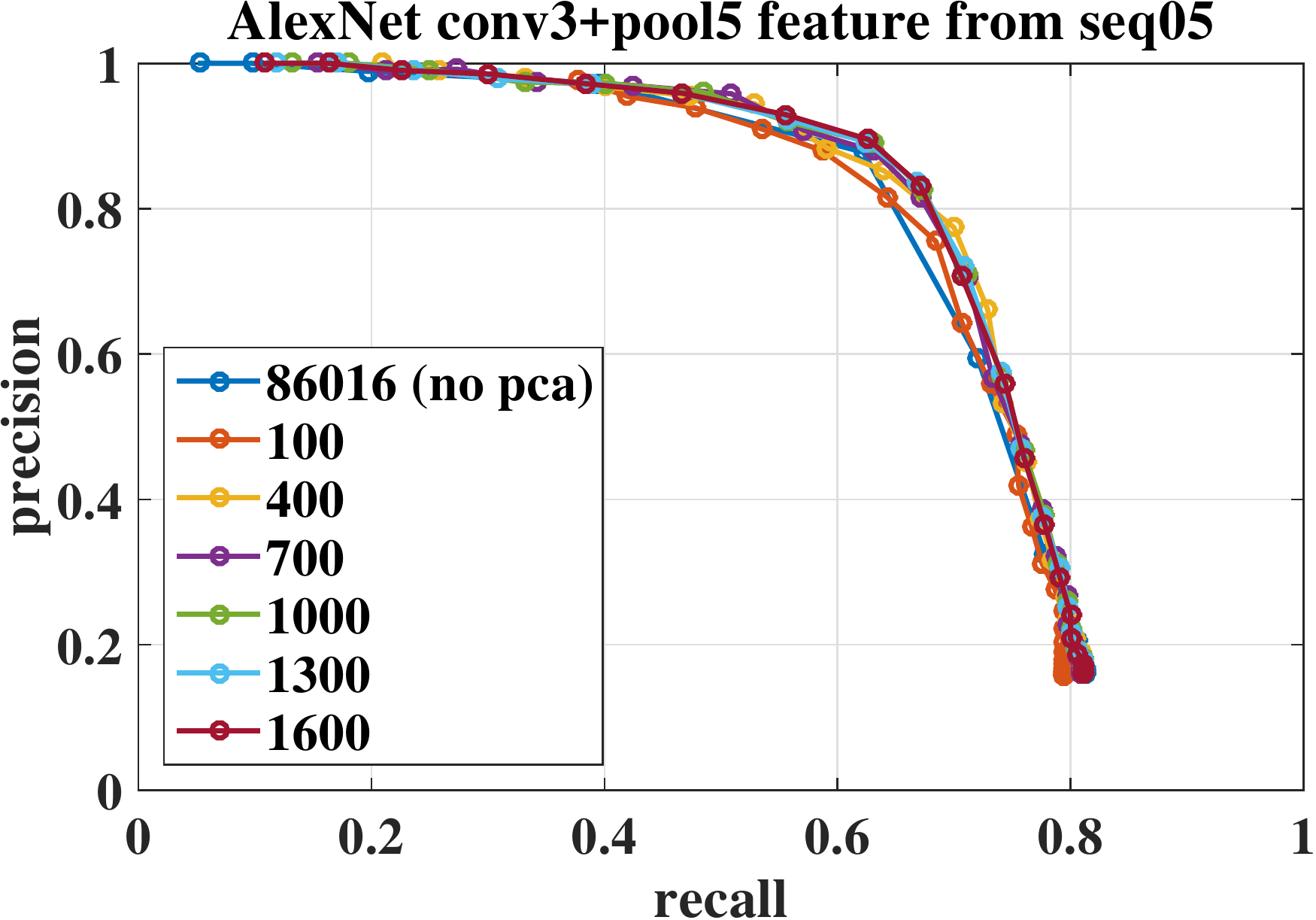}
	\includegraphics[width=0.32\textwidth,height=0.2\textwidth]{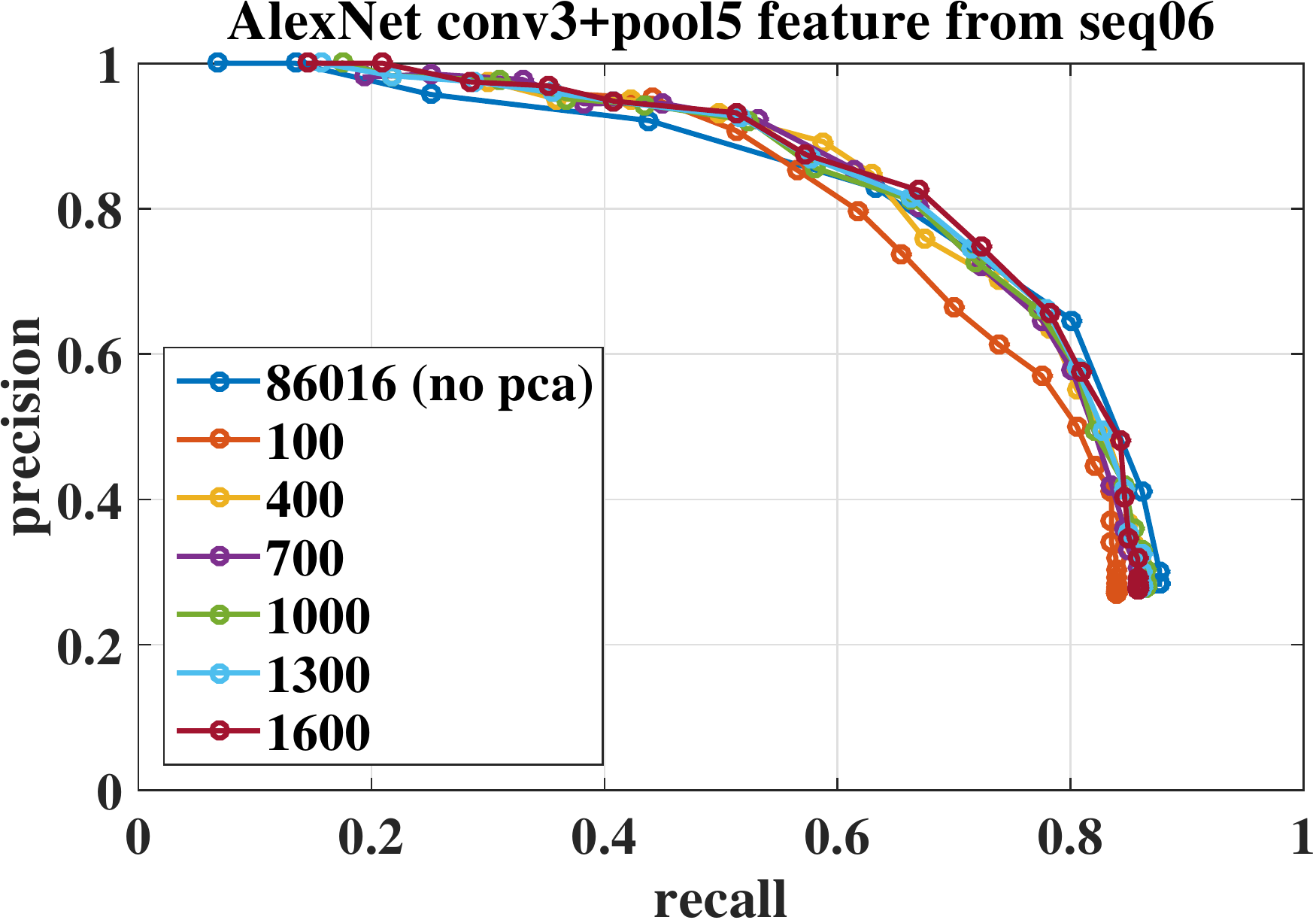}
	\includegraphics[width=0.32\textwidth,height=0.2\textwidth]{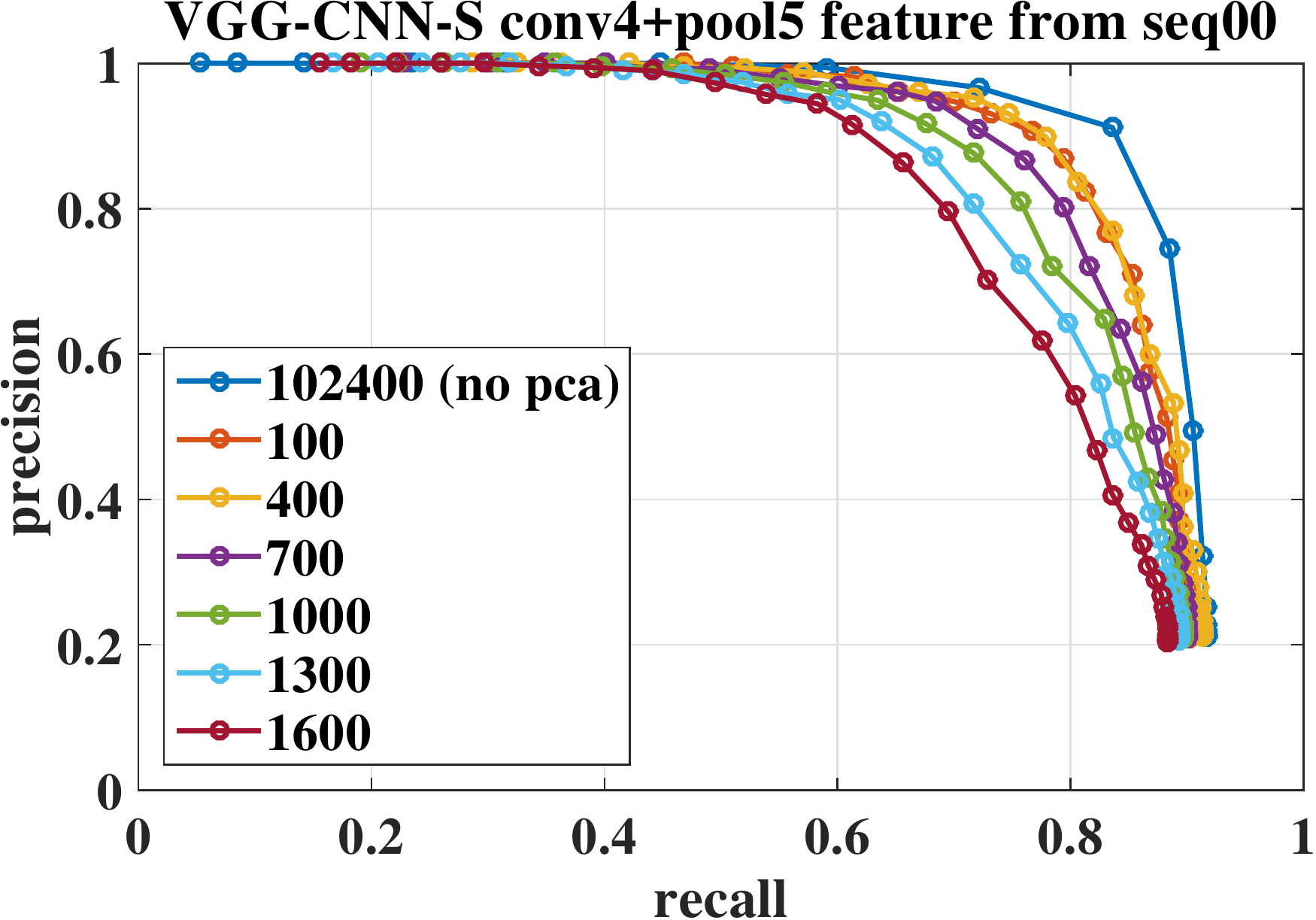}
	\includegraphics[width=0.32\textwidth,height=0.2\textwidth]{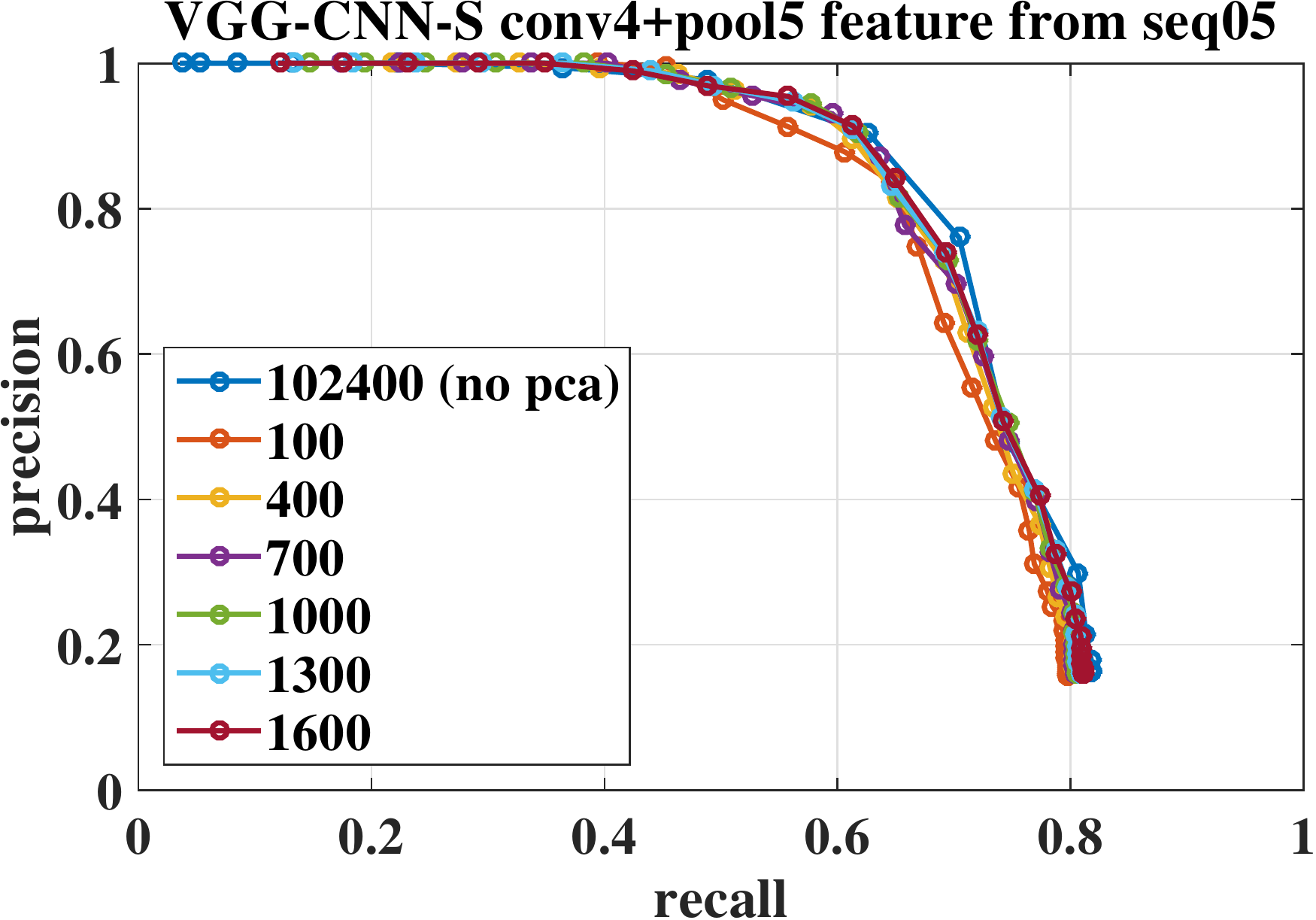}
	\includegraphics[width=0.32\textwidth,height=0.2\textwidth]{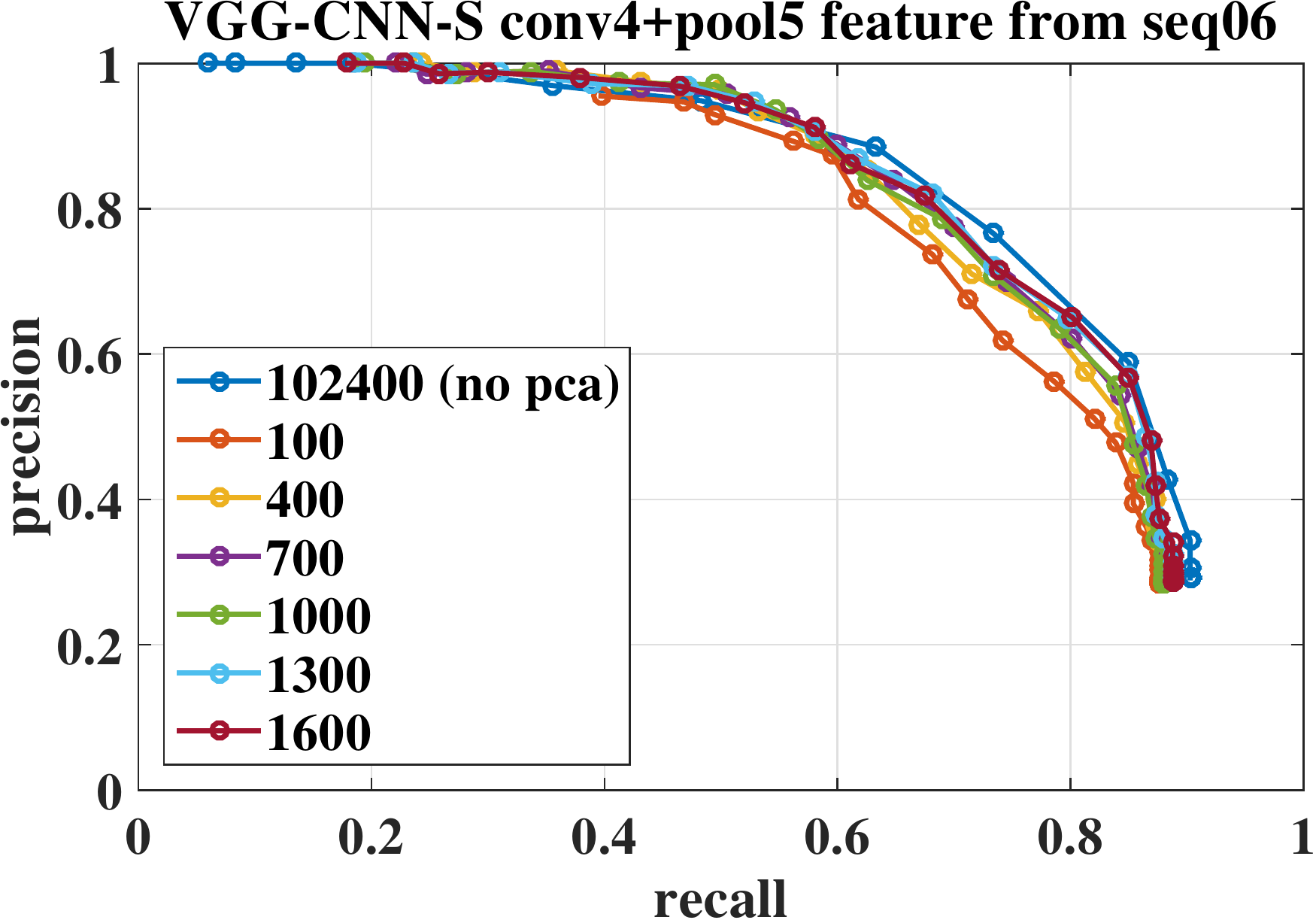}
	\includegraphics[width=0.32\textwidth,height=0.2\textwidth]{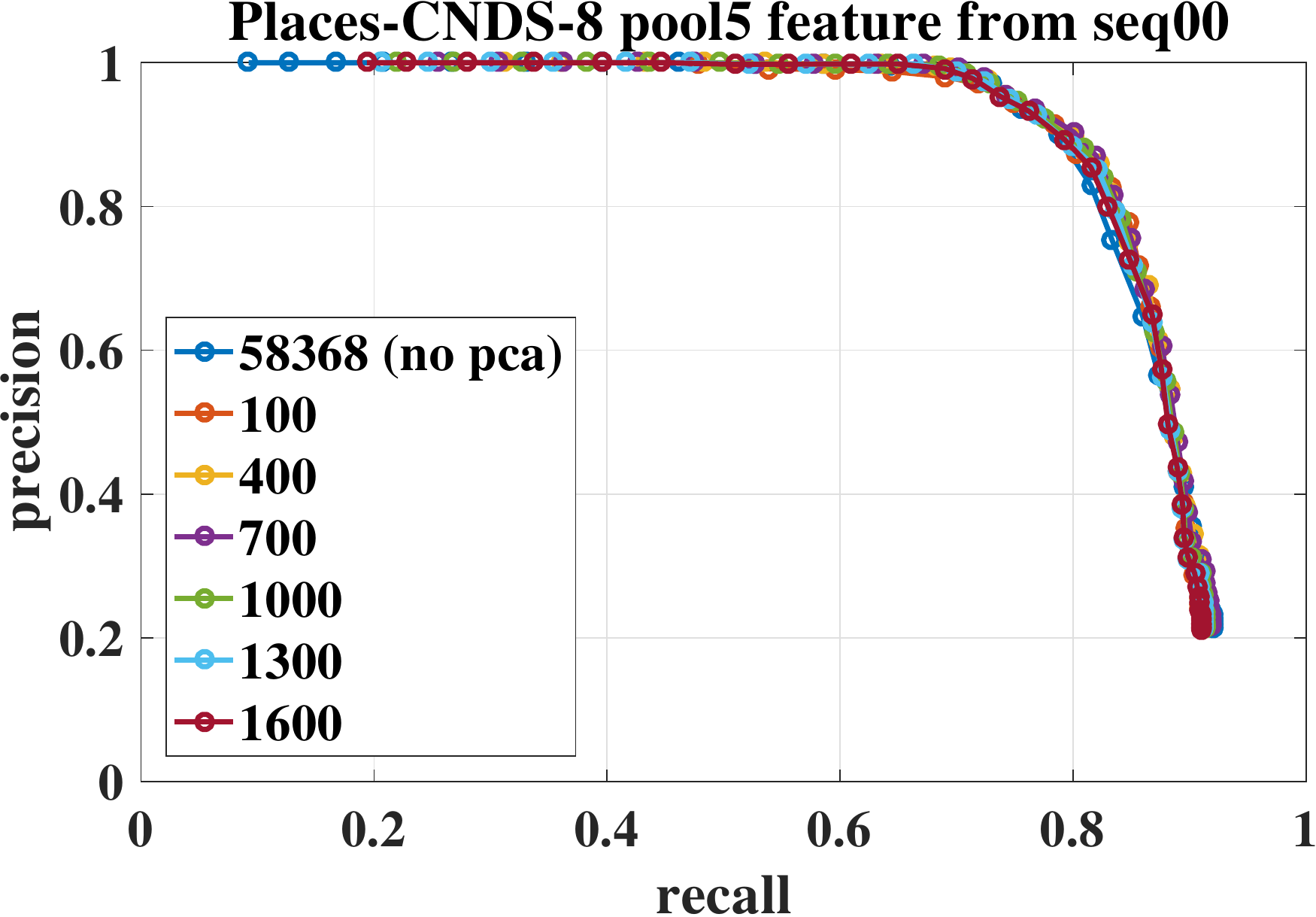}
	\includegraphics[width=0.32\textwidth,height=0.2\textwidth]{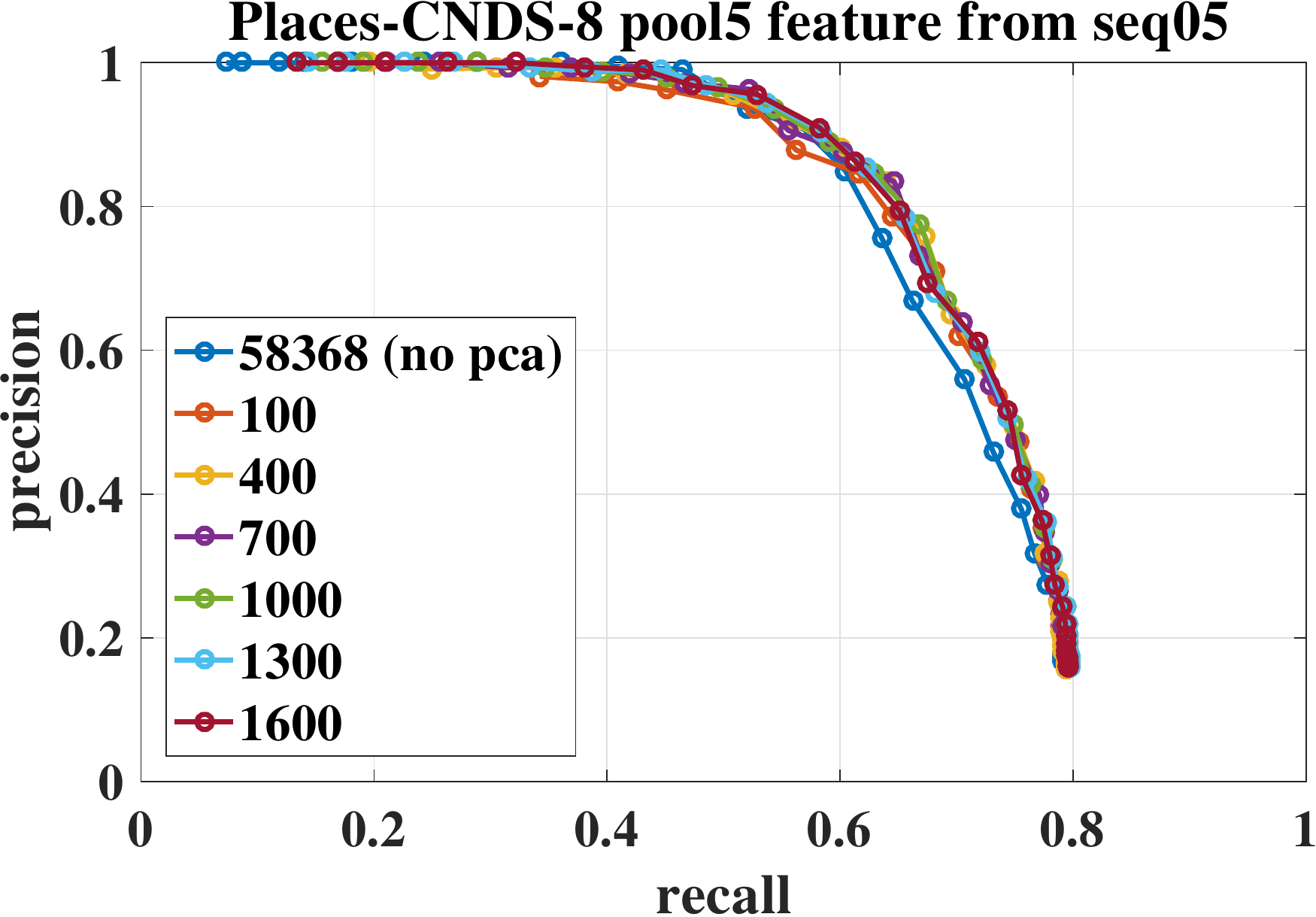}
	\includegraphics[width=0.32\textwidth,height=0.2\textwidth]{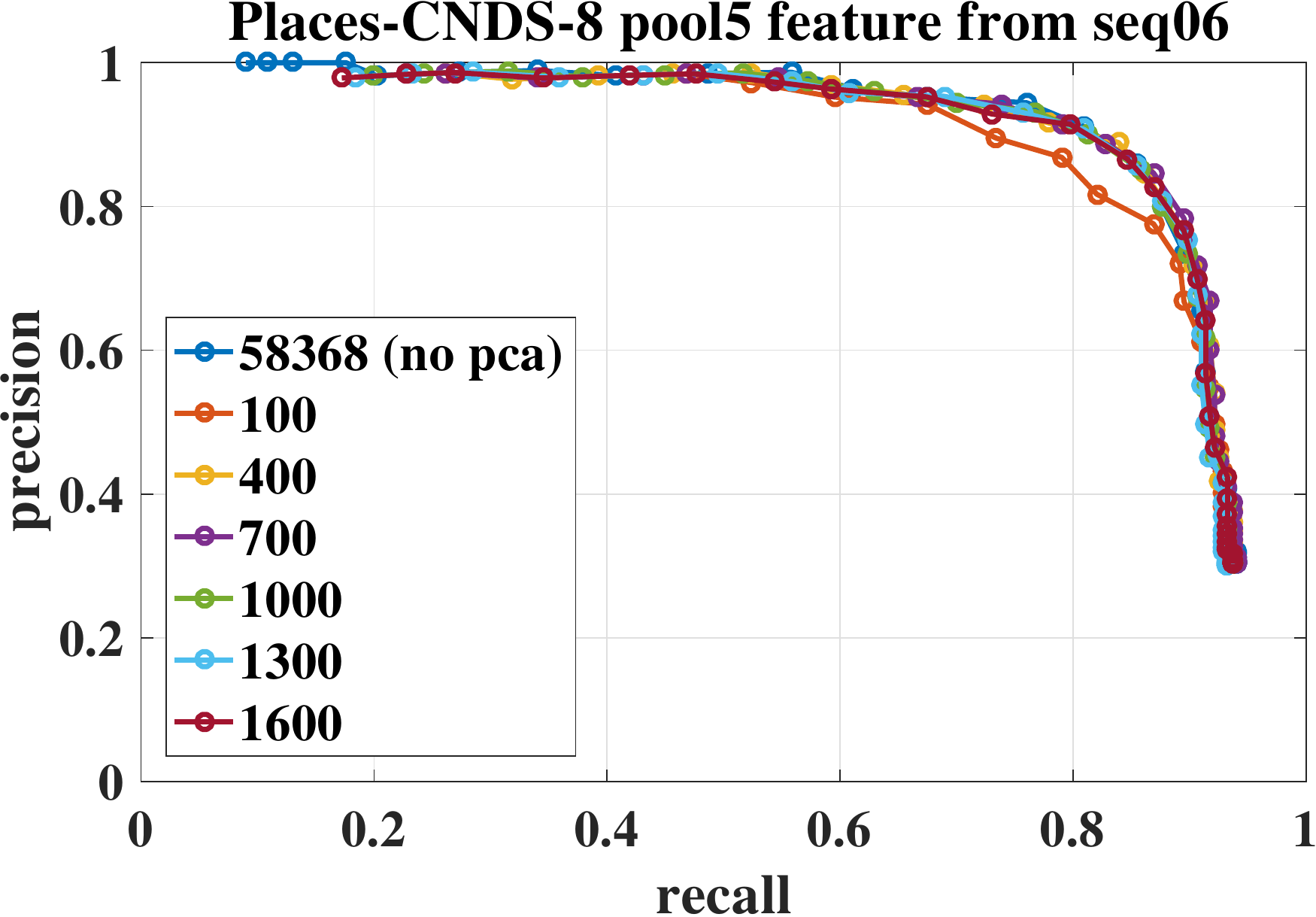}
	\caption{The precision-vs-recall performance with different remaining dimensions after PCA.  The original feature dimension and the remaining dimensions after PCA are listed in the legend.}
	\label{fig:pca}
\end{figure*}

In place recognition, each place is expected to have one unique and discriminative descriptor, so the variance of the feature indicates its discrimination ability.  Thus, in our system, PCA is adopted as a postprocessing step to obtain more compact features.  The precision-vs-recall performance with different numbers of remaining dimensions is shown in Figure~\ref{fig:pca} (better viewed in color).  It can be seen that in most of the cases, the curves are quite close to each other,  and the performance remains approximately the same even when the dimension is reduced from around 100,000 to a few hundred.  This result is consistent with our analysis, i.e. 1) the feature extracted by a pre-trained CNN has redundancy; and 2) after PCA dimension reduction, the main useful information of the features remains.  Notice that the lower dimensional features can outperform the higher dimensional ones.  This means that the PCA actually denoises the raw CNN features, e.g. there may be some cars on the road in a few scans, but in most of the scans the road region is flat and the variance corresponding to these elements is small, so that they are pruned during dimension reduction, eliminating the disturbance of the cars at the same time.

\subsection{Place recognition in bidirectional loop closure}
\label{subsec:bidirectional}
\begin{figure}
	\centering
	\captionsetup{justification=centering}
	\includegraphics[width=0.45\textwidth]{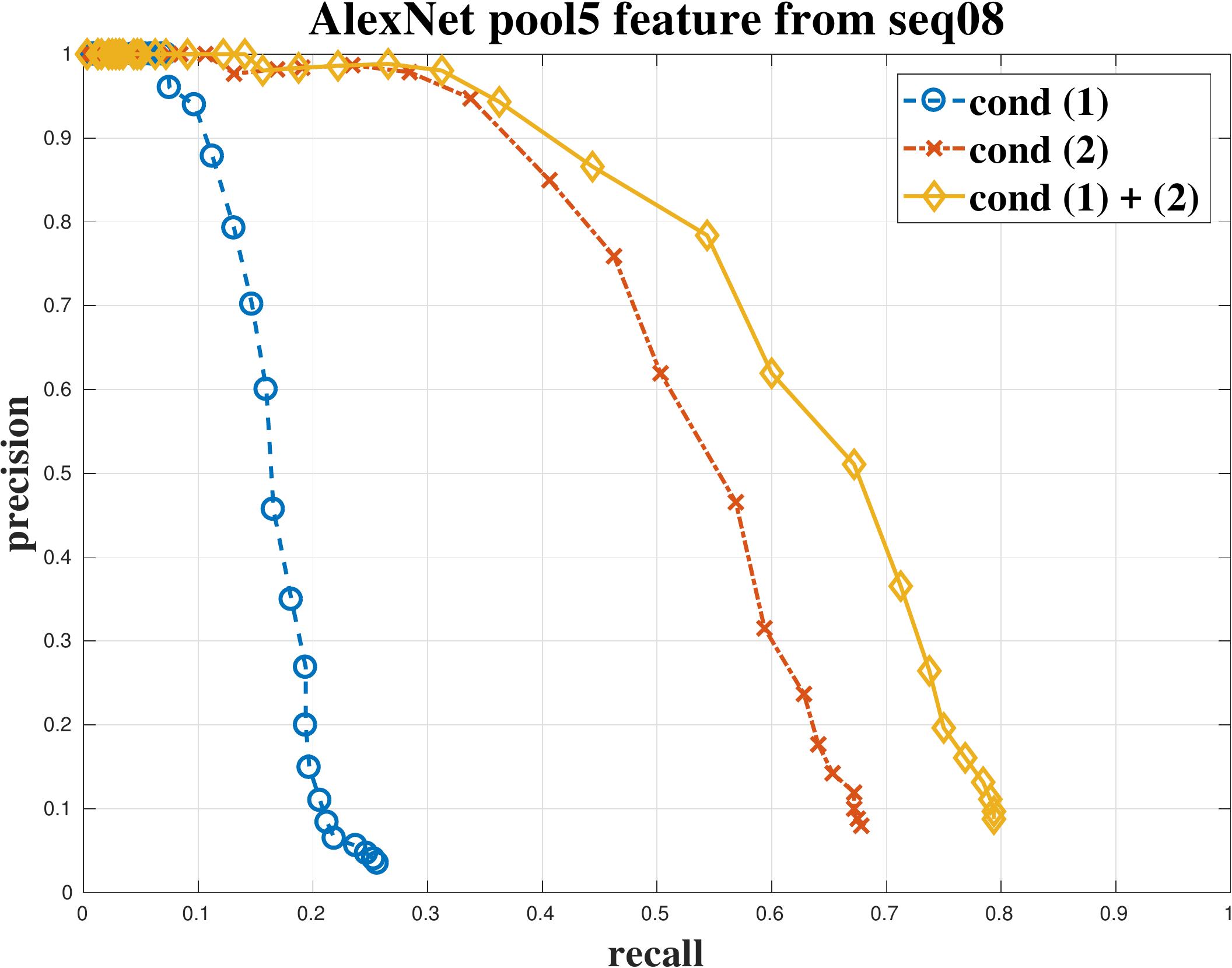}
	\caption{The precision-vs-recall performance with different PCA alignments.}
	\label{fig:bidirection}
\end{figure}

In this subsection, we test our system on sequence 08, which mainly contains bidirectional loop closures.  We tried using condition (\ref{equ:cond1}) or (\ref{equ:cond2}) alone, and considering both then choosing the match with the higher score.  The results are shown in Figure~\ref{fig:bidirection}.  It can be seen that only using one alignment case fails in this sequence, but combining both solves the problem.  We suggest that if the prior knowledge of only the unidirectional loop closure existing is available, we can use either (\ref{equ:cond1}) or (\ref{equ:cond2}) alone to generate one descriptor for each frame, otherwise consider using both, which doubles the computation.

\subsection{Comparison with other methods}
\begin{figure*}
	\centering
	\includegraphics[width=0.32\textwidth,height=0.2\textwidth]{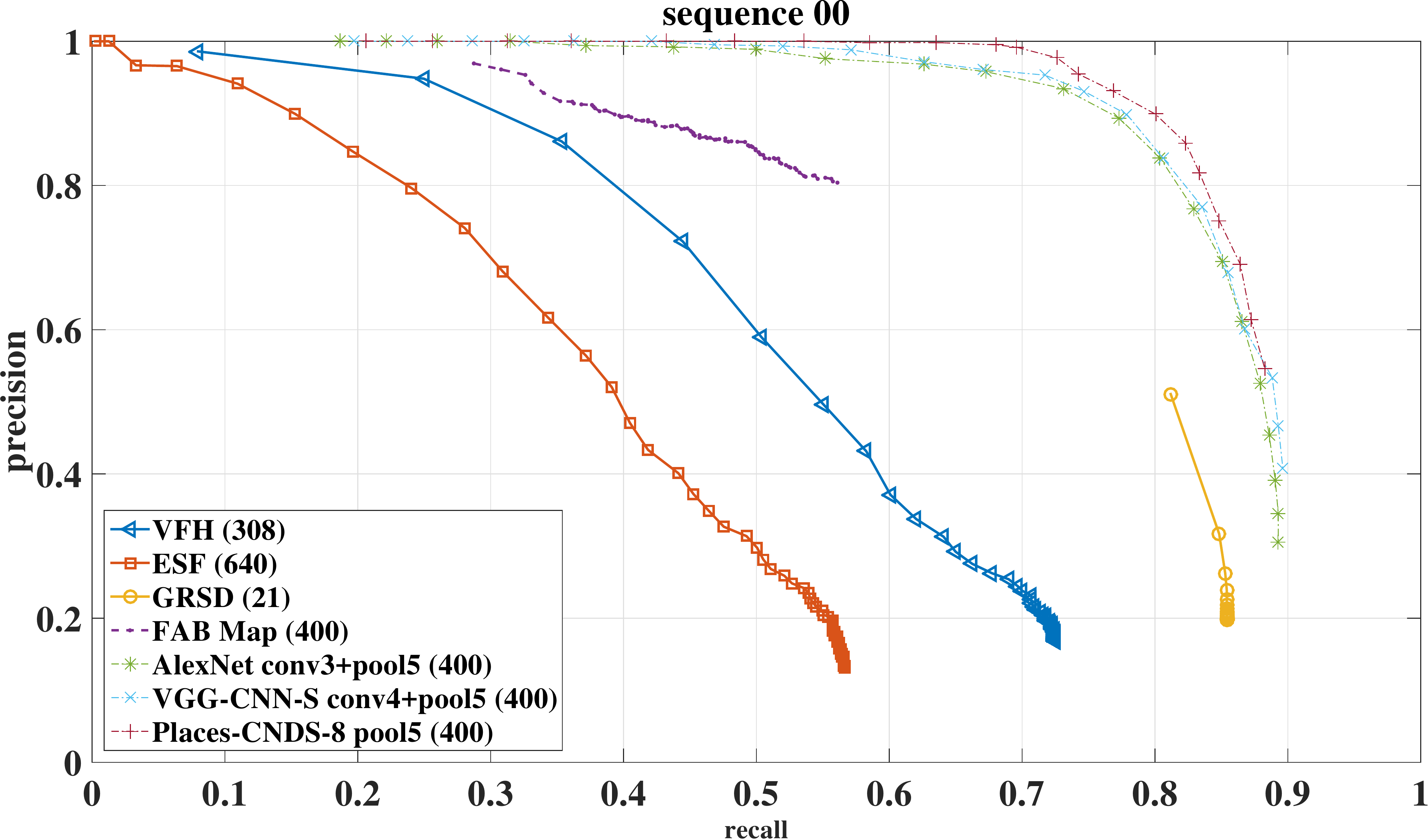}
	\includegraphics[width=0.32\textwidth,height=0.2\textwidth]{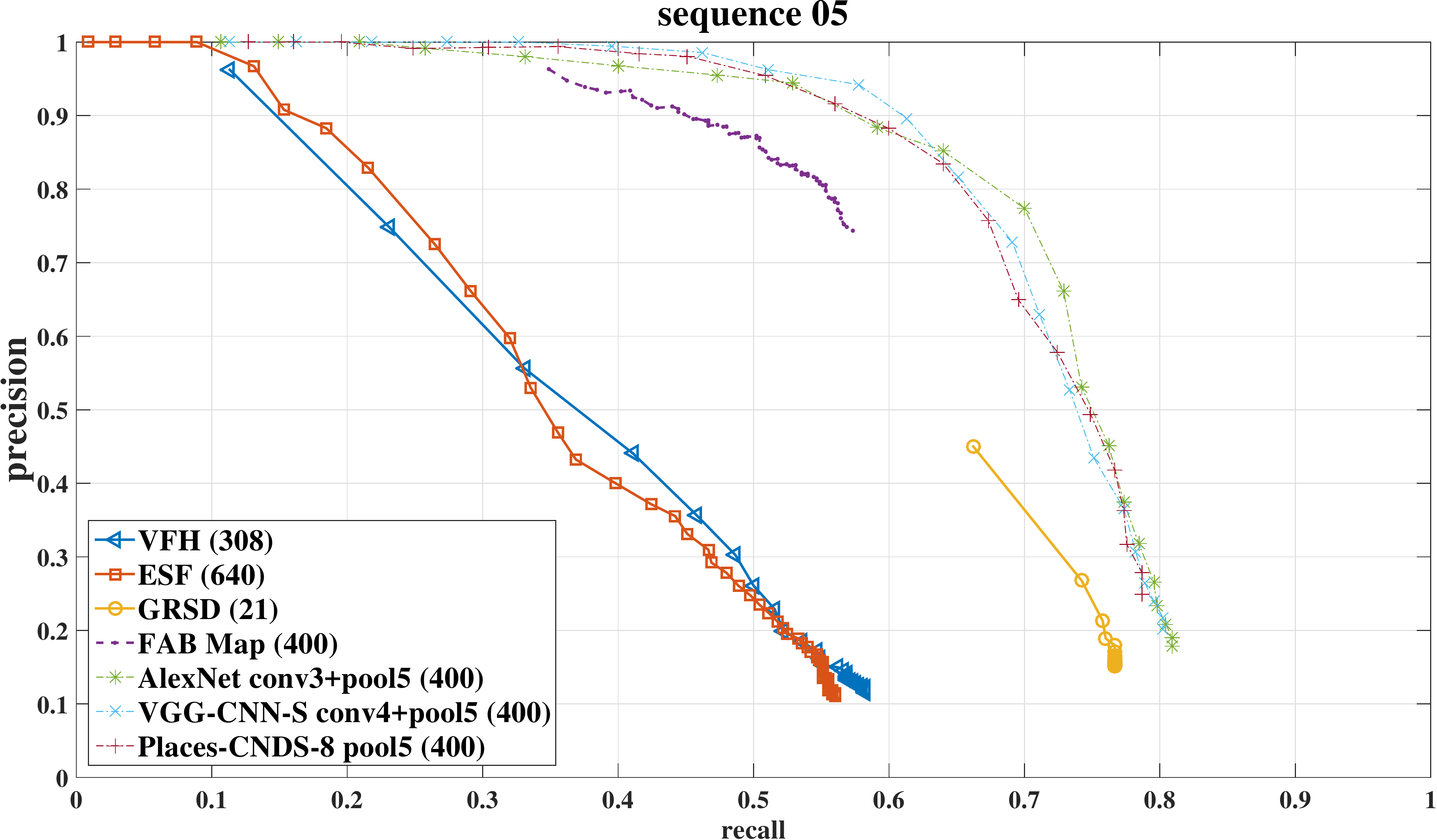}
	\includegraphics[width=0.32\textwidth,height=0.2\textwidth]{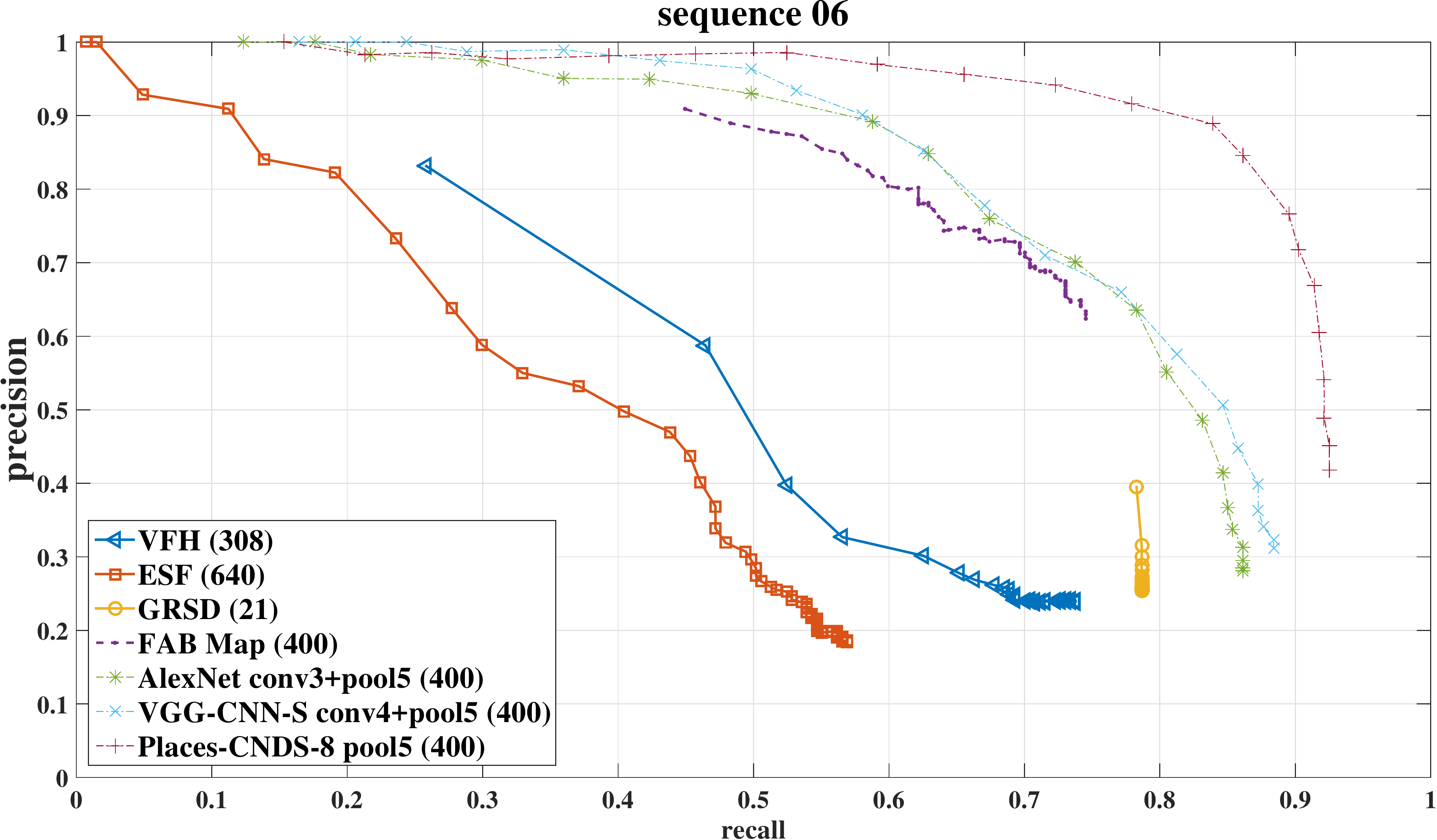}
	\caption{The precision-vs-recall performance with different methods.  The feature dimension of each method is annotated in the brackets.}
	\label{fig:compare}
\end{figure*}

Our proposed method is compared with the famous FAB-MAP \cite{FAB-MAP}, and 3 hand-crafted global features of point clouds: Viewpoint Feature Histogram (VFH) \cite{VFH}, Ensemble of Shape Functions (ESF) \cite{ESF} and Global Radius-based Surface Descriptor (GRSD) \cite{GRSD}.  We use the openFABMAP \cite{openfabmap} implementation in opencv library \cite{opencv}.  For all the hand-crafted features of point clouds we adopt the implementation in the Point Cloud Library (PCL) \cite{pcl} and normalize the features, then use cosine similarity for place retrieval.  The experiment is conducted on sequence 00, 05 and 06 of the KITTI dataset \cite{kitti} under `median difficulty' task settings.  The properties of each method are summarized in Table~\ref{tab:summary}, and the precision-vs-recall performance of all the methods is shown in Figure~\ref{fig:compare}.  It can be seen from Figure~\ref{fig:compare} that our proposed CNN features significantly outperform the hand-crafted point clouds features and FAB-MAP \cite{FAB-MAP} with comparable speed.  In Table~\ref{tab:summary}, notice that both FAB-MAP \cite{FAB-MAP} and our proposed method need training data.  However, FAB-MAP \cite{FAB-MAP} needs the testing data to be similar to the training data, which is expensive to collect for robotic application, while our method leverages the transfer learning ability of CNN, i.e. using the models pre-trained on abundant RGB images to extract features from a point cloud.

\begin{table*}[h]
	\begin{center}
		{\small
			\begin{tabular}{|p{5cm}|p{1.4cm}|p{1cm}|p{5cm}|p{1.5cm}|} 
				\hline
				\multicolumn{1}{|c|}{method} & feature dimension & need training data & parameters (if any)                                              & time (frames/sec) \\
				\hline\hline
				VFH \cite{VFH}              &  308              & No                 & normal estimation search radius = 0.03                           & 4.8 \\
				ESF \cite{ESF}              &  640              & No                 & --                                                               & 18.0 \\
				GRSD \cite{GRSD}            &  21               & No                 & normal estimation search radius = 0.05, GRSD search radius = 0.1 & 1.6 \\
				FAB-MAP \cite{FAB-MAP}      &  400              & Yes                & feature: SIFT, vocabulary size = 400                             & 8.7 \\
				proposed                     &  400              & Yes                & PCA remaining dimension = 400                                    & 2.0 (CPU), 5.4 (GPU) \\
				\hline
			\end{tabular}
		}
	\end{center}
	\caption{Summary of the the properties of the following methods: Viewpoint Feature Histogram (VFH) \cite{VFH}, Ensemble of Shape Functions (ESF) \cite{ESF}, Global Radius-based Surface Descriptor (GRSD) \cite{GRSD}, FAB-MAP \cite{FAB-MAP}, and our proposed method.  The GPU time of our proposed method is obtained by using the GPU version of Caffe \cite{Caffe} for CNN feature extraction.}
	\label{tab:summary}
\end{table*}

\subsection{Test on the HKUST dataset}
\begin{figure*}
	\centering
	\includegraphics[width=0.32\textwidth,height=0.2\textwidth]{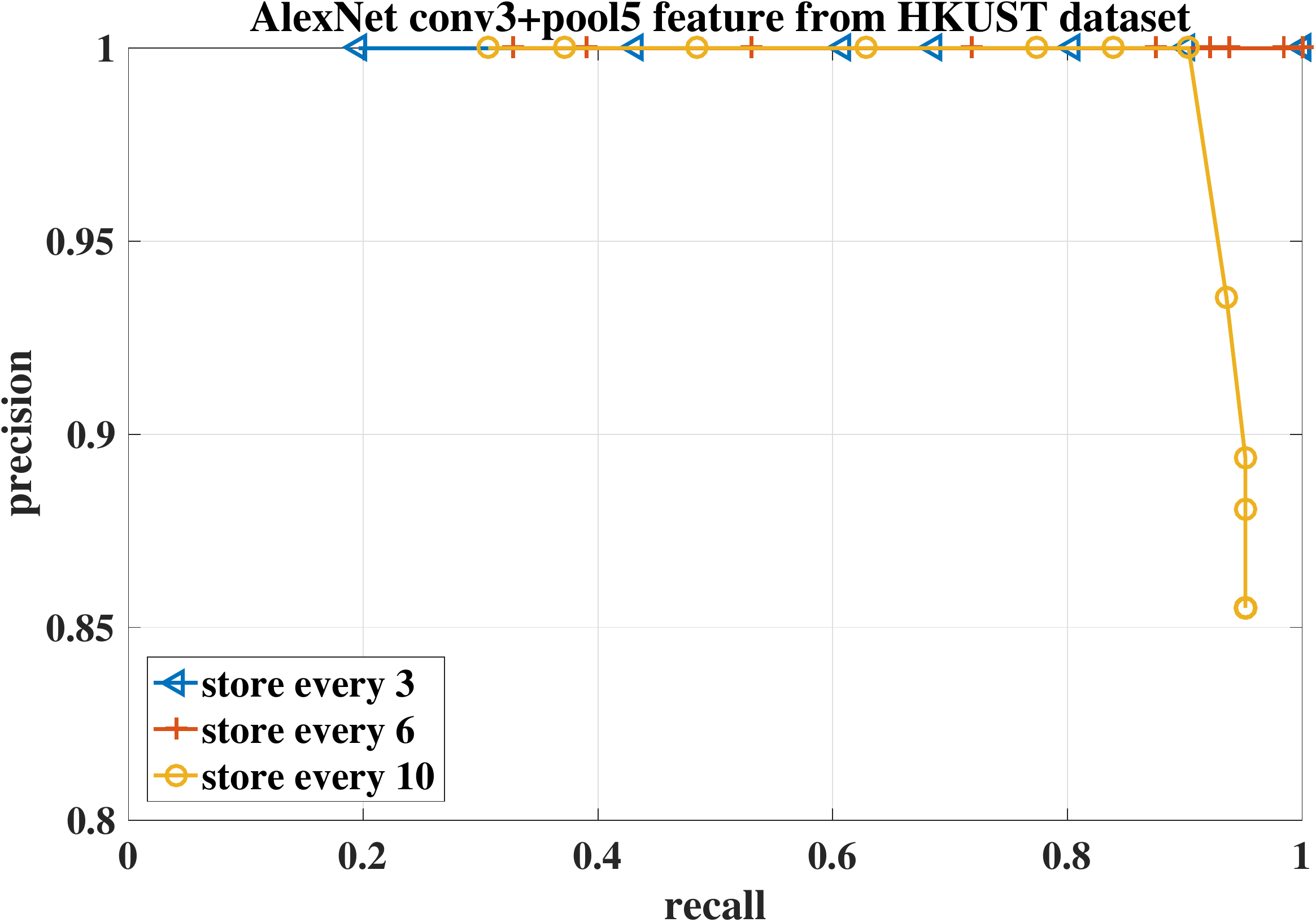}
	\includegraphics[width=0.32\textwidth,height=0.2\textwidth]{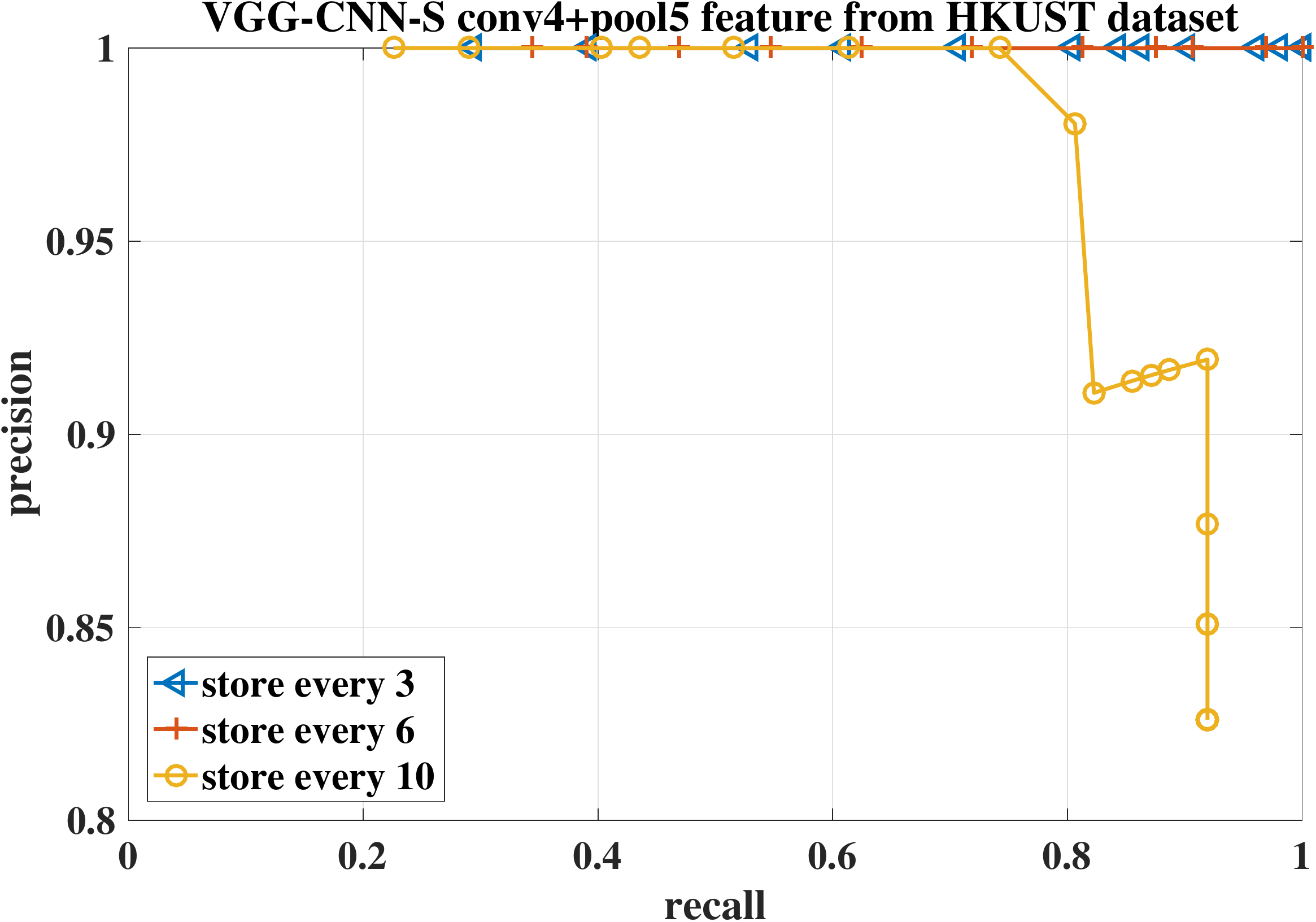}
	\includegraphics[width=0.32\textwidth,height=0.2\textwidth]{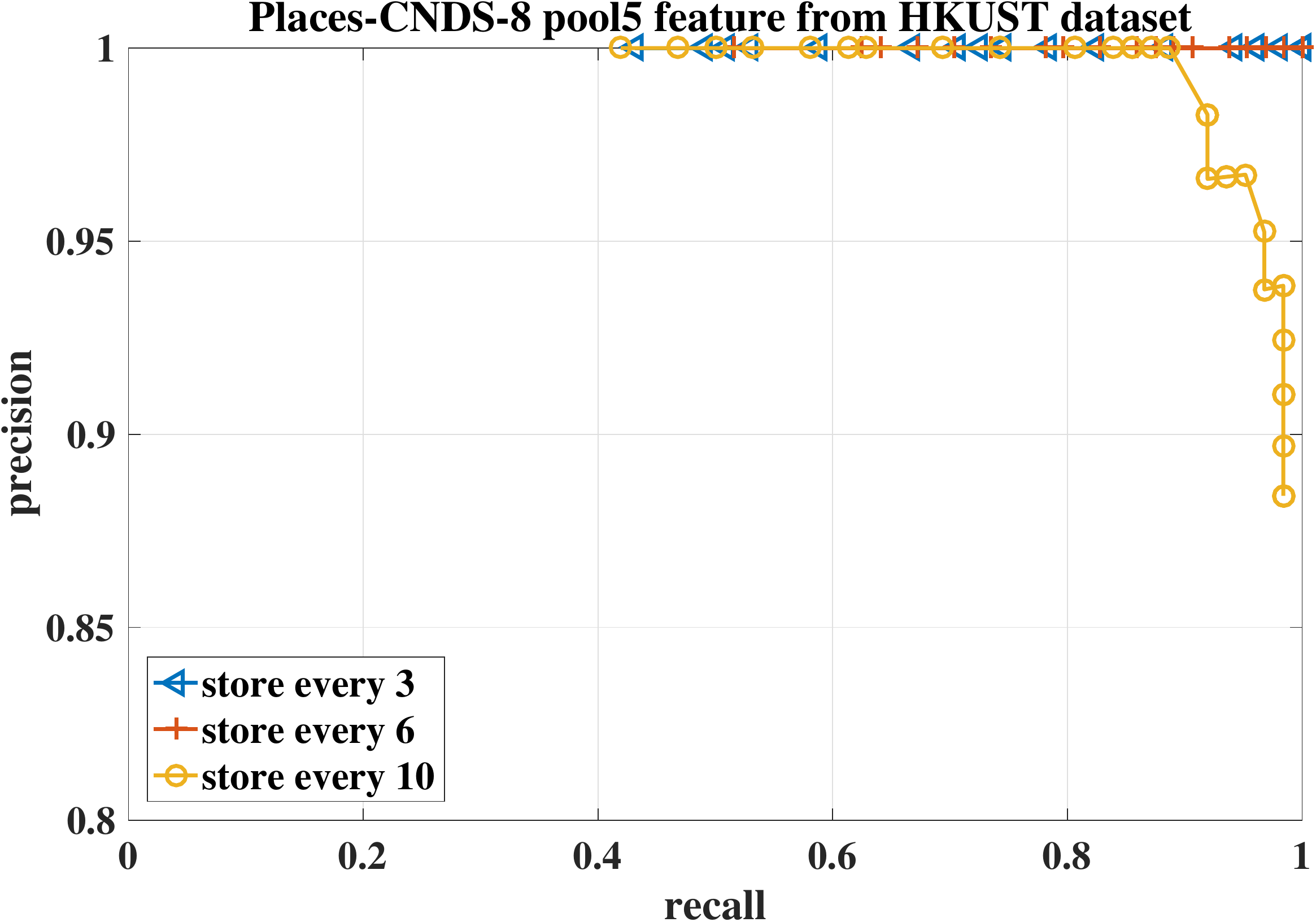}
	\caption{The precision-vs-recall performance on the rotation invariance testing dataset of HKUST.}
	\label{fig:hkust_rotation}
\end{figure*}

\begin{figure*}
	\centering
	\includegraphics[width=0.32\textwidth,height=0.2\textwidth]{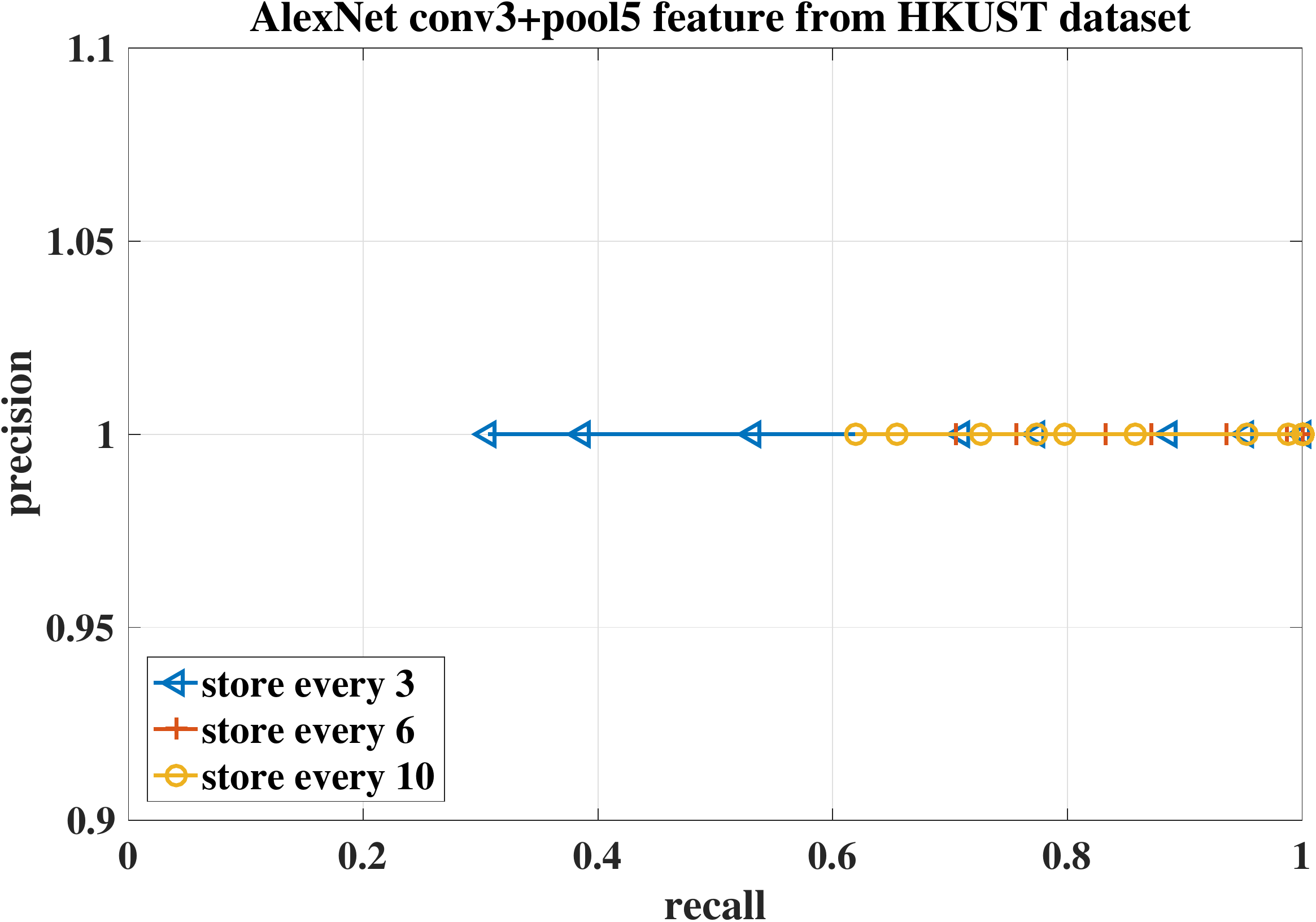}
	\includegraphics[width=0.32\textwidth,height=0.2\textwidth]{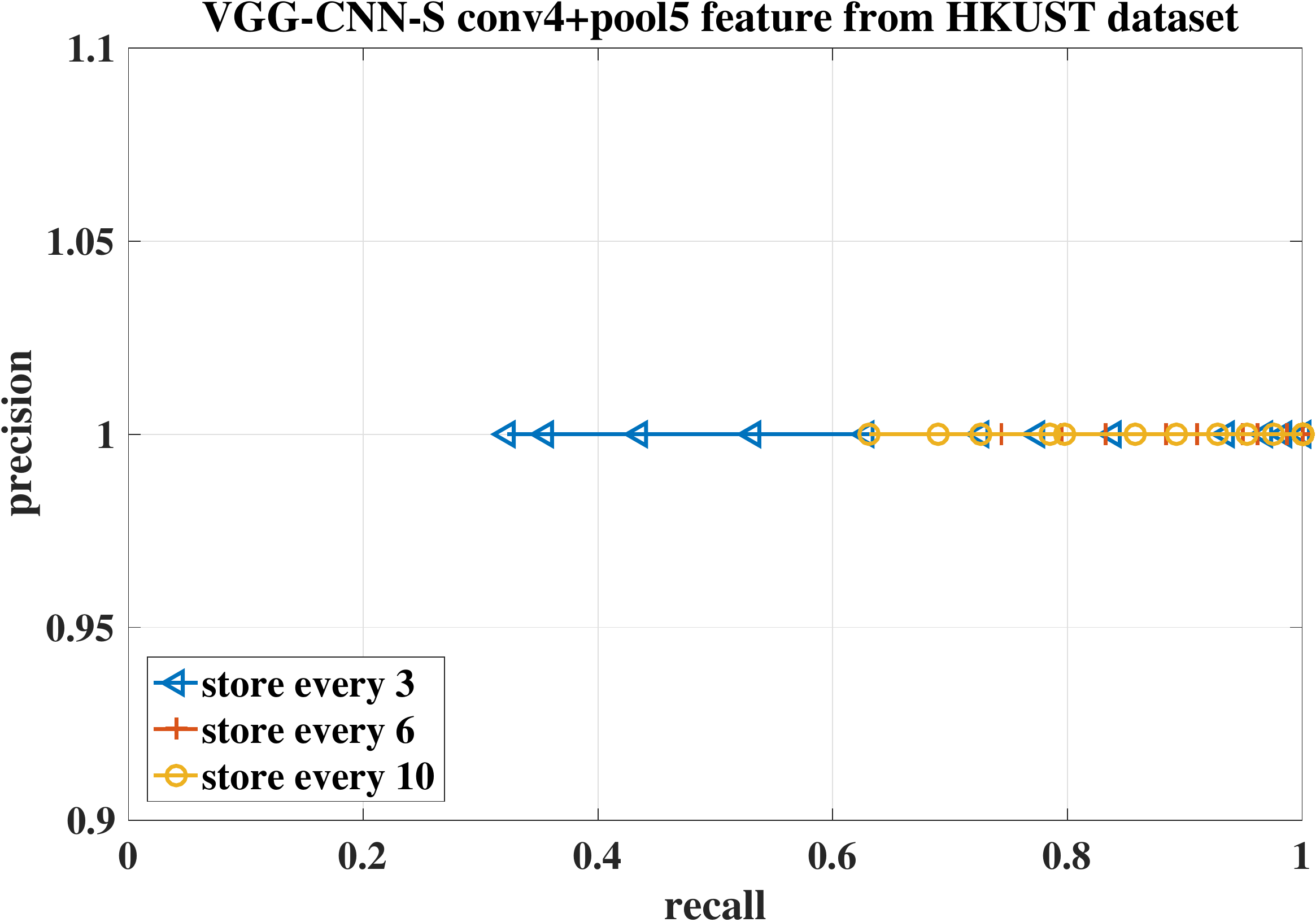}
	\includegraphics[width=0.32\textwidth,height=0.2\textwidth]{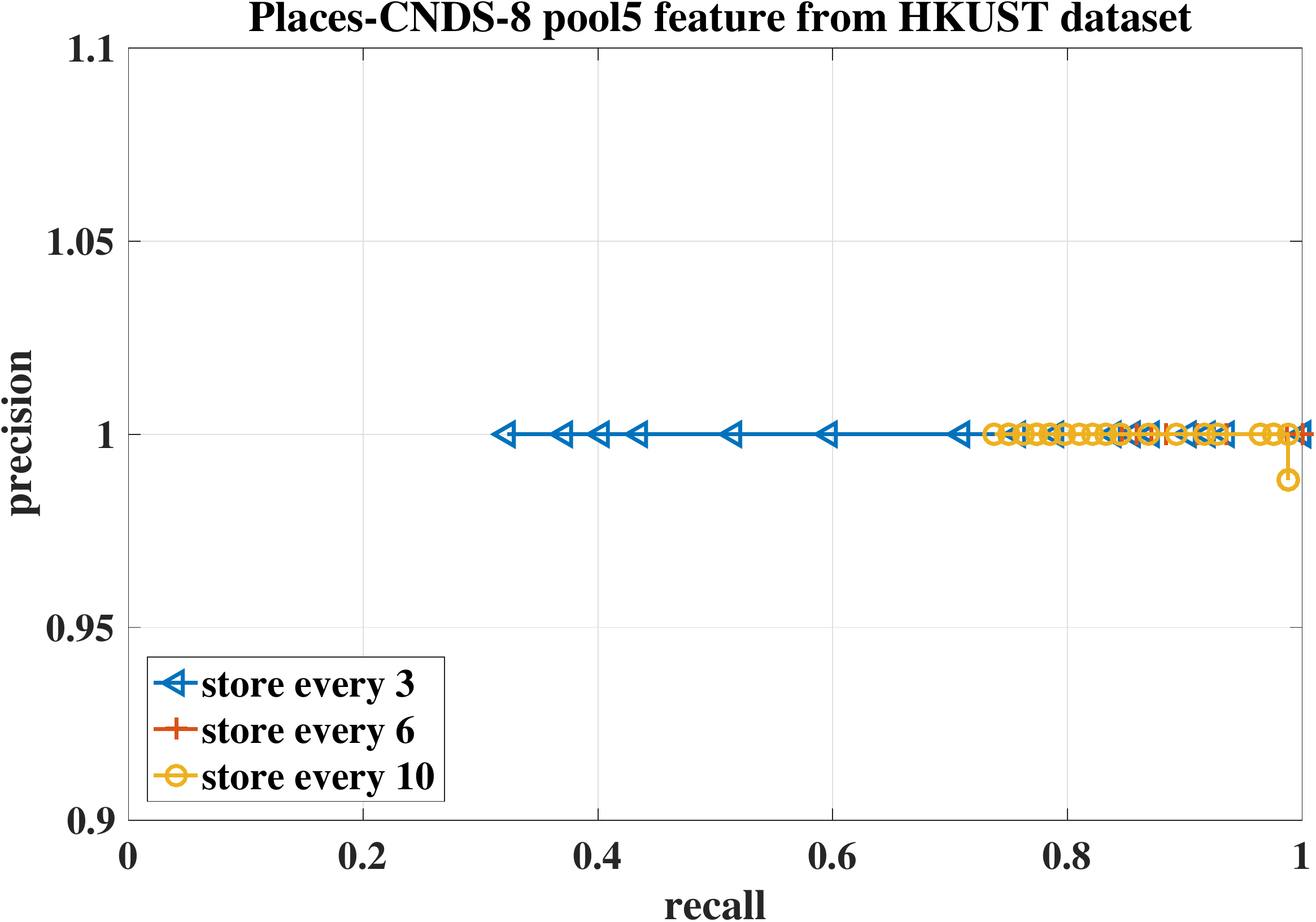}
	\caption{The precision-vs-recall performance on the robustness to unrelated small objects testing dataset of HKUST.}
	\label{fig:hkust_object}
\end{figure*}

As mentioned in Subsec.~\ref{subsec:dataset}, we collect two sets of data to test rotation invariance and robustness to small objects that are unrelated to the place identity.  We conduct place recognition on the two sets of data separately.  In each experiment, we sample 1/3, 1/6 and 1/10 of the scans as a stored map, and leave the rest as queries.  Since every one among the 7 locations has at least one scan stored, true matches exist for every query.  More stored scans means the query can find more similar matches, resulting in an easier task.  We collect about 10 scans at each location, so storing every 10\textit{th} scan means that on average, each place has one scan stored, and all the remaining scans should be able to match it.  The precision-vs-recall performance of rotation invariant testing and robustness to small unrelated objects are shown in Figure~\ref{fig:hkust_rotation} and Figure~\ref{fig:hkust_object}, respectively.  It can be seen from Figure~\ref{fig:hkust_rotation} that our system achieves very good performance under rotation (notice that the precision/vertical axis starts from 0.8), which is due to the PCA alignment and the spacial-invariant property of the CNN features.  Figure~\ref{fig:hkust_object} shows that our system can always recognize the correct place (notice that the precision/vertical axis starts form 0.9), which indicates that small unrelated objects basically do not affect our place recognition system.  For the sensor that covers a full $360^\circ$ environmental view, the proportion occupied by the unrelated objects like pedestrian and cars is very small, in other words, these sensors are unlikely to be severely blocked, which results in a robust system.

\section{Conclusion}
\label{sec:Conclusion}
In this paper, we proposed a novel point-cloud-based place recognition system leveraging CNN feature extraction.  Our method bridges the gap between powerful image-based deep learning approaches and point-cloud recognition.  Without using any range images for training, the CNN features obtained in our system significantly outperform hand-crafted features and the resultant system is illumination invariant, rotation invariant and robust to unrelated small moving objects.  We also introduce a new place recognition dataset containing both point cloud and grayscale images.  Both types of data cover a full $360^\circ$ environmental view, and the content of the dataset is organized to especially facilitate tests of rotation invariance and robustness to unrelated (moving) objects separately.

\ifCLASSOPTIONcaptionsoff
  \newpage
\fi



\begin{thebibliography}{}
\providecommand{\url}[1]{#1}
\csname url@rmstyle\endcsname
\providecommand{\newblock}{\relax}
\providecommand{\bibinfo}[2]{#2}
\providecommand\BIBentrySTDinterwordspacing{\spaceskip=0pt\relax}
\providecommand\BIBentryALTinterwordstretchfactor{4}
\providecommand\BIBentryALTinterwordspacing{\spaceskip=\fontdimen2\font plus
\BIBentryALTinterwordstretchfactor\fontdimen3\font minus
  \fontdimen4\font\relax}
\providecommand\BIBforeignlanguage[2]{{%
\expandafter\ifx\csname l@#1\endcsname\relax
\typeout{** WARNING: IEEEtran.bst: No hyphenation pattern has been}%
\typeout{** loaded for the language `#1'. Using the pattern for}%
\typeout{** the default language instead.}%
\else
\language=\csname l@#1\endcsname
\fi
#2}}

\end{thebibliography}


\begin{thebibliography}{10}
	\providecommand{\url}[1]{#1}
	\csname url@rmstyle\endcsname
	\providecommand{\newblock}{\relax}
	\providecommand{\bibinfo}[2]{#2}
	\providecommand\BIBentrySTDinterwordspacing{\spaceskip=0pt\relax}
	\providecommand\BIBentryALTinterwordstretchfactor{4}
	\providecommand\BIBentryALTinterwordspacing{\spaceskip=\fontdimen2\font plus
		\BIBentryALTinterwordstretchfactor\fontdimen3\font minus
		\fontdimen4\font\relax}
	\providecommand\BIBforeignlanguage[2]{{%
			\expandafter\ifx\csname l@#1\endcsname\relax
			\typeout{** WARNING: IEEEtran.bst: No hyphenation pattern has been}%
			\typeout{** loaded for the language `#1'. Using the pattern for}%
			\typeout{** the default language instead.}%
			\else
			\language=\csname l@#1\endcsname
			\fi
			#2}}
	
	\bibitem{scene-sequences}
	K.~L. Ho and P.~Newman, ``Detecting loop closure with scene sequences,''
	\emph{International Journal of Computer Vision}, vol.~74, no.~3, pp.
	261--286, 2007.
	
	\bibitem{FAB-MAP}
	M.~Cummins and P.~Newman, ``Fab-map: Probabilistic localization and mapping in
	the space of appearance,'' \emph{The International Journal of Robotics
		Research}, vol.~27, no.~6, pp. 647--665, 2008.
	
	\bibitem{SeqSLAM}
	M.~J. Milford and G.~F. Wyeth, ``Seqslam: Visual route-based navigation for
	sunny summer days and stormy winter nights,'' in \emph{proceedings of IEEE
		International Conference on Robotics and Automation}.\hskip 1em plus 0.5em
	minus 0.4em\relax IEEE, 2012, pp. 1643--1649.
	
	\bibitem{correct-loop}
	Y.~Latif, C.~Cadena, and J.~Neira, ``Robust loop closing over time for pose
	graph slam,'' \emph{The International Journal of Robotics Research}, vol.~32,
	no.~14, pp. 1611--1626, 2013.
	
	\bibitem{HMM}
	P.~Hansen and B.~Browning, ``Visual place recognition using hmm sequence
	matching,'' in \emph{proceedings of IEEE/RSJ International Conference on
		Intelligent Robots and Systems}.\hskip 1em plus 0.5em minus 0.4em\relax IEEE,
	2014, pp. 4549--4555.
	
	\bibitem{binary-sequences}
	R.~Arroyo, P.~F. Alcantarilla, L.~M. Bergasa, and E.~Romera, ``Towards
	life-long visual localization using an efficient matching of binary sequences
	from images,'' in \emph{proceedings of IEEE International Conference on
		Robotics and Automation}.\hskip 1em plus 0.5em minus 0.4em\relax IEEE, 2015,
	pp. 6328--6335.
	
	\bibitem{slam-across-seasons}
	T.~Naseer, M.~Ruhnke, C.~Stachniss, L.~Spinello, and W.~Burgard, ``Robust
	visual slam across seasons,'' in \emph{proceedings of IEEE/RSJ International
		Conference on Intelligent Robots and Systems}.\hskip 1em plus 0.5em minus
	0.4em\relax IEEE, 2015, pp. 2529--2535.
	
	\bibitem{location_models}
	E.~S. Stumm, C.~Mei, and S.~Lacroix, ``Building location models for visual
	place recognition,'' \emph{The International Journal of Robotics Research},
	vol.~35, no.~4, pp. 334--356, 2016.
	
	\bibitem{covisibility-graph}
	S.~Cascianelli, G.~Costante, E.~Bellocchio, P.~Valigi, M.~L. Fravolini, and
	T.~A. Ciarfuglia, ``Robust visual semi-semantic loop closure detection by a
	covisibility graph and cnn features,'' \emph{{Robotics and Autonomous
			Systems}}, vol.~92, pp. 53--65, 2017.
	
	\bibitem{3Dlidar}
	M.~Bosse and R.~Zlot, ``Place recognition using keypoint voting in large 3d
	lidar datasets,'' in \emph{proceedings of IEEE International Conference on
		Robotics and Automation}.\hskip 1em plus 0.5em minus 0.4em\relax IEEE, 2013,
	pp. 2677--2684.
	
	\bibitem{zhang2010lidar}
	W.~Zhang, ``Lidar-based road and road-edge detection,'' in \emph{{Intelligent
			Vehicles Symposium (IV), 2010 IEEE}}.\hskip 1em plus 0.5em minus 0.4em\relax
	IEEE, 2010, pp. 845--848.
	
	\bibitem{zhu20123d}
	Q.~Zhu, L.~Chen, Q.~Li, M.~Li, A.~N{\"u}chter, and J.~Wang, ``3d lidar point
	cloud based intersection recognition for autonomous driving,'' in
	\emph{{Intelligent Vehicles Symposium (IV), 2012 IEEE}}.\hskip 1em plus 0.5em
	minus 0.4em\relax IEEE, 2012, pp. 456--461.
	
	\bibitem{kitti}
	A.~Geiger, P.~Lenz, and R.~Urtasun, ``Are we ready for autonomous driving? the
	kitti vision benchmark suite,'' in \emph{Proceedings of the IEEE Computer
		Society Conference on Computer Vision and Pattern Recognition}.\hskip 1em
	plus 0.5em minus 0.4em\relax IEEE, 2012, pp. 3354--3361.
	
	\bibitem{ibisch2013towards}
	A.~Ibisch, S.~St{\"u}mper, H.~Altinger, M.~Neuhausen, M.~Tschentscher,
	M.~Schlipsing, J.~Salinen, and A.~Knoll, ``Towards autonomous driving in a
	parking garage: Vehicle localization and tracking using environment-embedded
	lidar sensors,'' in \emph{{Intelligent Vehicles Symposium (IV), 2013
			IEEE}}.\hskip 1em plus 0.5em minus 0.4em\relax IEEE, 2013, pp. 829--834.
	
	\bibitem{SIFT}
	D.~G. Lowe, ``Object recognition from local scale-invariant features,'' in
	\emph{Proceedings of the International Conference on Computer Vision},
	vol.~2.\hskip 1em plus 0.5em minus 0.4em\relax Ieee, 1999, pp. 1150--1157.
	
	\bibitem{steder2011place}
	B.~Steder, M.~Ruhnke, S.~Grzonka, and W.~Burgard, ``Place recognition in 3d
	scans using a combination of bag of words and point feature based relative
	pose estimation,'' in \emph{proceedings of IEEE/RSJ International Conference
		on Intelligent Robots and Systems}.\hskip 1em plus 0.5em minus 0.4em\relax
	IEEE, 2011, pp. 1249--1255.
	
	\bibitem{Imagenet}
	A.~Krizhevsky, I.~Sutskever, and G.~E. Hinton, ``Imagenet classification with
	deep convolutional neural networks,'' in \emph{Advances in neural information
		processing systems}, 2012, pp. 1097--1105.
	
	\bibitem{DeCAF}
	J.~Donahue, Y.~Jia, O.~Vinyals, J.~Hoffman, N.~Zhang, E.~Tzeng, and T.~Darrell,
	``Decaf: A deep convolutional activation feature for generic visual
	recognition,'' in \emph{International conference on machine learning}, 2014,
	pp. 647--655.
	
	\bibitem{cnn-feature-off-the-shelf}
	A.~Sharif~Razavian, H.~Azizpour, J.~Sullivan, and S.~Carlsson, ``Cnn features
	off-the-shelf: an astounding baseline for recognition,'' in \emph{Proceedings
		of the IEEE conference on computer vision and pattern recognition workshops},
	2014, pp. 806--813.
	
	\bibitem{very-deep}
	K.~Simonyan and A.~Zisserman, ``Very deep convolutional networks for
	large-scale image recognition,'' \emph{arXiv preprint arXiv:1409.1556}, 2014.
	
	\bibitem{going-deeper}
	C.~Szegedy, W.~Liu, Y.~Jia, P.~Sermanet, S.~Reed, D.~Anguelov, D.~Erhan,
	V.~Vanhoucke, and A.~Rabinovich, ``Going deeper with convolutions,'' in
	\emph{Proceedings of the IEEE Computer Society Conference on Computer Vision
		and Pattern Recognition}, 2015, pp. 1--9.
	
	\bibitem{convnet}
	Y.~LeCun, B.~E. Boser, J.~S. Denker, D.~Henderson, R.~E. Howard, W.~E. Hubbard,
	and L.~D. Jackel, ``Handwritten digit recognition with a back-propagation
	network,'' in \emph{Proceedings of Conference on Neural Information
		Processing Systems (NIPS)}, 1990, pp. 396--404.
	
	\bibitem{tai2016pca}
	L.~Tai, Q.~Ye, and M.~Liu, ``Pca-aided fully convolutional networks for
	semantic segmentation of multi-channel fmri,'' \emph{arXiv preprint
		arXiv:1610.01732}, 2016.
	
	\bibitem{sun2017improving}
	Y.~Sun, M.~Liu, and M.~Q.-H. Meng, ``Improving rgb-d slam in dynamic
	environments: A motion removal approach,'' \emph{{Robotics and Autonomous
			Systems}}, vol.~89, pp. 110--122, 2017.
	
	\bibitem{survey}
	S.~Lowry, N.~S{\"u}nderhauf, P.~Newman, J.~J. Leonard, D.~Cox, P.~Corke, and
	M.~J. Milford, ``Visual place recognition: A survey,'' \emph{IEEE
		Transactions on Robotics}, vol.~32, no.~1, pp. 1--19, 2016.
	
	\bibitem{BRIEF-Gist}
	N.~S{\"u}nderhauf and P.~Protzel, ``Brief-gist-closing the loop by simple
	means,'' in \emph{proceedings of IEEE/RSJ International Conference on
		Intelligent Robots and Systems}.\hskip 1em plus 0.5em minus 0.4em\relax IEEE,
	2011, pp. 1234--1241.
	
	\bibitem{SURF}
	H.~Bay, T.~Tuytelaars, and L.~Van~Gool, ``Surf: Speeded up robust features,''
	in \emph{Proceedings of the European Conference on Computer Vision}.\hskip
	1em plus 0.5em minus 0.4em\relax Springer, 2006, pp. 404--417.
	
	\bibitem{BRIEF}
	M.~Calonder, V.~Lepetit, P.~Fua, K.~Konolige, J.~Bowman, and P.~Mihelich,
	``Compact signatures for high-speed interest point description and
	matching,'' in \emph{Computer Vision, 2009 IEEE 12th International Conference
		on}.\hskip 1em plus 0.5em minus 0.4em\relax IEEE, 2009, pp. 357--364.
	
	\bibitem{BRISK}
	S.~Leutenegger, M.~Chli, and R.~Y. Siegwart, ``Brisk: Binary robust invariant
	scalable keypoints,'' in \emph{Proceedings of the International Conference on
		Computer Vision}.\hskip 1em plus 0.5em minus 0.4em\relax IEEE, 2011, pp.
	2548--2555.
	
	\bibitem{ORB}
	E.~Rublee, V.~Rabaud, K.~Konolige, and G.~Bradski, ``Orb: An efficient
	alternative to sift or surf,'' in \emph{Proceedings of the International
		Conference on Computer Vision}.\hskip 1em plus 0.5em minus 0.4em\relax IEEE,
	2011, pp. 2564--2571.
	
	\bibitem{LDB}
	X.~Yang and K.-T. Cheng, ``Ldb: An ultra-fast feature for scalable augmented
	reality on mobile devices,'' in \emph{{International Symposium onMixed and
			Augmented Reality (ISMAR)}}.\hskip 1em plus 0.5em minus 0.4em\relax IEEE,
	2012, pp. 49--57.
	
	\bibitem{D-LDB}
	R.~Arroyo, P.~F. Alcantarilla, L.~M. Bergasa, J.~J. Yebes, and S.~Bronte,
	``Fast and effective visual place recognition using binary codes and
	disparity information,'' in \emph{proceedings of IEEE/RSJ International
		Conference on Intelligent Robots and Systems}.\hskip 1em plus 0.5em minus
	0.4em\relax IEEE, 2014, pp. 3089--3094.
	
	\bibitem{WI-SURF}
	H.~Badino, D.~Huber, and T.~Kanade, ``Real-time topometric localization,'' in
	\emph{proceedings of IEEE International Conference on Robotics and
		Automation}.\hskip 1em plus 0.5em minus 0.4em\relax IEEE, 2012, pp.
	1635--1642.
	
	\bibitem{color-hist-map}
	I.~Ulrich and I.~Nourbakhsh, ``Appearance-based place recognition for
	topological localization,'' in \emph{proceedings of IEEE International
		Conference on Robotics and Automation}, vol.~2.\hskip 1em plus 0.5em minus
	0.4em\relax IEEE, 2000, pp. 1023--1029.
	
	\bibitem{cnn-performance}
	N.~S{\"u}nderhauf, S.~Shirazi, F.~Dayoub, B.~Upcroft, and M.~Milford, ``On the
	performance of convnet features for place recognition,'' in \emph{proceedings
		of IEEE/RSJ International Conference on Intelligent Robots and
		Systems}.\hskip 1em plus 0.5em minus 0.4em\relax IEEE, 2015, pp. 4297--4304.
	
	\bibitem{landmarks-cnn}
	N.~Sunderhauf, S.~Shirazi, A.~Jacobson, F.~Dayoub, E.~Pepperell, B.~Upcroft,
	and M.~Milford, ``Place recognition with convnet landmarks: Viewpoint-robust,
	condition-robust, training-free,'' \emph{Proceedings of Robotics: Science and
		Systems XII}, 2015.
	
	\bibitem{semi-semantic}
	S.~Cascianelli, G.~Costante, E.~Bellocchio, P.~Valigi, M.~L. Fravolini, and
	T.~A. Ciarfuglia, ``Robust visual semi-semantic loop closure detection by a
	covisibility graph and cnn features,'' \emph{{Robotics and Autonomous
			Systems)}, volume={92}, pages={53--65}, year={2017}, publisher={Elsevier}}.
	
	\bibitem{edge-boxes}
	C.~L. Zitnick and P.~Doll{\'a}r, ``Edge boxes: Locating object proposals from
	edges,'' in \emph{Proceedings of the European Conference on Computer
		Vision}.\hskip 1em plus 0.5em minus 0.4em\relax Springer, 2014, pp. 391--405.
	
	\bibitem{liu2014topological}
	M.~Liu and R.~Siegwart, ``Topological mapping and scene recognition with
	lightweight color descriptors for an omnidirectional camera,'' \emph{{IEEE
			Transactions on Robotics}}, vol.~30, no.~2, pp. 310--324, 2014.
	
	\bibitem{range-salient}
	P.~Newman and K.~Ho, ``Slam-loop closing with visually salient features,'' in
	\emph{proceedings of IEEE International Conference on Robotics and
		Automation}.\hskip 1em plus 0.5em minus 0.4em\relax IEEE, 2005, pp. 635--642.
	
	\bibitem{pcl-loop-clousure}
	K.~Granstr{\"o}m and T.~B. Sch{\"o}n, ``Learning to close the loop from 3d
	point clouds,'' in \emph{proceedings of IEEE/RSJ International Conference on
		Intelligent Robots and Systems}.\hskip 1em plus 0.5em minus 0.4em\relax IEEE,
	2010, pp. 2089--2095.
	
	\bibitem{pcl-descriptors}
	T.~Cieslewski, E.~Stumm, A.~Gawel, M.~Bosse, S.~Lynen, and R.~Siegwart, ``Point
	cloud descriptors for place recognition using sparse visual information,'' in
	\emph{proceedings of IEEE International Conference on Robotics and
		Automation}.\hskip 1em plus 0.5em minus 0.4em\relax IEEE, 2016, pp.
	4830--4836.
	
	\bibitem{liu2012dp}
	M.~Liu and R.~Siegwart, ``Dp-fact: Towards topological mapping and scene
	recognition with color for omnidirectional camera,'' in \emph{proceedings of
		IEEE International Conference on Robotics and Automation}.\hskip 1em plus
	0.5em minus 0.4em\relax Ieee, 2012, pp. 3503--3508.
	
	\bibitem{Generative}
	E.~Johns and G.-Z. Yang, ``Generative methods for long-term place recognition
	in dynamic scenes,'' \emph{International Journal of Computer Vision}, vol.
	106, no.~3, pp. 297--314, 2014.
	
	\bibitem{2D-lidar}
	R.~Zlot and M.~Bosse, ``Place recognition using keypoint similarities in 2d
	lidar maps,'' in \emph{Experimental Robotics}.\hskip 1em plus 0.5em minus
	0.4em\relax Springer, 2009, pp. 363--372.
	
	\bibitem{GmH}
	O.~Chum, M.~Perd'och, and J.~Matas, ``Geometric min-hashing: Finding a (thick)
	needle in a haystack,'' in \emph{Proceedings of the IEEE Computer Society
		Conference on Computer Vision and Pattern Recognition}.\hskip 1em plus 0.5em
	minus 0.4em\relax IEEE, 2009, pp. 17--24.
	
	\bibitem{deep-learning}
	Y.~LeCun, Y.~Bengio, and G.~Hinton, ``Deep learning,'' \emph{Nature}, vol. 521,
	no. 7553, pp. 436--444, 2015.
	
	\bibitem{deep-learning-overview}
	J.~Schmidhuber, ``Deep learning in neural networks: An overview,'' \emph{Neural
		networks}, vol.~61, pp. 85--117, 2015.
	
	\bibitem{my_fine_grained}
	T.~Sun, L.~Sun, and D.-Y. Yeung, ``Fine-grained categorization via cnn-based
	automatic extraction and integration of object-level and part-level
	features,'' \emph{{Image and Vision Computing}}, 2017.
	
	\bibitem{Visualizing-cnn}
	M.~D. Zeiler and R.~Fergus, ``Visualizing and understanding convolutional
	networks,'' in \emph{Proceedings of the European Conference on Computer
		Vision}.\hskip 1em plus 0.5em minus 0.4em\relax Springer, 2014, pp. 818--833.
	
	\bibitem{learning-deep-features}
	B.~Zhou, A.~Khosla, A.~Lapedriza, A.~Oliva, and A.~Torralba, ``Learning deep
	features for discriminative localization,'' in \emph{Proceedings of the IEEE
		Computer Society Conference on Computer Vision and Pattern Recognition},
	2016, pp. 2921--2929.
	
	\bibitem{scene-recognition}
	S.~Guo, W.~Huang, L.~Wang, and Y.~Qiao, ``Locally supervised deep hybrid model
	for scene recognition,'' \emph{{IEEE Transactions on Image Processing}},
	vol.~26, no.~2, pp. 808--820, 2017.
	
	\bibitem{Caffe}
	Y.~Jia, E.~Shelhamer, J.~Donahue, S.~Karayev, J.~Long, R.~Girshick,
	S.~Guadarrama, and T.~Darrell, ``Caffe: Convolutional architecture for fast
	feature embedding,'' in \emph{{Proceedings of the 22nd ACM international
			conference on Multimedia}}.\hskip 1em plus 0.5em minus 0.4em\relax ACM, 2014,
	pp. 675--678.
	
	\bibitem{return-devil}
	K.~Chatfield, K.~Simonyan, A.~Vedaldi, and A.~Zisserman, ``Return of the devil
	in the details: Delving deep into convolutional nets,'' \emph{arXiv preprint
		arXiv:1405.3531}, 2014.
	
	\bibitem{caffenet}
	L.~Wang, C.-Y. Lee, Z.~Tu, and S.~Lazebnik, ``Training deeper convolutional
	networks with deep supervision,'' \emph{arXiv preprint arXiv:1505.02496},
	2015.
	
	\bibitem{training_cnn}
	R.~Gomez-Ojeda, M.~Lopez-Antequera, N.~Petkov, and J.~Gonzalez-Jimenez,
	``Training a convolutional neural network for appearance-invariant place
	recognition,'' \emph{arXiv preprint arXiv:1505.07428}, 2015.
	
	\bibitem{pcl}
	R.~B. Rusu and S.~Cousins, ``3d is here: Point cloud library (pcl),'' in
	\emph{proceedings of IEEE International Conference on Robotics and
		Automation}.\hskip 1em plus 0.5em minus 0.4em\relax IEEE, 2011, pp. 1--4.
	
	\bibitem{VFH}
	R.~B. Rusu, G.~Bradski, R.~Thibaux, and J.~Hsu, ``Fast 3d recognition and pose
	using the viewpoint feature histogram,'' in \emph{proceedings of IEEE/RSJ
		International Conference on Intelligent Robots and Systems}.\hskip 1em plus
	0.5em minus 0.4em\relax IEEE, 2010, pp. 2155--2162.
	
	\bibitem{ESF}
	W.~Wohlkinger and M.~Vincze, ``Ensemble of shape functions for 3d object
	classification,'' in \emph{IEEE International Conference on Robotics and
		Biomimetics (ROBIO)}.\hskip 1em plus 0.5em minus 0.4em\relax IEEE, 2011, pp.
	2987--2992.
	
	\bibitem{GRSD}
	Z.-C. Marton, D.~Pangercic, R.~B. Rusu, A.~Holzbach, and M.~Beetz,
	``Hierarchical object geometric categorization and appearance classification
	for mobile manipulation,'' in \emph{The 10th IEEE-RAS International
		Conference on Humanoid Robots (Humanoids)}.\hskip 1em plus 0.5em minus
	0.4em\relax IEEE, 2010, pp. 365--370.
	
	\bibitem{openfabmap}
	A.~Glover, W.~Maddern, M.~Warren, S.~Reid, M.~Milford, and G.~Wyeth,
	``Openfabmap: An open source toolbox for appearance-based loop closure
	detection,'' in \emph{proceedings of IEEE International Conference on
		Robotics and Automation}.\hskip 1em plus 0.5em minus 0.4em\relax IEEE, 2012,
	pp. 4730--4735.
	
	\bibitem{opencv}
	G.~Bradski, ``The opencv library.'' \emph{{Dr. Dobb's Journal: Software Tools
			for the Professional Programmer}}, vol.~25, no.~11, pp. 120--123, 2000.
	
\end{thebibliography}
\end{document}